\documentclass{article}

\usepackage{booktabs}
\usepackage{float}
\usepackage{array}
\usepackage{multirow}
\usepackage{amssymb}
\usepackage[utf8]{inputenc} % allow utf-8 input
\usepackage[T1]{fontenc}    % use 8-bit T1 fonts
\usepackage{amsmath}
\usepackage{amssymb}
\usepackage{algorithm}
\usepackage{algpseudocode}
\usepackage{hyperref}       % hyperlinks
\usepackage{url}            % simple URL typesetting
\usepackage{booktabs}       % professional-quality tables
\usepackage{amsfonts}       % blackboard math symbols
\usepackage{nicefrac}       % compact symbols for 1/2, etc.
\usepackage{microtype}      % microtypography
\usepackage{lipsum}
\usepackage{fancyhdr}       % header
\usepackage{graphicx}       % graphics
\usepackage{float}
\usepackage{dcolumn}        % for stargazer decimal alignment
\usepackage{amsmath}
\graphicspath{{media/}}     % organize your images and other figures under media/ folder
\usepackage[authoryear,round]{natbib}

\setcitestyle{authoryear,round}

\usepackage[margin=1in]{geometry}

\title{Reasoning as Pattern Matching: Shared Mechanisms in Human and LLM Everyday Reasoning
}

\author{
\begin{minipage}[t]{0.45\textwidth}
\centering
\textbf{Zach Studdiford}\\
University of Wisconsin--Madison\\
Department of Psychology\\
\texttt{studdiford@wisc.edu}
\end{minipage}
\hfill
\begin{minipage}[t]{0.45\textwidth}
\centering
\textbf{Gary Lupyan}\\
University of Wisconsin--Madison\\
Department of Psychology\\
\texttt{lupyan@wisc.edu}
\end{minipage}
}

\begin{document}
\maketitle
\begin{abstract}
	When large language models (LLMs) fail to generalize or make content-sensitive errors in reasoning, it is often taken as evidence that LLMs are not truly reasoning, but rather performing a kind of pattern matching. The implication is that peoples’ behavior does not exhibit the same types of failures because human reasoning relies on principled and abstract world models. We test this assumption by first evaluating humans and LLMs on their ability to engage in common-sense reasoning about a variety of everyday situations. Our results reveal convergent patterns of reasoning across 46 LLMs and two cohorts of human participants. We then ask whether this behavioral convergence is due to LLMs having acquired abstract world models or a set of pattern-matching heuristics by characterizing the roles of abstract and content-sensitive model neurons in producing human-like responses. We find that while LLMs encode both abstract and content-sensitive representations, it is content-sensitive mechanisms which are causally responsible for aligning models with humans. Taken together, our results suggest that everyday causal reasoning in people and LLMs makes heavy use of pattern-matching.

\end{abstract}

\section{Introduction}
People routinely predict the outcomes of everyday actions such as inferring that a glass is more likely to break when it falls onto tile than onto a carpet or that a person facing North will face South after turning 180 degrees. It is often argued that underlying this ability are ``world models'':  structure-preserving, behaviorally efficacious representations of the entities and processes in the real world \citep{yildirimTaskStructuresWorld2024, tenenbaum2011grow, lake2017building, kemp2009structured}. That is, when we think about a glass falling onto a floor, we don’t simply retrieve a memorized fact or rely on associations between ``glass'' and ``floor'', but access a structured model that, e.g., computes the likelihood of a fragile object shattering when it hits a hard surface, or instantiates a geometric transformation given a starting point and a trajectory \citep{gerstenberg2021counterfactual}. Defining properties of these structured models include explicit representations of causal structure \citep{pearl2018book,sloman2009causal}: Someone who understands that facing North and turning 180 degrees implies facing South should also understand that facing South and turning 180 degrees implies facing North \citep{fodor1988connectionism}. These causal models are also thought to have role-filler independence \citep{quilty-dunn_best_2022}. For example, the representation of the relation \textit{North of} should be the same regardless of what landmark is under discussion \citep{fodor1988connectionism,quilty-dunn_best_2022}. If one understands that being North of something implies that the something is Southward, it should not matter what that something is.

Against this backdrop, it is striking that large language models (LLMs) are increasingly able to perform many tasks thought to require such structured world models. For example, LLMs can generate plausible causal explanations \citep{kiciman2023causal}, reason about the intentions of agents\citep{kosinski2024evaluating} and induce sequence-generation rules in context\citep{hendel2023context}. They learn to do this without any of the inductive priors \citep{lake2017building} or abstract representations \citep{wong2023word, pearl2018book} often argued to be essential for learning causal structure. Over the course of pretraining, LLM representations evolve from simple two and three word associations \citep{chang2022word} to eventually encoding patterns such as rare syntactic structures \citep{tenney2019bert, misra2024language} and structured representations of color and space \citep{marjieh2024large, fukushima2025advancements}. With sufficient training, there emerge even seemingly \textit{compositional} representations \citep{lepori2023break}, allowing LLMs to infer abstract relationships about symbol sequences \citep{hendel2023context}, game states \citep{hazineh2023linear}, and properties of places and objects \citep{khandelwal2025language}. 

One interpretation of these developments is that LLMs have learned structured world models of the type thought to underlie human reasoning \citep{lepori2023break, hahn2023theory}. Tempering this interpretation, however, is evidence that causal ``reasoning'' in LLMs often does not show the kind of compositionality, role-filler independence, and overall degree of abstraction that one would expect if they are using structured world models. Instead LLMs often succeed on one version of a problem while failing on other structurally identical versions \citep{mccoy2023embers, suresh2023conceptual, ivanova2025elements, ullman2023large, shojaeeIllusionThinkingUnderstanding2025}. This uneven performance profile has led some to argue that instead of relying on structured world models--assumed to underlie human reasoning--LLMs are \textit{merely} ``pattern-matching'' \citep{yildirimTaskStructuresWorld2024, mahowald2024dissociating, lewisUsingCounterfactualTasks2024}, relying on shallow associations rather than real understanding or reasoning \citep{ying2026grounding,chang2025characterizing, kumar2023disentangling}, e.g., inferring that a knocked over glass may break simply because of the co-occurrence of ``glass'' and ``break''.

A closer look at \textit{human} behavior, however, reveals some similar deviations from compositionality and role-filler independence that are expected if people are using structured world models. For example, people's performance on logical reasoning problems is highly content-sensitive. Success in applying a rule of the form ($P \rightarrow Q$) in one setting does not imply the ability to use it correctly in another setting  \citep{pollard1981effect, cox1982effects, wason1968reasoning}. People struggle to separate the logical structure of a problem from its content \citep{evansConflictLogicBelief1983a} and the ability to solve one version of a problem does not imply the ability to solve structurally identical problems \citep{kotovskyWhyAreProblems1985, johnson2010mental}. Knowledge of a rule does not imply the ability to apply it consistently. For example, despite being able to correctly state how to distinguish odd from even numbers, people not only take longer to judge that a number like 798 is even, but often mistake it for odd. Many of those who respond correctly nevertheless assert that 798 is ``less even'' than 400 \citep{lupyan_difficulties_2013}. In sum, many of the fail-states that are taken as evidence that LLMs do not ``really'' reason \citep{arkoudasGPT4CantReason2023} or think \citep{shojaeeIllusionThinkingUnderstanding2025} find close analogs in human behavior. 

One explanation for such deviations from normative behavior is that they arise arise from rational (model-based) inference that is constrained by time, memory, and computation \citep{griffiths2015rational, liederResourcerationalAnalysisUnderstanding2019, yildirimTaskStructuresWorld2024}. We consider another possibility: Pattern-matching is neither shallow nor a fallback mechanism, but the way neural networks solve problems \citep{rumelhart_parallel_1986}. Cognition, on this view, is ``built out of sequences of patterned responses to cues'' \citep[][p. 72]{margolisPatternsThinkingCognition1990} and reasoning involves relating the input to patterns: semantic knowledge at various levels of abstraction, with the entire system settling on a response in the manner of an analog computer \citep[][p. 68]{margolisPatternsThinkingCognition1990, rumelhart_parallel_1986}. On this view, processing of inputs is inherently contextual, recruiting background knowledge about the particular entities involved that supply further, often unstated, context. Unlike a structured world model, a pattern-matching system is not expected to be fully compositional and displays only partial role-filler independence. Its response is cued by the whole content-laden configuration, so seemingly irrelevant changes in content can change the inference: sometimes for the better, and sometimes for the worse. 

%misinterpretations of the task \citep{oaksford1994rational}.

If both human and LLM reasoning relies pattern matching, responses in both systems should be influenced by the full content of a problem rather than by its abstracted form. For example, consider predicting the outcome (by filling in the blank) in the following two scenarios: \textit{``Chicago is North of Ali. Ali turns around. Chicago is \texttt{BLANK} Ali''} and \textit{``The painting is North of Ali. Ali turns around. The painting is \texttt{BLANK} Ali''}. If one has the right world model (i.e., knows that facing direction has no bearing on whether one is North or South of a referent), the two prompts should be equally easy. The world model should abstract away from ostensibly irrelevant details like what the referent is. In a pattern-matching system, however, such details matter. The prompt about the painting may evoke an indoor scene such as a museum. This scene is unlikely to evoke a geocentric reference frame. In contrast, ``Chicago'' evokes an outdoor scene along with a geocentric frame in which cardinal directions are fixed features of the world; that turning around plainly leaves them unchanged.

\section{The present study}
\begin{figure}[H]

    \centering
    \includegraphics[width=0.90\linewidth]{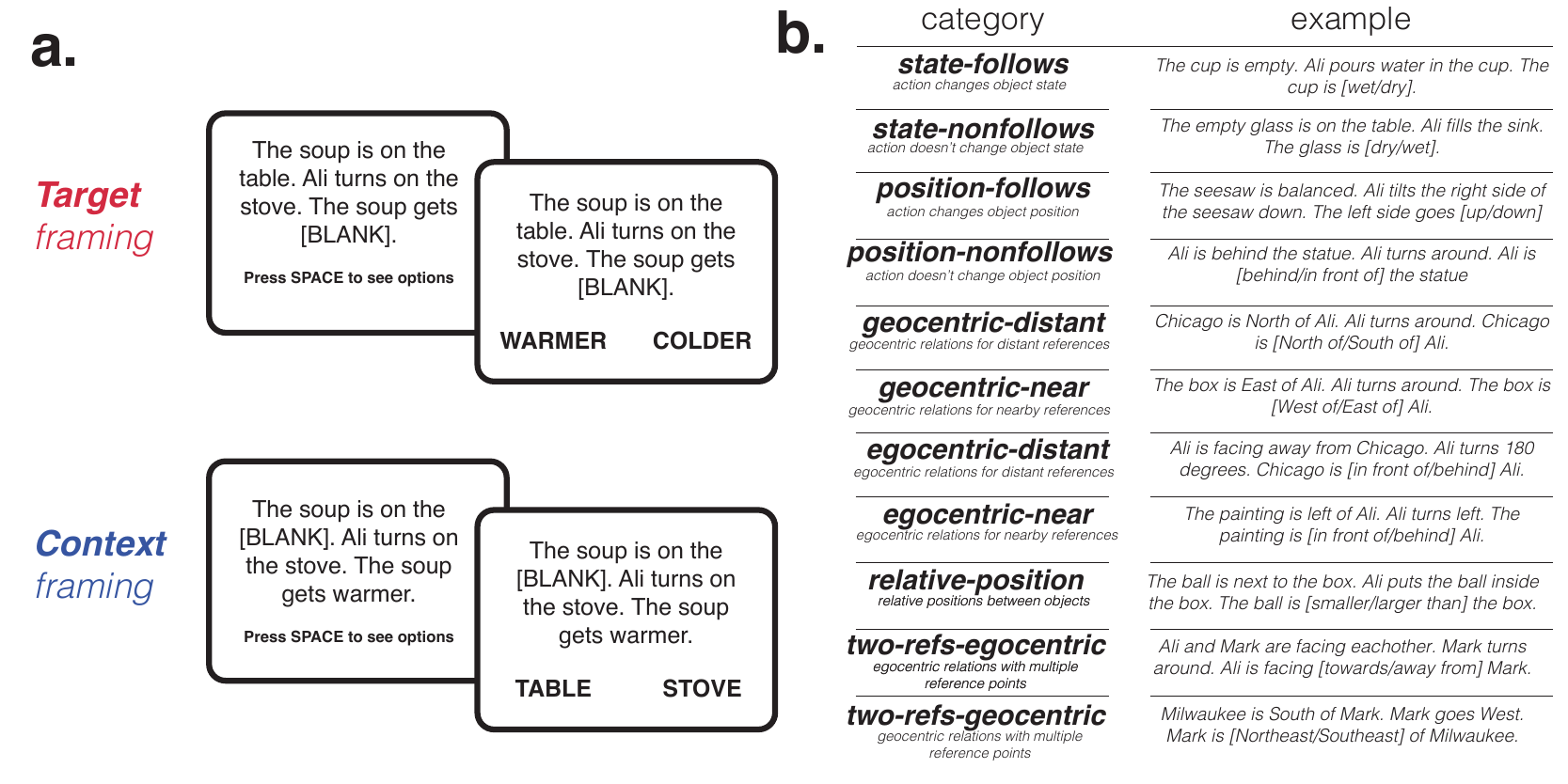}
    \caption{\textbf{Overview of our evaluation probing human and LLM causal reasoning}. 
    \textbf{a.} Illustration of the evaluation format used for testing causal reasoning in people and LLMs. For each prompt, subjects are first presented with the scenario. After reading the prompt, subjects press \texttt{SPACE} to see the two response options and then select the most likely completion. \textbf{b.} Summary of the 11 categories we used in our evaluation of everyday causal reasoning.
    }
    \label{fig:stim_overview}

\end{figure}

We evaluate whether the reasoning of people and LLMs is more consistent with abstract world models or pattern matching. Previous studies have investigated LLM and human reasoning on tasks including syllogisms \citep{lampinen2024language, eisape2024systematic}, logical rule induction \citep{loo2026llms}, and analogical reasoning \citep{webb2023emergent}. In contrast to these studies, where the correct response can be determined by applying abstract logical inference, we probe reasoning in content-rich everyday scenarios modeled after \citet{ivanova2025elements}. Successfully reasoning about causal relations tested in each scenario requires accessing relevant world knowledge, such as inferring that facing forward and turning around would result in facing the opposite direction, or that spilling a glass of water on a table would likely make the table wet. To ensure a more stringent test of world models as opposed to mere associations (e.g that "\textit{glass}" co-occurs with "\textit{breaks}"), we examine various measures of association strength and include scenarios that require inhibiting salient associations. While these problems possess some inherent ambiguity, we ask participants and LLMs to assess the \textit{relative} probability of one minimal-difference response over another \citep[as in][]{ivanova2025elements}. In each scenario, there are two options provided as completions for \texttt{BLANK}, such that one completion makes the statement more likely relative to the other (e.g., it is more likely that \texttt{BLANK} = ``\textit{breaks}'' if the glass is bumped off a table relative to ``\textit{doesn't break}'').  The scenarios we use draw on a variety of types of knowledge. For example, the categories \textit{geocentric-near} and \textit{egocentric-near} evaluate knowledge of geocentric and egocentric relations when reasoning about nearby objects (such as paintings, boxes, or people), while \textit{geocentric-distant} and \textit{egocentric-distant} evaluate the same knowledge in the context of distant references, such as mountains or cities. The full set of evaluation categories and example prompts is shown in Figure \ref{fig:stim_overview}b. 

In addition to testing subjects on their ability to reason about the \textit{outcomes} of various causal relations, we also probe the ability to infer the plausible \textit{cause} of an outcome in the same scenarios. For instance, in the prompt ``The glass is on the table. Ali bumps the \texttt{BLANK}. The glass breaks'', participants are presented with options \textit{BLANK=lamp} and \textit{BLANK=glass}. Successful inference in this case requires understanding that bumping the \textit{glass} would be the most likely cause of the glass breaking. Using a modular stimulus format inspired by \citet{ivanova2025elements}, we define a \textit{target framing} template for each prompt where the likely \textit{outcome} must be inferred, and a \textit{context framing} template which requires reasoning about plausible \textit{causes} given some outcome (see Figure \ref{fig:stim_overview}a). This allows us to test whether participants can robustly infer the likely effects and outcomes of a causal relation in multiple directions \citep{pearl2018book}. Full details of the causal reasoning evaluation are provided in supplementary methods (Section \ref{eval-overview}).

To foreshadow our results, we find that people's reasoning about everyday causal scenarios is sensitive to differences in context and to small differences in ostensibly irrelevant details. For example, people reason more accurately about the direction of a person who is North of \textit{Chicago} compared to a person who is North of a \textit{painting}. Moreover, accuracy is in some cases dependent on how this knowledge is probed: people reason with high accuracy about plausible causes in prompts such as ``The soup is on the \texttt{BLANK}. Ali turns on the stove. The soup gets colder.'' (answer=``table''), but frequently fail when tasked with inferring the outcome in prompts like ``The soup is on the table. Ali turns on the stove. The soup gets \texttt{BLANK}.'' (answer = ``colder''). Our results demonstrate that both people and LLMs show similar profiles characterized by graded generalization: small, surface-level changes to the content and format of prompts lead to seemingly unpredictable errors. 

Such inconsistencies do not necessarily imply that an underlying pattern-matching mechanism is at work \citep{lampinen2024language}. It could instead be the case that there is a core competency for abstract representations that is expressed in some cases but not others, both in people \citep{firestone2020performance} and in language models \citep{hu2024auxiliary}. Conversely, behavior in line with abstractions and content-invariant compositionality could possibly be realized through content-sensitive neural mechanisms \citep{enyan2026unifying}. This distinction between the \textit{performance} of a system (human or LLM) on a given task and the underlying \textit{competence} of that system's internal representations make it difficult to infer whether abstract mechanisms or pattern-matching based heuristics actually underlie the behavior of the system (See \citet{firestone2020performance} for an overview of this debate).

Although we cannot directly probe human cognitive representations, we \textit{can} causally manipulate those of LLMs \citep{olah2018building}. Using minimal difference prompt substitutions, we first locate neurons that are more sensitive to changes in specific content and those more sensitive to changes to abstract structural properties of the causal scenario. We then ask whether activity in these content-sensitive or structure-sensitive neurons provide better predictions of human behavior. Finally, we perturb the activity in these neurons and measure the consequences of these perturbations on the model's alignment to human behavior. These analyses show that although LLMs encode both abstract and content sensitive representations, it is the neurons that encode information about specific content of the scenario that are disproportionately responsible for model-human alignment. Taken together, our results support the conclusion that LLMs generalize in human-like ways \textit{because} they are pattern matching, rather than in spite of it.

\section{Results}

\subsection{Graded accuracy in human and LLM reasoning}
\label{sec:main-eval}
We begin by evaluating performance on our causal reasoning task in humans($n=142$) and 46 LLMs (42 open-weights and 4 proprietary models). Consistent with our hypothesis that causal reasoning in humans and LLMs can be explained as a kind of graded pattern-matching, we find that accuracy varies for seemingly arbitrary changes in the context and framing of a given prompt. Performance in both human participants and LLMs is surprisingly fragile: minor changes to prompts frequently induced fail states in both people and LLMs. These results are not explainable as inattention or guessing: in a replication using a second group of participants, we find that subjects tend to commit the same errors even when presented with the same prompt twice, and will even justify their incorrect responses (see Appendix \ref{app:retest}). Below we describe the results of our behavioral experiments in more detail. Additional details of our methods and evaluation are provided in section \ref{eval-overview}.

\begin{figure}
    \centering
    \includegraphics[width=1\linewidth]{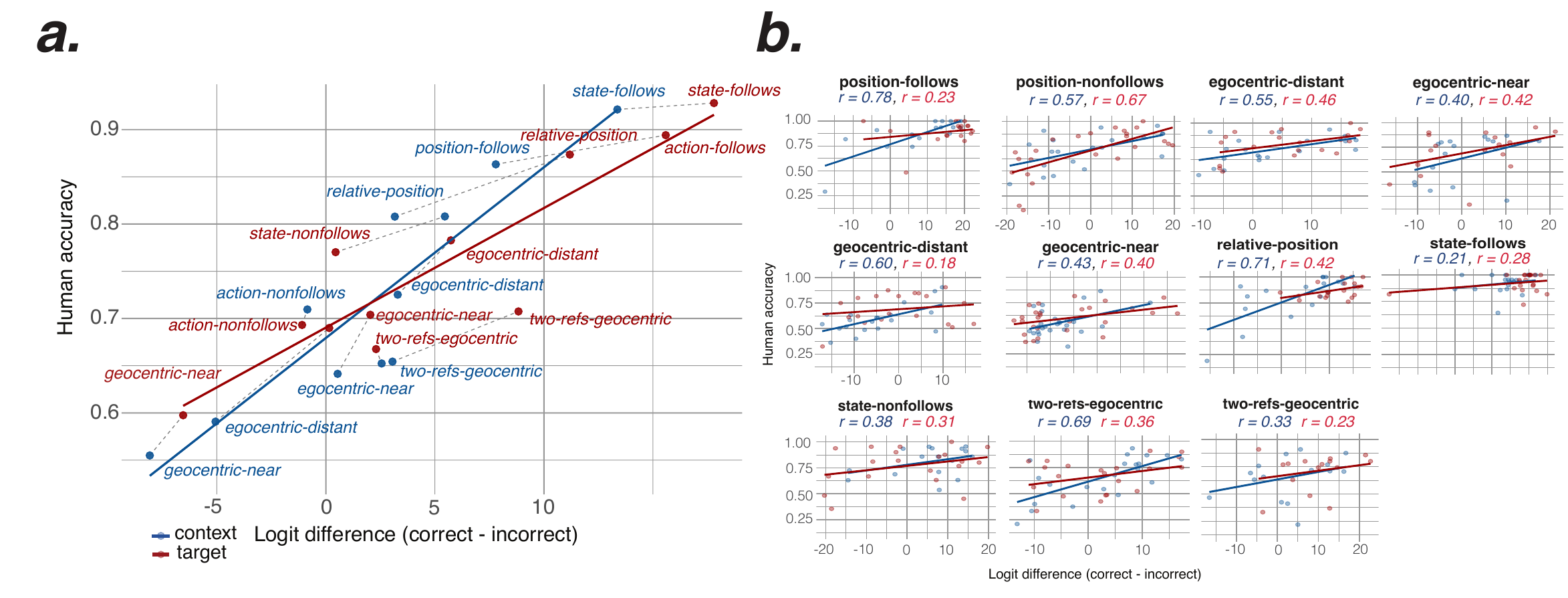}
    \caption{Behavioral results on our reasoning task (\texttt{gemma-4-12b}; results for other models are included in appendices \ref{app:cat-alignment} and \ref{app:item-align}). \textbf{a. Category alignment in humans and LLMs.} Humans and LLMs showed similar error patterns. The x-axis indicates the relative \textit{logit difference} between the correct next token output $A$ and the incorrect next token output $B$. The y-axis shows mean human accuracy for a given category and prompt type. Different prompt types of the same category are connected by dotted gray lines. Blue (context) and red (target) lines are best fits for target and context prompt types. {\textbf{b. Item-level accuracy.} Within-category accuracy and lines of best fit are shown for each prompt format condition. We include Pearson $r$ values for correlations with human accuracy separated by prompt format and category.} 
    }
    
    \label{fig:behavioral}
\end{figure}

\paragraph{People and LLMs make similar errors.}

Human accuracy on our reasoning evaluation was well below ceiling ($\mu=0.71$, $\sigma=0.21$), and varied widely across categories (Figure~\ref{fig:behavioral}a). The most accurate open-weights model, \texttt{gemma-4-31b},\footnote{All models described here underwent post-training unless otherwise specified.} performed on-par with human participants. Interestingly, our participants are outperformed by frontier models (\texttt{Deepseek-R1}: $\mu=0.90$; \texttt{GPT-5.2}: $\mu=0.87$; both $p<0.001$, two-proportion $z$-tests), although these models were less aligned with human responses (see Figure~\ref{fig:model_acc_predict} for $R^2$ comparisons of LLMs predicting human accuracy). In comparison, the accuracy of \texttt{gemma-4-12b}, the model with the best fit to human
data, was underwhelming in absolute terms, with an overall mean of 0.61.

\begin{figure}[]
    \centering
    \includegraphics[width=1\linewidth]{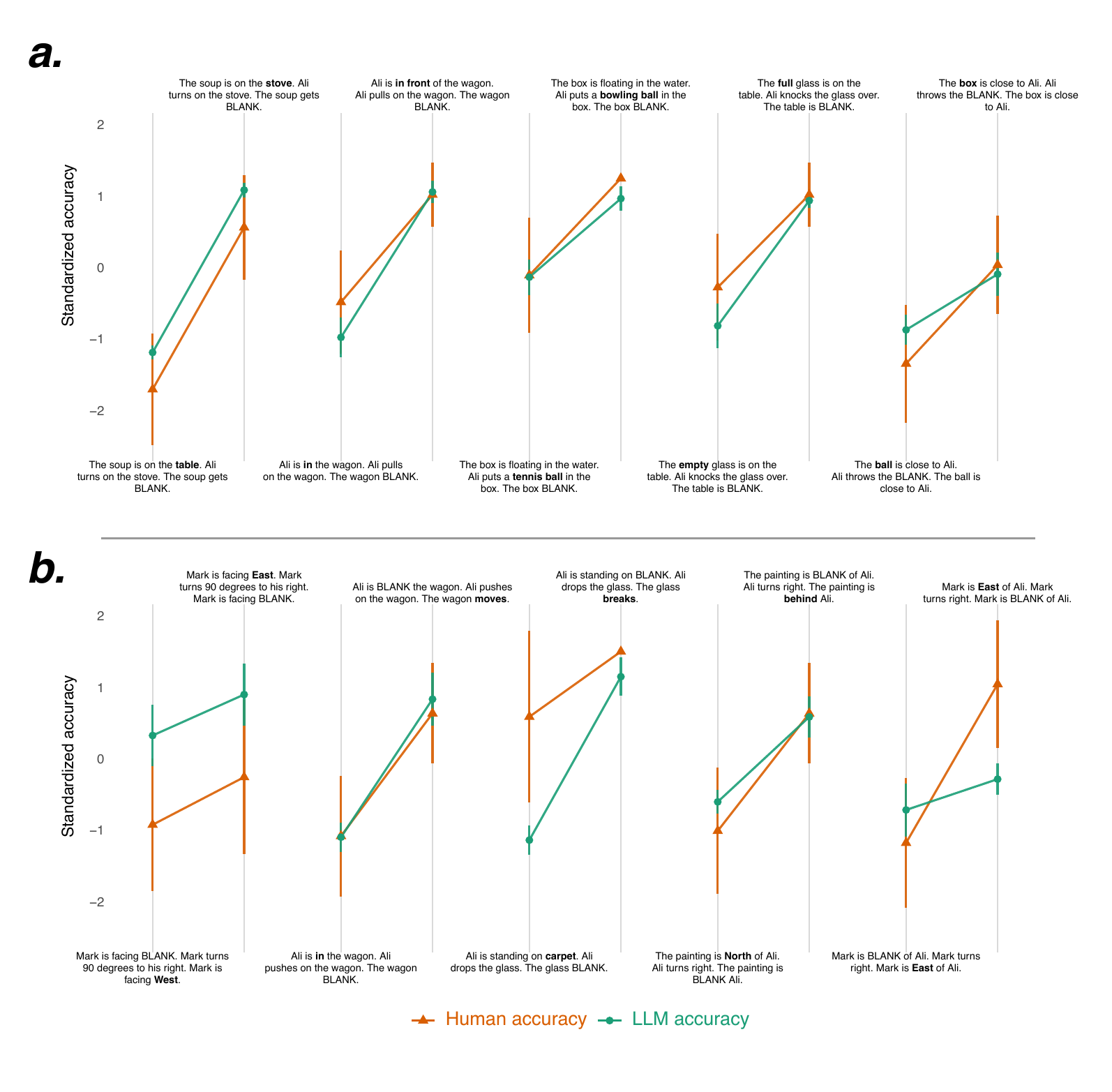}
    \caption{\textbf{Similar inconsistencies in human and LLM reasoning.} \textbf{a.} Minimal counterfactual changes to causal reasoning problems (changing one word to modify the ground-truth prompt outcome) elicit similar fail states in people and models. For each prompt pair, the left axis tick indicates standardized human and LLM accuracy for the lower prompt, while the right axis tick shows human and LLM for the upper prompt. Error bars correspond to 95\% confidence intervals in human and model responses. CIs and point estimates are obtained for LLM responses from all 46 open-weights models in our evaluation. \textbf{b.} Human-LLM response similarities for prompt pairs which differ only in context or target framing. 
    }

    % We include several examples of contrasting problem pairs in \textit{state-follows} with high accuracy and \textit{state-nonfollows} with low accuracy (top) and pairs of prompts in either the \textit{target} or \textit{context} stimulus framings (bottom). In all cases, problems only differ by 1-2 words, yet we observe large differences in human accuracy that are mirrored by the responses of \texttt{gemma-3-27b}.}

    \label{fig:min_pairs}
\end{figure}

Figure \ref{fig:behavioral}a shows alignment between human responses and \texttt{gemma-4-12b}.
The model shows high category-level alignment with human accuracy ($r(20) = 0.89$,
$p < 0.001$). For example, for both people and \textit{gemma-4-12b}, geocentric relations in the context of nearby objects (\textit{geocentric-near}, e.g.\ \textit{``The painting is North of Ali''};
$M_{\text{human}} = 0.58$, 95\% CI $[0.54, 0.61]$; $M_{\text{gemma}} = -7.31$,
$[-9.54, -5.08]$) were more difficult compared to geocentric relations for distant references
such as cities (\textit{geocentric-distant}, e.g.\ \textit{``Milwaukee is North of Ali''};
$M_{\text{human}} = 0.64$, $[0.59, 0.69]$; $M_{\text{gemma}} = -2.45$, $[-5.47, 0.56]$).
Egocentric relations were also more difficult in the context of nearby objects
(\textit{egocentric-near}, e.g.\ \textit{``The painting is in front of Ali''};
$M_{\text{human}} = 0.67$, $[0.60, 0.75]$; $M_{\text{gemma}} = 1.29$, $[-2.00, 4.58]$)
compared to distant references (\textit{egocentric-distant}, e.g.\
\textit{``Milwaukee is in front of Ali''}; $M_{\text{human}} = 0.75$, $[0.71, 0.80]$;
$M_{\text{gemma}} = 4.51$, $[1.72, 7.31]$). Although humans and \textit{gemma-4-12b} were able
to infer the outcome of events where a state or position change likely follows
(\textit{state-follows}, $M_{\text{human}} = 0.92$, $[0.90, 0.95]$; $M_{\text{gemma}} = 15.69$,
$[13.27, 18.12]$; \textit{action-follows}, $M_{\text{human}} = 0.88$, $[0.83, 0.93]$;
$M_{\text{gemma}} = 12.02$, $[8.53, 15.51]$), small changes to the prompt in
\textit{state-nonfollows} and \textit{action-nonfollows} (where a change in outcome is no
longer probable) lead to more frequent errors (\textit{state-nonfollows},
$M_{\text{human}} = 0.79$, $[0.73, 0.84]$; $M_{\text{gemma}} = 2.67$, $[-1.30, 6.65]$;
\textit{action-nonfollows}, $M_{\text{human}} = 0.70$, $[0.64, 0.76]$;
$M_{\text{gemma}} = -0.98$, $[-4.78, 2.82]$).

Humans and LLMs also show a moderate degree of convergence \textit{within} specific category and prompt frames. Figure \ref{fig:behavioral}b shows item level correlations for all categories\footnote{This alignment extends to additional measures of model behavior beyond accuracy and logit outputs: \textit{reasoning traces} from a model trained to use chain-of-though (\texttt{Deepseek-R1}) also offer fine-grained predictions of human accuracy, see Appendix \ref{app:reasoning_traces_predict}.}. People and \texttt{gemma-4-12b-it} are more aligned for the \textit{context} prompt framing relative to the \textit{target} prompt framing ($b = -0.035$, $SE = 0.015$, $t(409) = -2.41$, $p = 0.017$). A mixed-effects model predicting LLM accuracy from human responses with fixed effects for category and prompt format and random effects for each model indicates that item-level convergence is broadly consistent across models ($b = 0.027$, $SE = 0.003$, $t(41) = 8.52$, $p < 0.001$). Even when accounting for general measures of word distributions--mean trigram frequency and \texttt{GPT-2} surprisal--models remains predictive of human behavior ($b = 0.026$, $SE = 0.003$, $t(41) = 8.31$, $p < 0.001$; see Appendix \ref{app:regression-item-level} for regression details, see \ref{app:item-align} for item-level correlations in other models).

Figure \ref{fig:min_pairs} shows examples of item-level similarities in human and model responses, including prompts that differ by just a single word (\ref{fig:min_pairs}a), and test the same problems that differ in prompt framing (\ref{fig:min_pairs}b). These single word substitutions are sufficient to elicit more or less accurate responses in both models and people, suggesting that high accuracy in one problem does not necessarily generalize to problems that  ostensibly recruit the same world model. For example, high accuracy on the problem \textit{The soup is on the \textbf{stove}. Ali turns on the stove. The soup gets \texttt{BLANK}} (answer=\textit{``warmer''})--achieved by both people and LLMs--might suggest that both should reason competently about the scenario\textit{The soup is on the \textbf{table}. Ali turns on the stove. The soup gets \texttt{BLANK}} (answer=\textit{``colder''}), yet both people and models do not. Similarly, the same problem interpolated in a different framing sometimes elicits different responses: Both people and models are highly accurate for \textit{Ali is BLANK the wagon. Ali pushes on the wagon. The wagon moves}, and systematically incorrect for \textit{Ali is \textbf{in} the wagon. Ali pushes on the wagon. The wagon BLANK}.

\paragraph{Consistency in human and LLM errors} The relatively low accuracy of human subjects raises the possibility that errors stem from inattention or random guessing rather than consistent reasoning errors. To determine whether this is the case, we evaluated a second group of participants ($n = 70$). After presenting subjects with an initial set of prompts, we solicited additional responses for questions they previously answered incorrectly (presented randomly at a later point in the trial sequence). We observe moderately consistent test-retest reliability in participant accuracy between the first and second prompt presentations ($r = 0.60$, $p < 0.001$). Additionally, we fit a mixed effects regression predicting whether a participant was correct on the second stimulus presentation from first presentation correctness, with random intercepts for subject and prompt. We find that getting a question wrong on the first presentation corresponds to a 3.84 likelihood increase in getting that question wrong on the second presentation, (95\% CI $[2.92, 5.07]$, $z = 9.57$, $p < 0.001$). In sum, individual participant errors cannot be fully explained by general inattention or random guessing, and are broadly consistent across stimulus presentations (we note variability in this consistency across prompt categories, see Appendix \ref{app:retest}).

We also note a strong degree of consistency in the responses of various language models: the particular reasoning errors and successes described above are remarkably consistent across the range of LLMs we evaluated. As seen in Appendix \ref{app:mean-cat-alignment}, which includes category and item-level accuracies averaged across all models tested, patterns of responses--such as greater accuracy in \textit{geocentric-distant} relative to \textit{geocentric-near}, and lower accuracy in \textit{egocentric-near} relative to \textit{egocentric-distant}--are not a quirk of one particular model but characterize the responses of language models more generally. Even at the item level, seemingly idiosyncratic failure modes are instead consistent across different models (see Figure \ref{fig:min_pairs}). Moreover, we find that language models are often as aligned within category and prompt format as individual human subjects are with each other: \texttt{gemma-4-12b} is as correlated with individual participant responses as participants are with themselves, for 15 of the 22 prompt format - category combinations (see Appendix \ref{app:held-out-predict}).

Is it possible that the close alignment between human and LLM responses is inflated by accidentally including LLM responses in our human data? Although our participants were verified through CloudResearch and the data were collected prior to recent advancements in agentic models \citep{kay2025you}, we examined whether human-labeled response patterns were suspiciously close 
to LLMs, that is, whether there is any evidence of a bimodal pattern in human-to-model alignment suggesting some participants were actually LLMs. Hartigans' dip test provides no evidence against unimodality ($D = 0.031$, $p = 0.433$, $n = 142$ participants, see Appendix \ref{app:dip-test}).

\subsection{Content-sensitive mechanisms drive human-like responses in LLMs} 
\label{sec:internal-mech-predict}

Our behavioral results reveal notable inconsistencies in both human and LLM commonsense reasoning. Prompts that one might assume to be testing the same world knowledge--such as \textit{The painting is North of Ali. Ali turns around...} and \textit{The mountain is North of Ali. Ali turns around}--elicit similar discrepancies in accuracy across multiple participant cohorts and many different LLMs.

We next examined the internal mechanisms (sets of attention heads) underlying LLM behavior to determine whether these patterns of responses are attributable to graded (but still abstract) representations, or sets of more content-sensitive heuristics consistent with the models performing forms of pattern matching. Specifically, we focused on the role of model \textit{attention heads} in reasoning behavior, and operationalized \textbf{abstraction-sensitive} and \textbf{content-sensitive} attention heads as follows:

\begin{enumerate}
    \item \textbf{Abstraction-sensitive heads:} Attention heads encoding abstract representations should increase the probability of the model making a specific answer choice conditional on whether a given abstract concept/relation is present in the input. For example, a neuron that encodes an abstract representation of \textit{``X is North of Y''} should affect the output distribution conditional on whether \textit{``North of''} is in the prompt (as opposed to, e.g., \textit{``South of''} or \textit{``left of''}. Moreover, the neuron should display a degree of compositionality: differences in ``X'' and ``Y'' should not affect the neuron's activation.
    \item \textbf{Content-sensitive heads:} Attention heads encoding content-sensitive representations should display the inverse pattern of behavior. Given a relation such as \textit{``X is North of Y''}, the neuron should affect the output conditional on the values of ``X'' and/or ``Y''  and not ``North of''. For example, the neuron should contribute similarly to the output for \textit{``The \textbf{painting} is North of Ali''} and \textit{``The \textbf{painting} is South of Ali''}, and differ between \textit{``The \textbf{painting} is North of Ali''} and \textit{``The \textbf{mountain} is North of Ali''}.
\end{enumerate}

%GL: THe caption below still mentions gemma-2. Intentional?
\begin{figure}[h]
    \centering
    \includegraphics[width=0.75\linewidth]{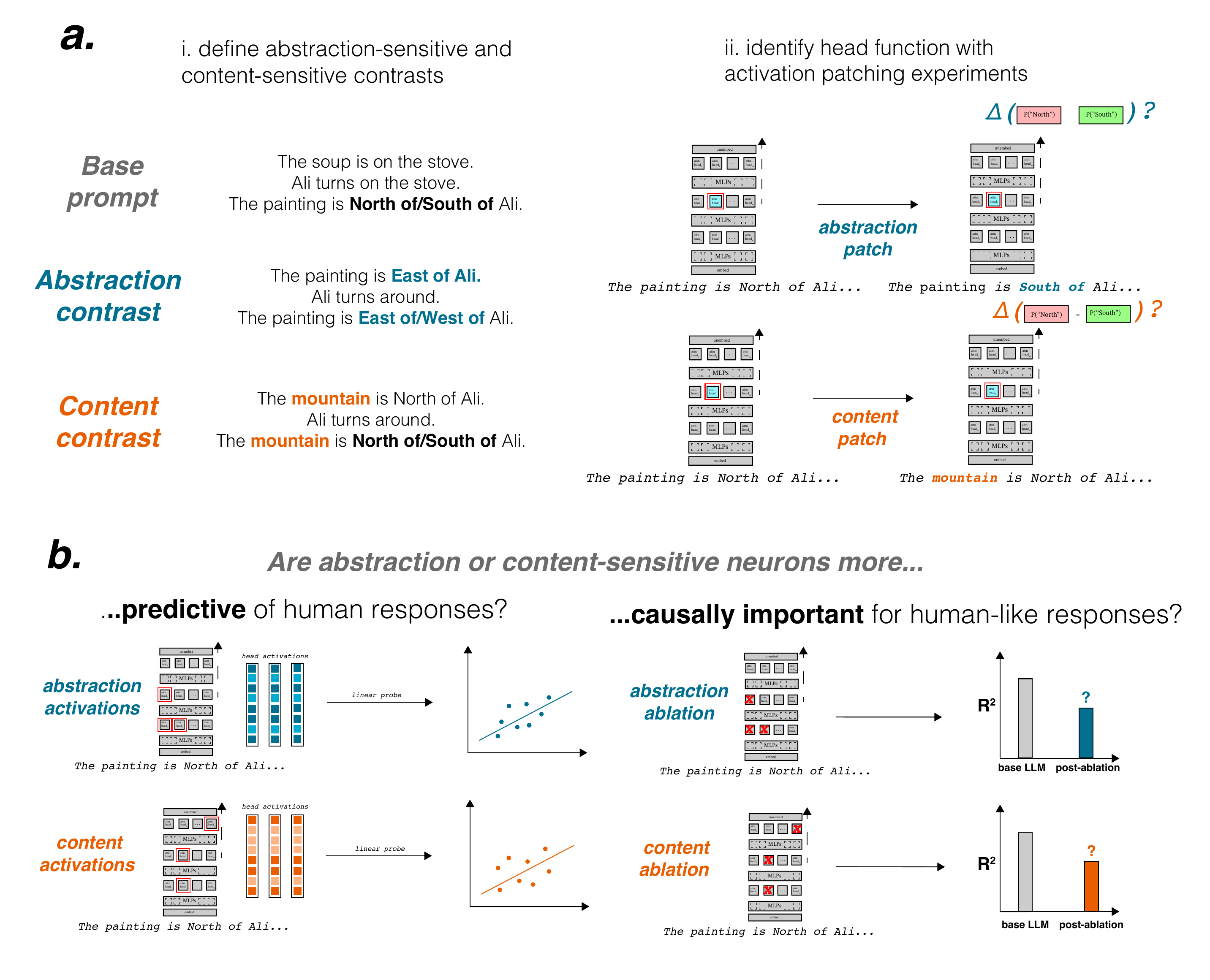}
    \caption{Comparing the roles of abstraction and content-sensitive attention heads in human-like responses. \textbf{a (i). Overview of our procedure for characterizing abstraction and content-sensitivity of model attention heads}. For each base prompt in our evaluation (gray), we define three \textit{abstraction contrasts} where a critical abstract relation in the prompt is modified (blue), and three \textit{content contrasts} where the non-relevant prompt content is modified (orange). \textbf{(ii).} For each attention head in a given model, we replace (patch) the head's activations while processing the base prompt with the head's activations while processing one of the content or critical contrasts. The resulting change in model outputs (measured as the change in logits of the correct and incorrect responses) indicates whether a given head encodes (and causally effects model outputs conditional on) \textit{abstraction} or \textit{content} sensitive information. \textbf{b. Relating abstraction and content-sensitive heads to human behavior.} We first test whether content-sensitive and abstraction-sensitive head outputs predict human behavior. We then perform \textit{causal interventions} to test whether abstraction-sensitive or content-sensitive heads are more important for aligning models to patterns of human responses.
    }
    
    \label{fig:proc-ablations}
\end{figure}

Figure \ref{fig:proc-ablations}a describes our method for characterizing the degree of \textit{abstraction} and \textit{content-sensitivity} in each attention head. For each problem in our original evaluation, we first define three minimally-different versions of the same problem differing only in the \textit{abstraction} tested. Importantly, this abstraction should matter for the ground-truth correct response, such as changing \textit{The painting is North of Ali} to \textit{The painting is South of Ali}. We then define three minimally-different versions of the same problem differing in the arbitrary \textit{content} tested, such as changing \textit{The painting is North of Ali} to \textit{The mountain is North of Ali}. Importantly, these content substitutions should not matter for the outcome of the problem.

These contrasts (six for each problem, totaling 2730 contrast pairs, see Appendix \ref{app:patching-stim} for examples) allow us to broadly characterize each neuron in the model along the axes of content sensitivity and abstraction sensitivity. For a given model neuron processing some base prompt from the original evaluation, we patch in (replace) that neuron's activations with its activations when processing an \textit{abstract} or \textit{content} contrast. We then measure how changing this single neuron's activations affects the LLM's outputs as a whole. If patching in abstract prompt activations for this neuron leads to greater changes in model outputs, that neuron is more \textit{abstraction-sensitive}, and vice-versa. This approach allows us to determine the function of each neuron (on our evaluation) insofar as it relates to actual model output behavior.

\begin{figure}[H]
    \centering
    \includegraphics[width=1\linewidth]{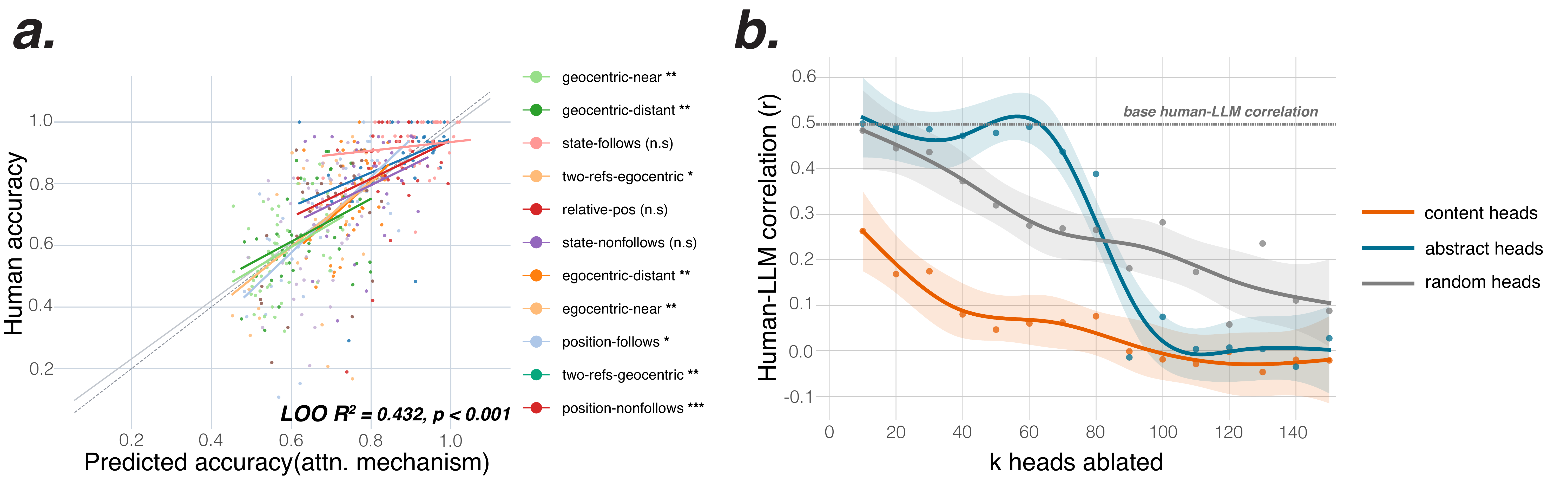}
    \caption{\textbf{Comparing the roles of content-sensitive and abstract attention head representations in LLM-human alignment.} \textbf{a. Predictions of human behavior from content-sensitive heads (k=10)}. Actual (y axis) versus predictions (x axis) of human-behavior on held-out stimuli from a ridge regression using the activations of the top 10 content-sensitive attention heads. We include a line-of-best-fit for within category predictions. Asterisks indicate significance of linear model fit within category. \textbf{b. Effects of content-sensitive and abstraction-sensitive head ablations on human alignment.} We compare the effects of ablating the top-$k$ content-sensitive heads (orange) and abstraction-sensitive heads (blue). We also include ablations of a random top-$k$ heads (gray) selected from the complement of all other attention heads (averaged over 5 random head selections for each $k$).
    }
    \label{fig:predict-acc-mech}
\end{figure}

\paragraph{Content-sensitive attention heads predict human accuracy.} After characterizing the abstraction-sensitivity and content-sensitivity of all neurons in \texttt{gemma-4-12b}\footnote{Distributions of content-sensitive and abstraction-sensitive attention heads show that content sensitive heads tend to appear in earlier model layers relative to abstraction sensitive heads (Appendix \ref{app:head-distributions}). This is commensurate with prior work demonstrating that more ``abstract'' operations tend to be computed in later model layers \citep{tenney2019bert}}, asked whether human responses were better predicted by content- or abstraction-sensitive neurons (attention heads). Our method is illustrated in Figure \ref{fig:proc-ablations}b (i): We first selected the top-$k$ most content-sensitive attention heads as well as the top-$k$ most abstraction-sensitive heads. We then tested whether the held-out predictions of these attention heads (derived from linear probes of head activations) improve our ability to predict human responses beyond the problem category and problem framing. We also included simpler \textit{distributional baselines} \citep{trott2026large}--mean trigram frequency of the input and small model (\texttt{GPT-2}) surprisal--to test whether the internal mechanisms of interest were predictive above simpler measures of distributional semantics. Section \ref{sec:head-predictions} includes the full implementation details of model fitting and our predictors. Comparisons of held-out $R^2$ for all values of $k$ can be found in Section \ref{app:partial-corrs}.

Comparisons of model fit for \textit{content-sensitive} and \textit{abstraction-sensitive} heads
revealed that while the activations of both head types improved linear model fit to human
behavior, the predictions of \textit{content-sensitive} heads were better than those of
\textit{abstraction-sensitive} heads at all values of $k$ heads ($F$s$(k, 392\text{--}430) =
11.2\text{--}72.8$ vs $9.1\text{--}50.9$, all $p < 0.001$), when including fixed effects for
category and prompt type ($F$s$(k, 381\text{--}419) = 6.1\text{--}20.8$ vs $4.2\text{--}9.5$,
all $p \leq 0.011$), and when accounting for distributional baselines and the logit outputs of
\texttt{gemma-4-12b} ($F$s$(k, 378\text{--}416) = 1.6\text{--}3.8$ vs $0.5\text{--}2.1$,
$p = 0.002\text{--}0.818$) \ref{fig:predict-acc-mech}b). Figure \ref{fig:predict-acc-mech}a
shows held out predictions of human behavior for the top $k=10$ content-sensitive attention
heads. These head activations captured graded within-category variance in human responses,
with significant within-category fits in 8 of the 11 categories (See Appendix \ref{app:per-category-breakdown}). Permutation tests also reveal
that content-sensitive heads offer better predictions relative to abstraction-sensitive heads
in isolation ($\Delta R^2 = 0.064\text{--}0.108$ for $k \geq 6$, $p < 0.05$ at 11 of 12 values
of $k$) and when including fixed effects and distributional baselines as controls
($\Delta R^2 = 0.003\text{--}0.051$, $p < 0.05$ for all $k > 4$). This pattern of results is also consistent for the other 14 models evaluated: Appendix \ref{app:partial-corrs} includes F-tests and permutations aggregated across all remaining LLMs.

% \textit{Content-sensitive} heads significantly improved fit to human behavior while \textit{abstraction-sensitive} heads did not. In fact, adding these as predictors lowered model-human alignment when accounting for the model's own outputs and distributional baselines. Figure \ref{}a shows held out predictions for \textit{content-sensitive} heads at $k=10$: the activations of these heads capture within-category variance in human responses which is not represented in abstraction-sensitive heads. These findings hold for all values of $k$ heads selected and all models evaluated (see Appendix \ref{}).

\paragraph{Content-sensitive attention heads align LLMs with human responses.}

The analyses above show that content-sensitive representations in attention heads improve predictions of human accuracy. But do these content-sensitive representations produce human-like behavior in LLMs? To find out we progressively ablated (i.e., set the output activations of the attention heads to zero) the top-$k$ content and abstraction sensitive heads from the model, measuring how removing either content-sensitive or abstract representations modulated the alignment between the model and human behavior.

In line with our correlational analyses, ablating content-sensitive heads had a significantly greater impact on decreasing the alignment of model outputs with human behavior. Figure \ref{sec:internal-mech-predict}b shows correlations between human accuracy and content-sensitive head ablations (orange) and abstraction-sensitive head ablations (blue) for all values of $k$. As a control, we also randomly ablate $k$ random heads and repeated this ablation across 5 folds for every value of $k$. Our results show that while content-sensitive head ablations immediately deteriorate human-model alignment at small values of $k$, the same is not true of abstraction-sensitive ablations: indeed, ablating abstract heads has less of an effect on model alignment than ablating random heads. A spline regression (appropriate for modeling the non-linear effects of attention head ablation) confirms that the relationship between
\textit{content-sensitive} head ablations and human accuracy is indeed greater than that of
abstraction-sensitive or random heads (abstraction vs.\ content: $b = 0.20$, $SE = 0.015$,
$t = 13.85$; content vs.\ random: $b = -0.21$, $SE = 0.015$, $t = -14.69$; condition $\times$
spline interaction: $\chi^2(10) = 138.94$, $p < 0.001$; Full results for the spline model fit can be found in Appendix \ref{app:head-ablation-general}). This effect holds when aggregating across the top 15 models in our evaluation (abstraction vs.\ content: $b = 0.138$, $SE = 0.012$, $t = 11.73$; content vs.\ random: $b = -0.124$, $SE = 0.009$, $t = -13.58$; condition $\times$ spline interaction: $\chi^2(10) = 144.59$, $p < 0.001$), and is also true of model accuracy (abstraction vs.\ content: $b = 0.271$, $SE = 0.131$, $t = 2.07$; content vs.\ random: $b = -0.258$, $SE = 0.101$, $t = -2.54$; condition $\times$ spline interaction: $\chi^2(10) = 20.83$, $p = 0.022$, see Appendix \ref{app:head-ablation-general}).
\begin{figure}[H]
    \centering
    \includegraphics[width=0.90\linewidth]{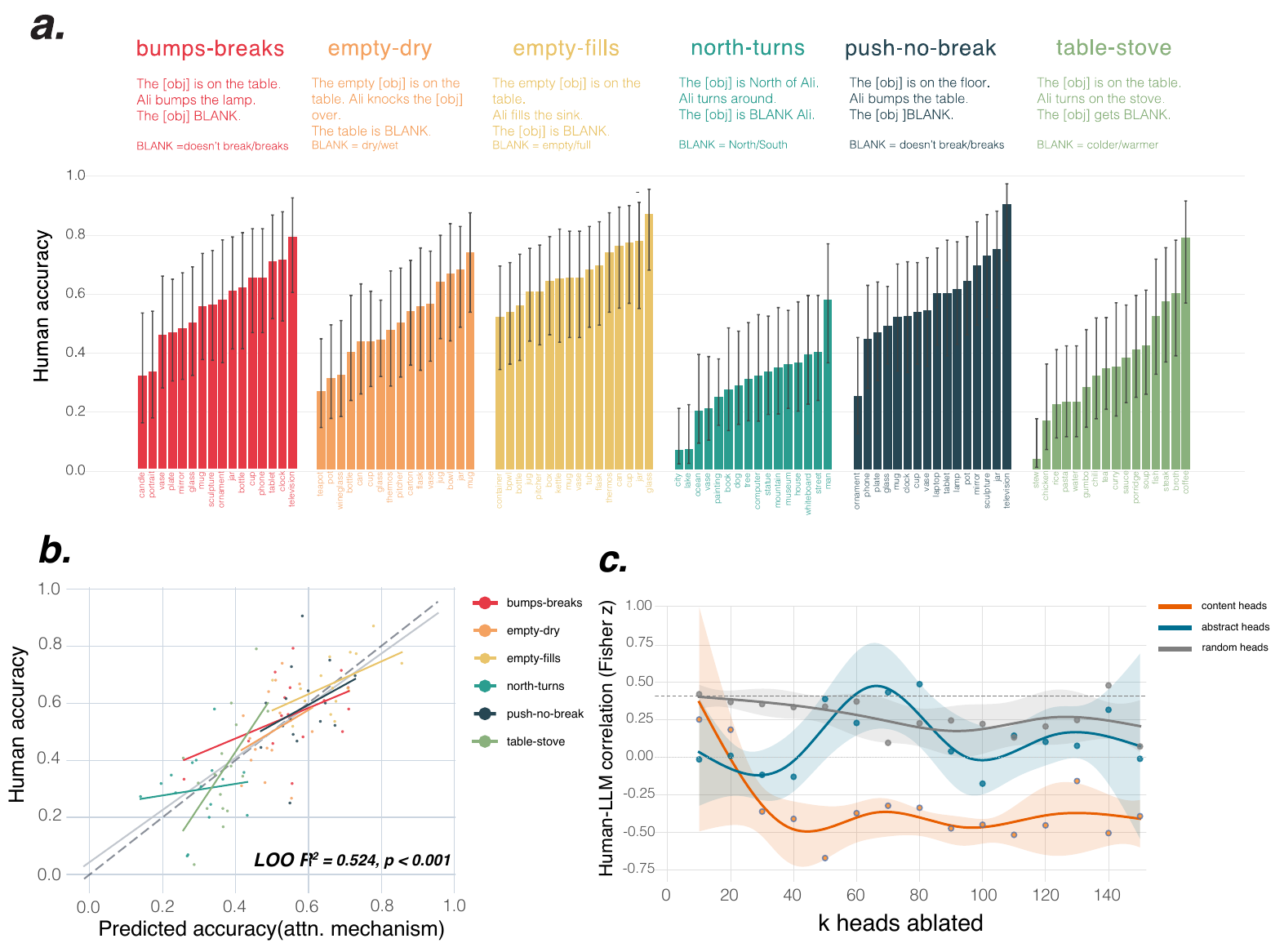}
    \caption{\textbf{Content-sensitive attention heads are responsible for similar content effects in LLM and human responses.} \textbf{a. Human accuracies for six sets of minimally different \textit{scenarios}.} Labels above show scenario templates, words on x axis ticks indicate words substituted in the \textit{[obj]} template slot for each scenario. \textbf{b. Content-sensitive heads predict human behavior.} Predictions from the activations of content-sensitive heads explain a majority of variance in human responses. Predictions are included from a linear model with fixed effects for prompt scenario and held-out predictions from the top 3 content sensitive heads. \textbf{c. Effects of content-sensitive and abstraction-sensitive head ablations on human alignment.} We compare the effects of ablating the top-$k$ content-sensitive heads (orange) and abstraction-sensitive heads (blue). We also include ablations of a random top-$k$ heads selected from the complement of all other attention heads (averaged over 5 random head selections for each $k$). The dotted line indicates the base correlation between humans and \texttt{gemma-4-12b}.}

    % Content effects in human reasoning are explained by causally important attention heads.} \textbf{a.}Human accuracies for six sets of minimally different \textit{scenarios}. Labels above show scenario templates, words on x axis ticks indicate words substituted in the \textit{[obj]} template slot for each scenario. \textbf{b.} ground truth human accuracy versus held out model predictions from a regression with top-$k$ attention heads ($k=7$ selected using BIC) and a fixed effect for scenario. Perfect fit ($y= \hat{y}$) is indicated by the dotted gray line. \textbf{c.} Comparisons of $R^2$ distributions from linear model permutation tests, for models fit \textit{only} from attention head features. We fit 1000 regressions predicting human accuracy with features randomly sampled from a standard normal distribution (blue) and non-causally important attention heads (gray);  p-values indicate the probability of the top-$k$ linear model $R^2$ (red line) being greater than the distribution of $R^2$ values from the random normal or random attention head distributions.
    
    %GL: Same confusion as in the previous version of the figure... What catches the reviewers' eye is the p=.086 and the reader then tries to relate this to the text... 
    
    \label{fig:pattern-match-overview}
\end{figure}

\subsection{Content-sensitive heads explain content-specific variance in human behavior}
\label{sec:content-effects-min-diff}
The previous analyses show that broadly human-like patterns of responses in LLMs are causally dependent on attention heads that are sensitive to ostensibly ``arbitrary'' prompt content. However, the heterogeneity of our stimuli-- testing a range of world knowledge grounded in many different real-world scenarios--makes it difficult to determine if changes to prompt content \textit{alone} are sufficient for manipulating the responses of people and language models in predictable (and content-dependent) ways. To address this limitation we created a new set of minimal-difference scenarios variants. For example, the scenario template \textit{The \texttt{[obj]} is on the table. Ali bumps the lamp. The [obj] BLANK.} is instantiated with \textit{phone} and with \textit{jar}. As in the previous evaluation, the particular content variation should bear no importance on the problem outcome. 

Despite content-variations in these stimuli being irrelevant to the prompt outcome, human
accuracy varies dramatically depending on this content (see Figure
\ref{fig:pattern-match-overview}a). For example, people correctly recognize (with 79\% accuracy)
that for the prompt \textit{The coffee is on the table. Ali turns on the stove. The coffee gets
\texttt{BLANK}}(warmer/colder), the coffee should get colder as it is on the table, not the stove.
However, less than 1\% of our participants recognize that for \textit{The \textbf{stew} is on
the table}, the stew should also become colder. We observe similar content sensitivity in our
set of language models: \texttt{gemma-4-12b} shows even greater content sensitivity than people
and covaries with human responses to a moderate degree ($r(94) = 0.40$, $p < 0.001$, these
results are consistent across models, see Appendix \ref{app:correlation-pred-min-diff}). That
this content-sensitive variation in human behavior is predictable by a range of different LLMs
and is not fully explained by simpler distributional baselines and the model's own outputs ($b = 0.026$, $SE = 0.003$,
$t(41) = 8.31$, $p < 0.001$, controlling for mean trigram frequency and \texttt{GPT-2}
surprisal) indicates that people, rather than guessing, show a similarly graded sensitivity to
prompt content consistent with language models.

We replicated our mechanistic analyses for these minimal-difference stimuli \footnote{We used
minimal-contrasts designed for these stimuli to ensure attention heads captured the desired
behaviors (see Section \ref{app:abstract-content-min-diff-stim}). The identified neurons appear
in similar layers to our original evaluation, see Appendix
\ref{app:min-diff-head-distributions}.}. We again found that the parts of the LLM that were
most sensitive to content differences explained most of the variance when predicting human
responses. Activations of content-sensitive heads explain significant within-scenario variance,
above and beyond that predicted by abstraction-sensitive heads (content: $F$s$(k, 70\text{--}88)
= 2.84\text{--}14.43$, all $p \leq 0.001$; abstraction: $F$s $= 0.53\text{--}2.38$, significant
at 6 of 10 values of $k$; $\Delta R^2 = 0.105\text{--}0.181$). Disrupting
these content-sensitive neurons led to greater misalignment with human data than disrupting
neurons implicated in representing more abstract information (abstraction vs.\ content:
$b = 0.452$, $SE = 0.083$, $t = 5.48$; content vs.\ random: $b = -0.608$, $SE = 0.064$,
$t = -9.51$), see Figures \ref{fig:pattern-match-overview}b and \ref{fig:pattern-match-overview}c. In sum, the findings from our original evaluation
stimuli generalize to stimuli that differ \textit{only} in the arbitrary content present in the
prompt, again suggesting that these arbitrary differences elicit similar effects in people and
LLMs at the level of behavior and mechanism.

\section{Discussion}
When large language models are probed to reason about simple causal relations in everyday scenarios, they show sensitivity to seemingly inconsequential changes to the prompt. For example, the models reason correctly about the facing direction of someone who is North of Chicago after turning around, but reason incorrectly about the same situation when someone is described as being North of a painting. As a stronger test of such content sensitivity, we created versions of the same prompt that differed only in the role filler, e.g., whether an empty jar (or teapot) that is resting on a table makes the table wet when the object is knocked over. The model is much more likely to respond correctly for teapot than jar.

It is tempting to interpret such content sensitivity (and the errors is sometimes leads to) as evidence that a system is ``not really reasoning'' \citep{arkoudasGPT4CantReason2023} or understanding \citep{mitchell_debate_2023} and is rather \textit{merely} pattern-matching \citep{chen2026llms, shojaeeIllusionThinkingUnderstanding2025}. However, these same fail states are also clearly evident observed in human reasoning (see also \citet{lampinen2024language} \citet{loo2026llms} \citet{eisape2024systematic}). Just like LLMs our participants were more accurate when asked to use cardinal directions North/South/East/West when considering more distant referents (cities, mountains, and statues), and more accurate for egocentric (front, back, left, right) when reasoning about more proximal references (paintings, people, and rooms). People are twice as likely to think that a knocked over empty teapot made the table wet than an empty jar, and so did LLMs. Like LLMs, people tended to expect causal outcomes where there ought to be none, readily inferring that bumping a table with a glass on it might cause water in the glass to spill, but failing to notice when an agent is described as bumping a different object (e.g., a lamp instead of a table) which should have no effect on the glass. Applying the same standard to such content sensitivity in human reasoning would lead to the dubious conclusion that humans don't ``really reason'' either. 

On their own, human errors of the sort we report here may be brushed aside as performance errors (e.g., stemming from momentary lapses of attention) which tell us little about underlying human competence. If people's errors were sporadic or unpredictable we too might be tempted to brush them aside. Instead, they are consistent (as revealed with retests) and are \textit{highly} predictable as revealed by their close alignment with the behavior produced by LLMs. Our investigations of the mechanisms behind the ability of LLMs to predict responses at both category, item, and minimal-difference level showed that the model-human alignment was strongly driven by neurons that encode the content of the prompt rather than its abstract form. Intervening on these neurons drastically reduced model-human alignment. 

Taken together, our results strongly point to the models performing a type of pattern-matching and that this pattern matching is what enables the model to respond in human-like ways. These results do not imply the absence of bona fide world models in LLMs. Indeed we were able to locate neurons that were sensitive to the abstract structure of the prompted scenarios. But ablating these neurons had much less effect on the ability of the model to respond in human-like ways. Our results also do not imply that people are unable to represent abstract causal world models. After all, the very same system that thinks a person North of a painting is South of the painting when they turn around is also capable of stepping back and appreciating that they have made an error. Rather, the results show that causal reasoning people and LLMs use when predicting causes and outcomes of everyday scenarios are strongly content sensitive, exhibiting all the signatures of a pattern-matching process rather than the robust content-independent generalization expected of an abstract world model.

\subsection{Why pattern matching?}
Pattern matching may seem like a highly non-optimal way to reason. The usefulness of pattern matching becomes clearer when we consider the nature of the problems people (and LLMs trained on human-produced data) face \citep{margolisPatternsThinkingCognition1990, mercier_enigma_2017}. Although an abstract world model is beneficial when the problem space is well-defined, most tasks people face are far more open-ended with complex and multi-directional causal influences that are conditional on specific content. In such a task environment it is rarely advantageous to abstract away from the specifics in favor of an abstract model. From this perspective, it is not a coincidence that tasks that \textit{do} require such abstraction, e.g., syllogistic reasoning, deductive inference, and arithmetic are sources of difficulty and are associated with frequent generalization failures \citep[e.g.,][]{lampinen2024language, wason1968reasoning, banerjeeChildrensArithmeticSkills2025}. Succeeding on these tasks frequently involves leaning on external resources such as Euler diagrams and formally learned algorithms executed with the aid of pen-and-paper executed algorithms \cite[e.g.][]{bauerHowDiagramsCan1993}. While such training and reliance on external media helps with ``formal'' problem solving, we are not aware of evidence showing generalization to the far more common everyday reasoning problems of the sort tested here. In sum, we believe that people and LLMs rely on pattern matching precisely because this mechanism is a better match to the ecology of everyday causal reasoning problems that people tend to face.

The best-performing LLMs in our analyses had superhuman performance. They were notably less aligned with human behavior because they avoided mistakes that people often made. Because we cannot inspect the internal mechanisms of these models we cannot make any strong conclusions about whether these models do in fact implement bona fide abstract world models. Considering that the performance gains of these models do not appear to come at the expense of engaging in highly effective pattern matching \cite{lupyanUnreasonableEffectivenessPattern2026}, a parsimonious conclusion is that these models are not better because their representations are more abstract, but because they have learned (possibly through more sophisticated post-training) task-settings that allow the model to form abstractions that are more suitable to the task at hand.

\clearpage
\section{Supplemental Methods}

{\renewcommand\thefootnote{}\footnotetext{Code for all experiments is available at https://github.com/zstud04/llm-shared-mechanisms}}

\subsection{Behavioral measures}
\subsubsection{Evaluating human responses} We collected data from 162 participants from Mechanical Turk (pre-screened by Cloud Research) excluding 20 participants who failed more than 1 of 5 basic attention checks or who averaged less than one second per response. We note that participant responses were collected in January 2025 prior to the popularization of agentic AI platforms which have since made online participant data collection more difficult \citep{kay2025you}. Each participant completed 55 prompts (excluding attention checks) yielding $8910$ individual prompt observations.

\paragraph{Evaluation details.} On each trial participants were shown a prompt (with a $BLANK$ included in the prompt string indicating where the answer should be filled in) and were instructed to press $SPACE$ to reveal the two answer options at which point they should select the the  ``\textit{the most likely word/phrase to replace ``BLANK''}''. An example trial is included in Appendix section \ref{app:human-eval-details}). In addition to the final choice, we collected two response time measures: \textit{read-rt} defined as the time spent reading the prompt before pressing $SPACE$, and \textit{choice-rt}, defined as the time spent selecting the answer option.

\subsubsection{Testing the consistency of human judgments}
\paragraph{Re-test procedure and participant justifications.} To assess the robustness of our results, we replicated our behavioral evaluation in an additional sample of participants ($n=49$). To determine whether subjects are randomly guessing, we presented every question a subject gets wrong a second time (randomly inserted later in trial order). After the second stimulus presentation, participants were asked to write a brief response justifying their answer choice.

\subsubsection{Collecting language model responses}
\label{app3:lm-responses}
We tested 25 LLMs (21 open source and 4 proprietary closed models) on our world knowledge evaluation, across a range of parameter sizes, training modalities and post-training regimens (see Appendix \ref{app:model_battery} for properties of all models evaluated). Model responses were queried individually for each stimulus using the following prompt:

\begin{quote}
\texttt{What is the most likely word/phrase to replace ''BLANK''? Respond ONLY with the word/phrase.}

\texttt{\{prompt\}}

\texttt{Option A: \{a\}}\\
\texttt{Option B: \{b\}}

\texttt{BLANK =}
\end{quote}

For each prompt, we randomly shuffled the ordering of the correct and incorrect options into the response slots $(a, b)$. For instruction-tuned models, we applied the appropriate instruction or chat template as defined by the model provider. 

\subsubsection{Measures of language model accuracy}
\label{sec:h-model-inf}

\paragraph{Open-weights models} For open weights models, we obtained the raw output logits for the correct and incorrect answer tokens, and defined model accuracy as the difference between the correct and incorrect logits, $\mathrm{logit}(r^{+}) - \mathrm{logit}(r^{-})$. This value will be positive when the correct answer choice is more likely relative to the incorrect choice. We measured logit values directly rather than probing to avoid potentially confounding task demands associated with interpreting the question template \citep{hu2023prompting}, and to obtain a more fine-grained measure of model confidence.

\paragraph{Proprietary models} For proprietary models where full activations are not available, We used a binary measure of accuracy determined by whether the correct response option appears in the first five token outputs when generating at zero temperature.

% \subsubsection{Measuring language model reasoning effort} Inspired by \citet{de2025cost}, we evaluated the reasoning model \texttt{DeepSeek-R1} and obtain the string length of reasoning traces for all model generations (allowing for indefinite length in reasoning traces). We then log-transformed all string-length counts,
% $\ell_i = \log\!\left(1 + n_i\right)$,
% where $n_i$ denotes the number of characters in the $i$-th reasoning trace. Analyses of the relationship between reasoning effort and human accuracy can be seen in Appendix \ref{app:reasoning_traces_predict}.

\subsection{Characterizing content-sensitivity and abstraction-sensitivity of attention heads}
\label{sec:attention-interp}
\paragraph{Contrast definitions} We first determined whether each attention head was more sensitive to substitutions in \emph{abstract} prompt information (i.e., changes to the underlying causal relation being tested, which alter the correct response) or in prompt \emph{content} that has no bearing on the correct response. Formally, for each base prompt and its correct and incorrect responses $(p_b, r_b^{+}, r_b^{-})$, we constructed sets of abstraction contrasts $\mathcal{A}_b = \{(p_a^{(i)}, r_a^{(i)+}, r_a^{(i)-})\}_{i=1}^{3}$ and content contrasts $\mathcal{C}_b = \{(p_c^{(i)}, r_c^{(i)+}, r_c^{(i)-})\}_{i=1}^{3}$ such that $r_a^{(i)+} \neq r_b^{+}$ and $r_c^{(i)+} = r_b^{+}$ for all $i$ (correct response is the same for all content contrasts, different for all abstraction contrasts).

\paragraph{Patching procedure} Consider one set of base, abstraction contrast, and content contrast prompts. For each attention head $k$ in some transformer layer $\ell$ in $H$, we extracted the activations of that attention head in the forward pass of the abstraction contrast and content contrast prompts, $h_a$ and $h_c$ respectively. Then, we patched $h_a$ and $h_c$ into the activations of head $k$ while processing the base prompt, $h_b$. The forward pass of $H$ on the base prompt was the same except for the minimal-difference patch at $h_b$. The measured change in the logit difference between the correct and incorrect responses, $\mathrm{LD} = \mathrm{logit}(r_b^{+}) - \mathrm{logit}(r_b^{-})$, under the forward pass with the abstraction patch, $\Delta_a = \mathrm{LD}(p_b) - \mathrm{LD}(p_b \mid h_b \leftarrow h_a)$, and under the content patch, $\Delta_c = \mathrm{LD}(p_b) - \mathrm{LD}(p_b \mid h_b \leftarrow h_c)$, gives us an abstraction-score and a content-score for each attention head. Roughly, this corresponds to how the attention head modulates model responses, conditional on differences in the \emph{abstraction} or \emph{content} in the prompt. We repeated this procedure for all attention heads in the model and all base prompts, averaging $\Delta_a$ and $\Delta_c$ over the three contrasts of each type and across base prompts to obtain each head's final abstraction and content scores, $S^{\mathrm{abs}}_{\ell,k}$ and $S^{\mathrm{con}}_{\ell,k}$.

\paragraph{Linear approximations of attention head contributions} As described in \citet{elhage2021mathematical}, the final transformer hidden state can be defined as a linear combination of each component across layers:
\begin{equation}
    x_L = x_0 + \sum_{\ell=1}^{L} \Big( \sum_{k} h_{\ell,k} + m_\ell \Big),
\end{equation}
where $x_0$ is the embedding and $h_{\ell,k}$ and $m_\ell$ are the outputs of head $k$ and the MLP in layer $\ell$. This allows us to directly estimate the contribution of a given component by projecting its output onto the logit-difference direction using a method known as direct logit attribution (DLA):
\begin{equation}
    \mathrm{DLA}_{\ell,k} = h_{\ell,k} \cdot \big( W_U[:, r_b^{+}] - W_U[:, r_b^{-}] \big).
\end{equation}
DLA captures only direct effects: thus, while it does not give an exact value for the causal contribution of a given attention head, empirically we note a very strong degree of alignment between actual ablations and DLA-estimated ablations (see Appendix \ref{app:dla-effect}). We use DLA to estimate the causal effect of content and abstraction contrasts as defined above for each attention head. This heuristic allows us to feasibly run our mechanistic analyses on an expanded set of models.

\paragraph{Selecting top-$k$ content sensitive and abstraction sensitive attention heads} We defined the top-$k$ content sensitive attention heads as the $k$ heads with the greatest content score relative to abstraction score, $S^{\mathrm{con}}_{\ell,k} - S^{\mathrm{abs}}_{\ell,k}$, and vice versa for the top-$k$ abstraction sensitive heads, $S^{\mathrm{abs}}_{\ell,k} - S^{\mathrm{con}}_{\ell,k}$.

\subsection{Predicting human responses from attention head activations}
\label{sec:head-predictions}
Here we describe our method for predicting human accuracy from linear probes fit to attention head outputs. 

\paragraph{Measuring top-$k$ head activations} For some selection of top-$k$ attention heads, we extracted each attention head's output activations at the last token in each prompt, $h_j(p_i)$ for head $j \in \{1, \dots, k\}$ and prompt $p_i$. We then fit linear probes (with L1 regularization) predicting human responses $y_i$ using a leave-one-out procedure: each prediction generated from model activations is for a prompt \emph{not} used to fit model weights,
\begin{equation}
    \hat{y}_{ij} = \mathbf{w}_j^{(-i)\top} h_j(p_i) + b_j^{(-i)},
\end{equation}
where $\mathbf{w}_j^{(-i)}$ and $b_j^{(-i)}$ were fit on all prompts except $p_i$. This yields a matrix $\hat{Y} \in \mathbb{R}^{n \times k}$ (where $n$ is the number of prompts in the evaluation) containing the held-out predictions of $k$ attention head output vectors for each prompt.

\paragraph{Distributional baselines} We included two distributional baselines to assess whether attention heads are uniquely predictive of human responses above and beyond generic measures of word co-occurrence and small-model surprisal. Specifically, we included a measure of mean \emph{trigram frequency}, defined as the average trigram probability over a sliding window of each prompt string $w_1, \dots, w_T$,
\begin{equation}
    \bar{P}_{\mathrm{tri}}(p) = \frac{1}{T-2} \sum_{t=3}^{T} P(w_t \mid w_{t-2}, w_{t-1}),
\end{equation}
and \emph{GPT-2 surprisal}, $-\sum_{t} \log P_{\text{GPT-2}}(w_t \mid w_{<t})$, which has empirically been shown to be an effective operationalization of human reading time \citep{oh2026model}

\paragraph{Fitting linear models} We combined the linear-probe derived predictions of model attention heads with the distributional baselines (described above) and fixed effects for category and prompt type. We also asked whether these mechanistic predictors explained additional variance above and beyond model logit outputs (for the respective LLM), including these outputs as an additional control in a separate set of regression results. For all linear models, we assess the variance explained on \textit{held-out} predictions using a leave-one-out procedure for all stimuli.

\subsection{Estimating the causal effects of content and abstraction sensitive heads}

In addition to our correlational analyses of attention head function, we perform a series of \textit{ablations} to determine the causal effect of content-sensitive and abstraction-sensitive attention heads.

\paragraph{Head ablation procedure} For each $k \in \{0, 1, \dots, |\mathcal{H}|/2\}$, where $|\mathcal{H}|$ is the total number of attention heads, we selected the top-$k$ content-sensitive heads $\mathcal{H}^{\mathrm{con}}_k$ and the top-$k$ abstraction-sensitive heads $\mathcal{H}^{\mathrm{abs}}_k$. Then, for each subset of model components, we ablated the activations of those attention heads by setting them to zero. These ablations were applied at \emph{all} sequence positions, $h^{(t)}_{\ell,j} \leftarrow \mathbf{0}$ for all $t \in \{1, \dots, T\}$. We repeated this procedure for all values of $k$.

\paragraph{Random $k$-selection control} We also compared the effect of content-sensitive and abstraction-sensitive head ablations relative to ablations of random heads in the model. Specifically, at every value of $k \in \{0, 1, \dots, |\mathcal{H}|/2\}$, we randomly select $k$ attention heads from the model disjoint from the top-$k$ content-sensitive and abstraction-sensitive sets. We repeated this selection five times, selecting different random heads with replacement. We then average across these random selections at each value of $k$ to estimate the causal effect of random head ablations with 95\% confidence intervals.

\paragraph{Measuring effects on human alignment and model accuracy} The above procedure yields three ablations (content-sensitive heads, abstraction-sensitive heads, and random heads), resulting in three different perturbed distributions of model output logits for the correct and incorrect response options. Let $\mathrm{LD}^{(c,k)}_i$ denote the logit difference on prompt $p_i$ under ablation condition $c \in \{\mathrm{con}, \mathrm{abs}, \mathrm{rand}\}$ with $k$ heads ablated. For each perturbed distribution, we measured the correlation of logit outputs with human responses, $r\big(\mathrm{LD}^{(c,k)}, y\big)$, and the absolute accuracy of model responses, $\frac{1}{n}\sum_{i=1}^{n} \mathrm{LD}^{(c,k)}_i$.

\newpage

\appendix

\section{Behavioral evaluation details}

\subsection{Causal reasoning stimuli}
\label{eval-overview}

We introduced a novel causal reasoning evaluation to evaluate the ability of humans and LLMs to flexibly deploy and reason about world knowledge across different scenarios and prompt formats. In this section we specify four broad domains of world knowledge (based on prior studies and debates in cognitive science) and define 11 \textit{categories} for our evaluation based on these domains. To test world knowledge in humans and LLMs, we evaluated reasoning across various everyday domains that ostensibly do not require specialized knowledge. Following \citet{ivanova2025elements}, we adopted a modular stimulus format that allows for flexibly recombining starting context and target states within different domains. Each stimulus prompt $S$ consists of a context state $C$ and a target state $T$, which together define a causal relation $R(C \rightarrow T)$. In the \textit{context-completion condition}, a word in the context state $C$ is omitted and must be inferred given a binary choice between two alternatives $(a, b)$. In the \textit{target-completion condition}, a component of the target state $T$ is omitted and must be inferred from two options given the context $C$.

\paragraph{Stimuli domains.} We designed our evaluation stimuli based on four broadly defined domains of world knowledge:

\textit{Changes in the positions of objects.} The ability to reason about changes in the positions of objects--as they are thrown, pushed, pulled, lifted, and dropped-- is frequently interrogated in tests of perception \citep{spelke1990principles} intuitive physics \citep{xu2021bayesian, mccloskey1983intuitive}, and causality \citep{kruschke2019perception}. Moreover, evaluations of object positions and movements have often served as a proving ground for symbolic \citep{newell1980physical, tenenbaum2002theory} and connectionist \citep{ullman2000high, kipf2018neural} models of human inference. We included three categories evaluating the ability to reason about changes in object positions: \textit{action-follows}, where an object changes in position following some event, \textit{action-nonfollows}, where some action takes place but a change in object position is not entailed, and \textit{relative-position}, where we test knowledge about positions between objects. Evaluating scenarios where a change in position both is and isn't entailed allows us to probe for additional sources of potential error, such as a bias towards inferring causal relations where they do not exist.

\textit{Changes in the states of objects.} Much of early human reasoning is evaluated on the ability to reason about stability in the states and properties of objects \citep{baillargeon1995physical}, and transitions between states \citep{gopnik2007causal}. Early knowledge of object states is also at the center of debates about the existence of conceptual primitives \citep{carey2000origin} and causality \citep{gopnik2012reconstructing}. We defined a category to test scenarios where an action results in a change of state in an object (\textit{state-follows}) and a category where the state of an object should be invariant to changes in state (\textit{state-nonfollows}), allowing us to probe intuitions about scenarios where state should be preserved and changed.

\textit{Egocentric relations.} Egocentric (from one's own perspective) experience has long been posited as a critical source of grounded knowledge for humans \citep{piaget2013construction, tversky2005functional} and AI systems \citep{vong2024grounded, varela2017embodied}. Our evaluation includes three categories to test the ability of humans and LLMs to reason in egocentric reference frames about nearby objects (\textit{egocentric-near}) faraway cities and landmarks (\textit{egocentric-distant}), and the egocentric perspectives of others (\textit{two-refs-egocentric}).

\textit{Geocentric relations.} Geocentric (North, South, East, West) relations, while rooted in the physical world, are learned by humans through cultural transmission and language \citep{levinson2003space, majid2004can}. Geocentric relations are not innate to infants \citep{mishra2013culture} and are frequently misunderstood by adults \citep{hegarty2005individual}. To evaluate how these relations are represented in humans and LLMs, we created three categories: \textit{geocentric-near} (geocentric relations for nearby objects), \textit{geocentric-distant} (geocentric relations for distant objects), and \textit{two-refs-geocentric} (geocentric relations between multiple people and places). 

\clearpage

\subsection{Model battery}
 We collect behavioral judgments from a suite of 42 open-source and 4 proprietary models. The behavioral measures obtained for each LLM (as described in Section \ref{sec:h-model-inf}) depended on whether the model was open source (with accessible logits) or only available through API calls. The table below lists the properties of all models.

\label{app:model_battery}
\label{app:model_battery}

\begin{table}[H]
\centering
\small
\begin{tabular}{lcccc}
\hline
\textbf{Model} & \textbf{Model family} & \textbf{Parameters (Billions)} & \textbf{Modality} & \textbf{Post-training} \\
\hline
Gemma-2-2B                 & Gemma    & 2.6  & Text-only  & Base \\
Gemma-2-2B-IT              & Gemma    & 2.6  & Text-only  & IT   \\
Gemma-2-9B                 & Gemma    & 9.2  & Text-only  & Base \\
Gemma-2-9B-IT              & Gemma    & 9.2  & Text-only  & IT   \\
Gemma-2-27B                & Gemma    & 27.2 & Text-only  & Base \\
Gemma-2-27B-IT             & Gemma    & 27.2 & Text-only  & IT   \\
Gemma-3-4B                 & Gemma    & 4.3  & Multimodal & Base \\
Gemma-3-4B-IT              & Gemma    & 4.3  & Multimodal & IT   \\
Gemma-3-12B                & Gemma    & 12.2 & Multimodal & Base \\
Gemma-3-12B-IT             & Gemma    & 12.2 & Multimodal & IT   \\
Gemma-3-27B                & Gemma    & 27.4 & Multimodal & Base \\
Gemma-3-27B-IT             & Gemma    & 27.4 & Multimodal & IT   \\
Gemma-4-12B-IT             & Gemma    & 12.0 & Multimodal & IT   \\
Gemma-4-26B-A4B-IT         & Gemma    & 27.0 & Multimodal & IT   \\
Gemma-4-31B-IT             & Gemma    & 33.0 & Multimodal & IT   \\
GPT-2 XL                   & OpenAI   & 1.5  & Text-only  & Base \\
GPT-OSS-20B                & OpenAI   & 21.0 & Text-only  & IT   \\
LLaMA-2-7B                 & Llama    & 6.7  & Text-only  & Base \\
LLaMA-2-7B-Chat            & Llama    & 6.7  & Text-only  & IT   \\
LLaMA-2-13B                & Llama    & 13.0 & Text-only  & Base \\
LLaMA-2-13B-Chat           & Llama    & 13.0 & Text-only  & IT   \\
LLaMA-3.1-8B               & Llama    & 8.0  & Text-only  & Base \\
LLaMA-3.1-8B-Instruct      & Llama    & 8.0  & Text-only  & IT   \\
T\"ulu-3-8B-SFT            & Llama    & 8.0  & Text-only  & SFT  \\
OLMo-2-1124-7B-SFT         & OLMo     & 7.3  & Text-only  & SFT  \\
OLMo-2-1124-7B-DPO         & OLMo     & 7.3  & Text-only  & DPO  \\
OLMo-2-1124-7B-Instruct    & OLMo     & 7.3  & Text-only  & IT   \\
OLMo-2-1124-13B-SFT        & OLMo     & 13.7 & Text-only  & SFT  \\
OLMo-2-1124-13B-Instruct   & OLMo     & 13.7 & Text-only  & IT   \\
OLMo-2-0325-32B-Instruct   & OLMo     & 32.0 & Text-only  & IT   \\
OLMo-3-7B-Instruct         & OLMo     & 7.3  & Text-only  & IT   \\
OLMo-3.1-32B-Instruct      & OLMo     & 32.0 & Text-only  & IT   \\
Phi-3-mini-128k-Instruct   & Phi      & 3.8  & Text-only  & IT   \\
Phi-3.5-mini-Instruct      & Phi      & 3.8  & Text-only  & IT   \\
Phi-3-medium-4k-Instruct   & Phi      & 14.0 & Text-only  & IT   \\
Phi-3-medium-128k-Instruct & Phi      & 14.0 & Text-only  & IT   \\
Qwen-2-7B-Instruct         & Qwen     & 7.6  & Text-only  & IT   \\
Qwen-2.5-7B-Instruct       & Qwen     & 7.6  & Text-only  & IT   \\
Qwen-2.5-14B-Instruct      & Qwen     & 14.8 & Text-only  & IT   \\
Qwen-2.5-32B-Instruct      & Qwen     & 32.8 & Text-only  & IT   \\
Qwen-3-4B-Instruct (FP8)   & Qwen     & 4.0  & Text-only  & IT   \\
Qwen-3-30B-A3B-Instruct    & Qwen     & 30.5 & Text-only  & IT   \\
\hline
GPT-4.1                    & OpenAI   & --   & Multimodal & IT   \\
GPT-5.2                    & OpenAI   & --   & Multimodal & IT   \\
Gemini-2.0-Flash           & Gemini   & --   & Multimodal & IT   \\
DeepSeek-R1                & DeepSeek & --   & Text-only  & IT   \\
\hline
\end{tabular}
\caption{Model families and architectural metadata for the 46 models evaluated: 42 open-weight models (top) and 4 closed-weight API models (bottom).}
\end{table}

\newpage

\subsection{Measures of human and model responses}

\label{sec:behavioral-eval}
We defined 11 different categories testing the ability to reason about causal relations in broad domains of knowledge as pecified in section \ref{eval-overview}. Definitions of these domains are as follows (category summaries also available in Figure \ref{fig:stim_overview}b):

\begin{table*}[ht]
\centering
\small
\begin{tabular}{p{3.2cm} p{4.5cm} p{7cm}}
\hline
\textbf{Category} & \textbf{Definition} & \textbf{Example} \\
\hline
state-follows & Action changes object state & The cup is empty. Ali pours water in the cup. The cup is [wet/dry]. \\
\hline
state-nonfollows & Action doesn't change object state & The empty glass is on the table. Ali fills the sink. The glass is [dry/wet]. \\
\hline
position-follows & Action changes object position & The seesaw is balanced. Ali tilts the right side of the seesaw down. The left side goes [up/down]. \\
\hline
position-nonfollows & Action doesn't change object position & Ali is behind the statue. Ali turns around. Ali is [behind/in front of] the statue. \\
\hline
geocentric-distant & Geocentric relations for distant references & Chicago is North of Ali. Ali turns around. Chicago is [North of/South of] Ali. \\
\hline
geocentric-near & Geocentric relations for nearby references & The box is East of Ali. Ali turns around. The box is [West of/East of] Ali. \\
\hline
egocentric-distant & Egocentric relations for distant references & Ali is facing away from Chicago. Ali turns 180 degrees. Chicago is [in front of/behind] Ali. \\
\hline
egocentric-near & Egocentric relations for nearby references & The painting is left of Ali. Ali turns left. The painting is [in front of/behind] Ali. \\
\hline
relative-position & Relative positions between objects & The ball is next to the box. Ali puts the ball inside the box. The ball is [smaller/larger than] the box. \\
\hline
two-refs-egocentric & Egocentric relations with multiple reference points & Ali and Mark are facing each other. Mark turns around. Ali is facing [towards/away from] Mark. \\
\hline
two-refs-geocentric & Geocentric relations with multiple reference points & Milwaukee is South of Mark. Mark goes West. Mark is [Northeast/Southeast] of Milwaukee. \\
\hline
\end{tabular}
\caption{Overview of the 11 evaluation categories with definitions and example stimuli. Correct answers are shown first in brackets.}
\label{tab:categories}
\end{table*}

\paragraph{Probing knowledge of plausible causes and effects} A common criteria of world models is the ability to infer the likely outcome of an event, as well as to reason about plausible causes of an event given some observed outcome \citep{pearl2018book}. To evaluate participants' ability to model both plausible causes and effects, we presented each stimulus in two different \textit{framings}: we created a set of \textit{target-evaluation} prompts where subjects must infer the target outcome of a scenario, and a set of \textit{context-evaluation} prompts where the context that caused some observed outcome must be inferred. An example pair of target and context stimuli (for the same scenario) can be seen in Figure \ref{fig:stim_overview}a. The full set of 11 categories (with 2 framings per category) yields 433 total prompts for our evaluation.

\subsection{Human evaluation}
\label{app:human-eval-details}

\subsubsection{Evaluation format}
On each trial, participants saw the prompt without the answer options. They were asked to read the prompt and pres the spacebar to reveal the options. An example trial sequence is included below.

\begin{figure}[h]

    \centering
    \includegraphics[width=1\linewidth]{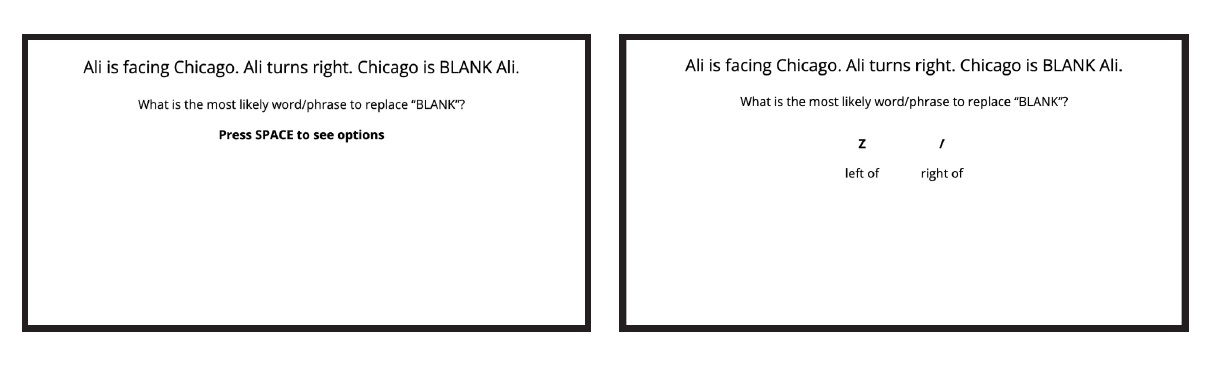}
    \caption{\textbf{Human evaluation trial sequence.} Participants are first presented with the prompt and asked to press $SPACE$ to reveal answer options. They are then instructed to press $z$ or $/$ corresponding to an answer choice (order randomized).}
    
    \label{fig:stim_overview2}

\end{figure}

\subsubsection{Human re-test evaluation}
\label{app:retest}
To determine whether low accuracy in human causal reasoning is consistent or the result of random noise, we deploy our evaluation for an additional set of subjects ($n=70$). In addition, we re-evaluate the same subjects on questions they initially got wrong, presenting the same (initially incorrect) problems at a second point in the evaluation. Subjects were then prompted to provide a free response justification of their choice following the second stimulus presentation. Figure \ref{fig:h_retest_cat} shows subject-level reliability by category. We note moderate consistency in the relationship between a subject's first and second responses, although this varies by category.

\begin{figure}

    \centering
    \includegraphics[width=0.40\linewidth]{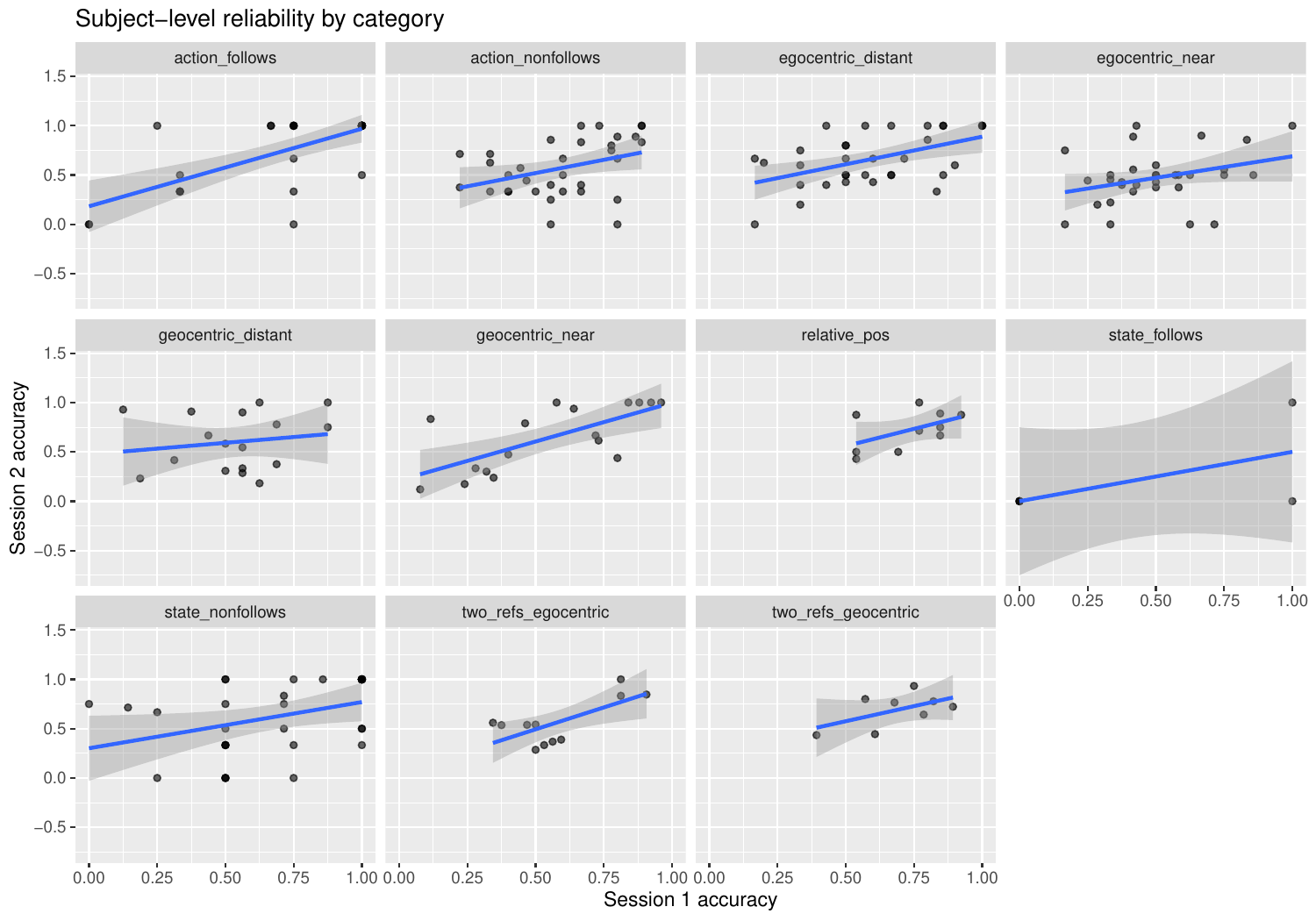}
    \caption{Consistency between first and second responses for a given subject, separated by category. X-axis indicates scores for each participant on the initial round (session) of prompts, y-axis indicates participant scores on the second presentation of prompts. }

    \label{fig:h_retest_cat}

\end{figure}

\subsubsection{Distribution of participant accuracies}
\label{app:dip-test}
To rule out the possibility that the close alignment between human and LLM responses was driven by some human participants actually being LLM-powered bots, we examined the distribution of item-wise correlations between each participant and two LLMs available at time of testing (gemma-3-27b-it, and GPT 4.1). A bimodal distribution would indicate that some participants may in fact be bots. A Hartigan's dip test revealed that the distributions were not significantly different from a unimodal normal distribution (see results in the main text).

\begin{figure}

    \centering
    \includegraphics[width=0.60\linewidth]{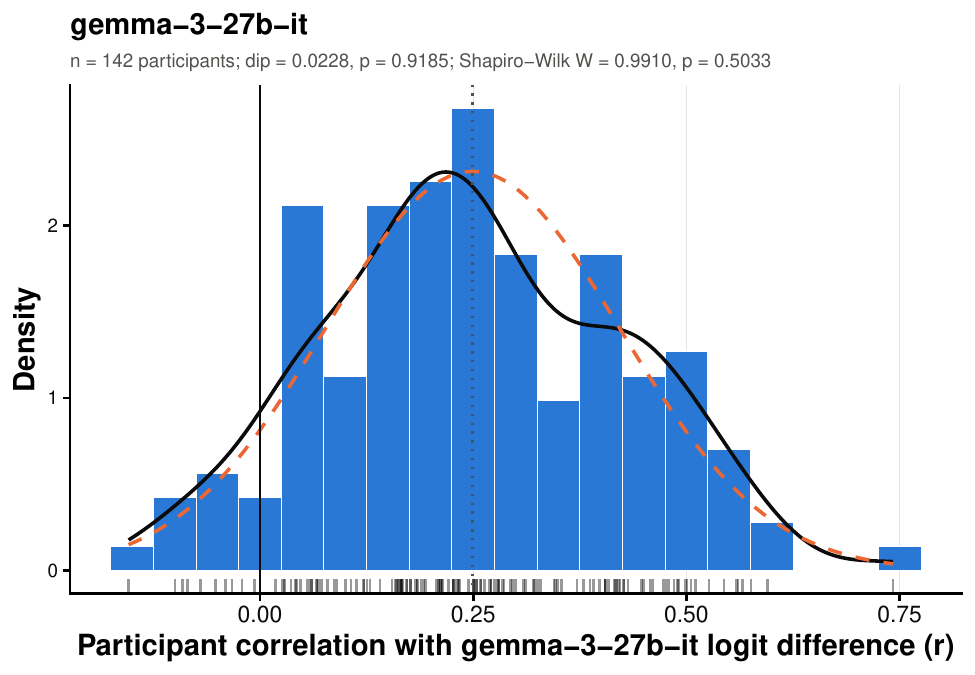}
    \caption{Distribution of participant correlations with \texttt{gemma-3-27b} on our general evaluation (A frontier LLM at the time of data collection). The dotted red line indicates a normal distribution with the mean and standard deviation of our participant distribution, for comparison.}

    \label{fig:h_retest_cat}

\end{figure}

\begin{figure}

    \centering
    \includegraphics[width=0.60\linewidth]{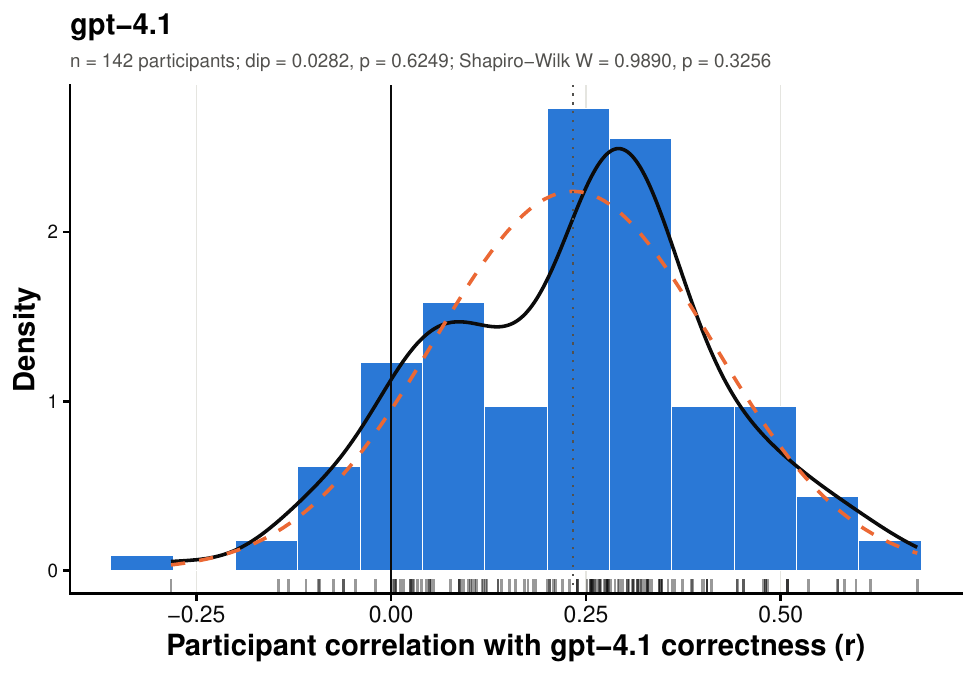}
    \caption{Distribution of participant correlations with \texttt{GPT-4.1} on our general evaluation (A frontier LLM at the time of data collection). The dotted red line indicates a normal distribution with the mean and standard deviation of our participant distribution, for comparison.}

    \label{fig:h_retest_cat}

\end{figure}

\pagebreak
\subsection{LLM behavioral results}

\subsubsection{Accuracy measures}
\begin{figure}

    \centering
    \includegraphics[width=0.4\linewidth]{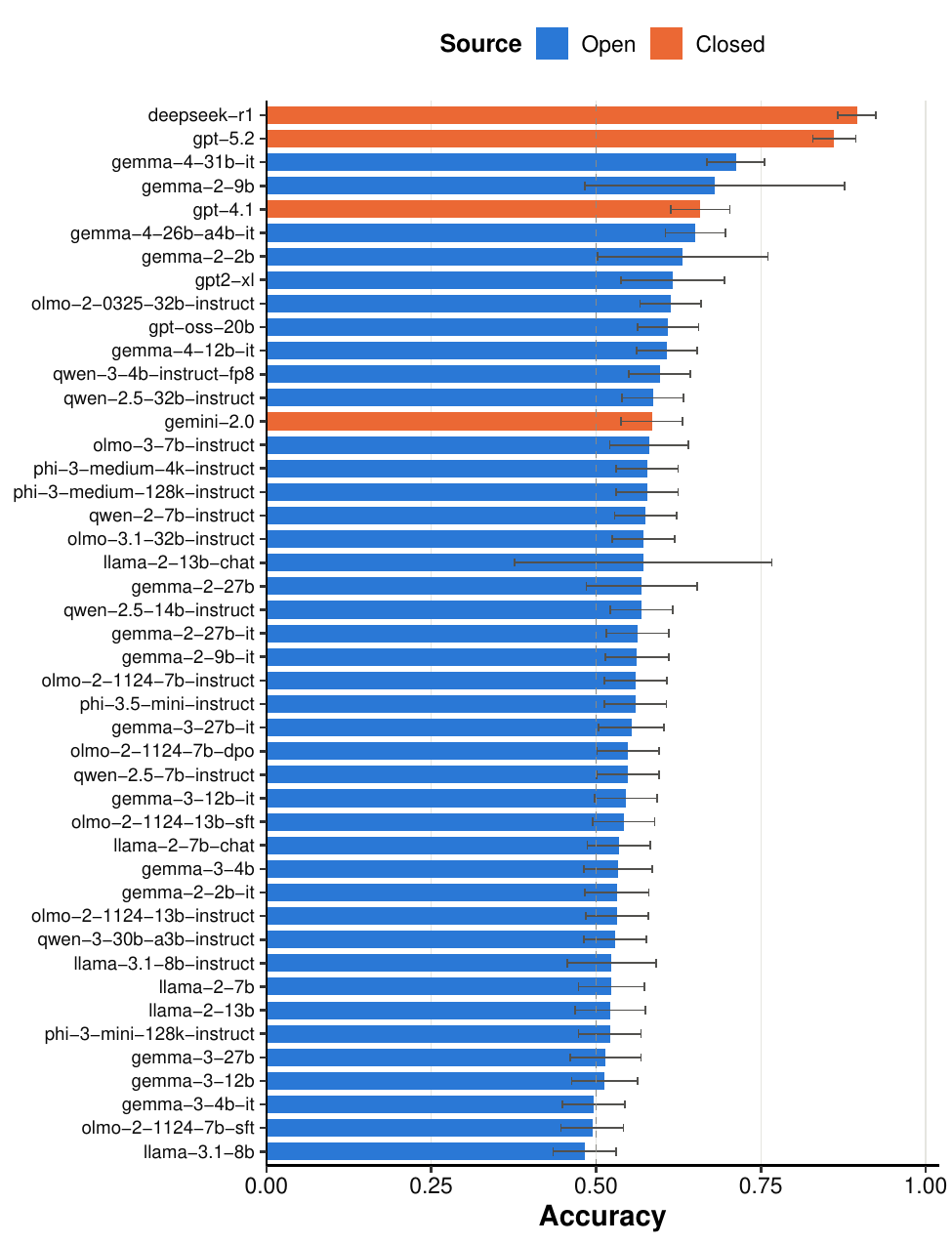}
    \caption{Mean accuracy of open and closed source models as given by accuracy measured from model outputs. Error bars indicate 95\% confidence intervals.}

    \label{fig:model_acc}

\end{figure}

\begin{figure}

    \centering
    \includegraphics[width=0.4\linewidth]{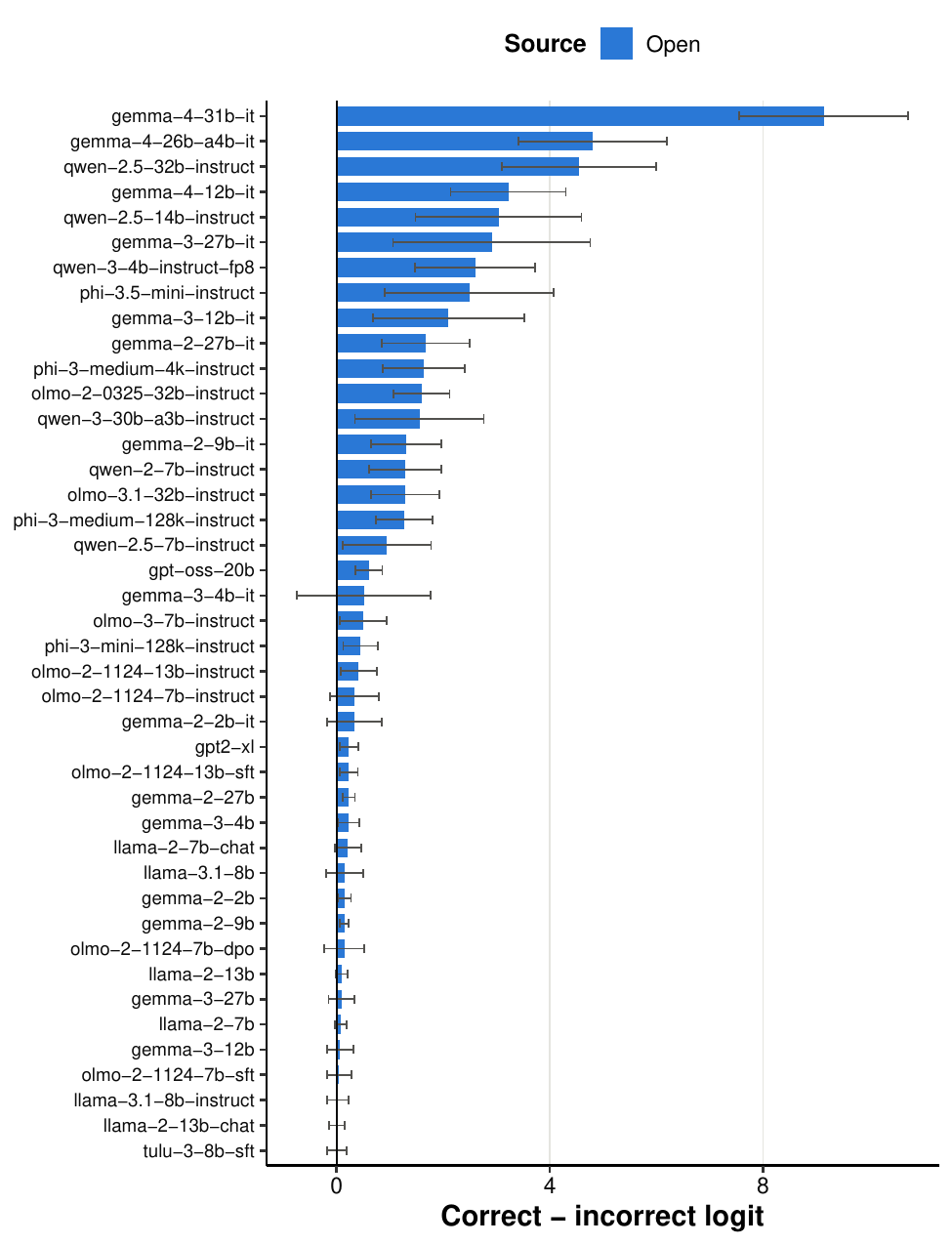}
    \caption{Mean accuracy of open and closed source models as given by the difference of correct and incorrect logit outputs. Error bars indicate 95\% confidence intervals.}

    \label{fig:model_logits}

\end{figure}

We obtain \textit{logit difference} measures for all open-source models, and binary response measures (whether the model completion contains the correct answer token) for all closed and open-source models. Figure \ref{fig:model_acc} and Figure \ref{fig:model_logits} show mean response accuracies and logit difference metric values by model, respectively.

\begin{figure}

    \centering
    \includegraphics[width=0.4\linewidth]{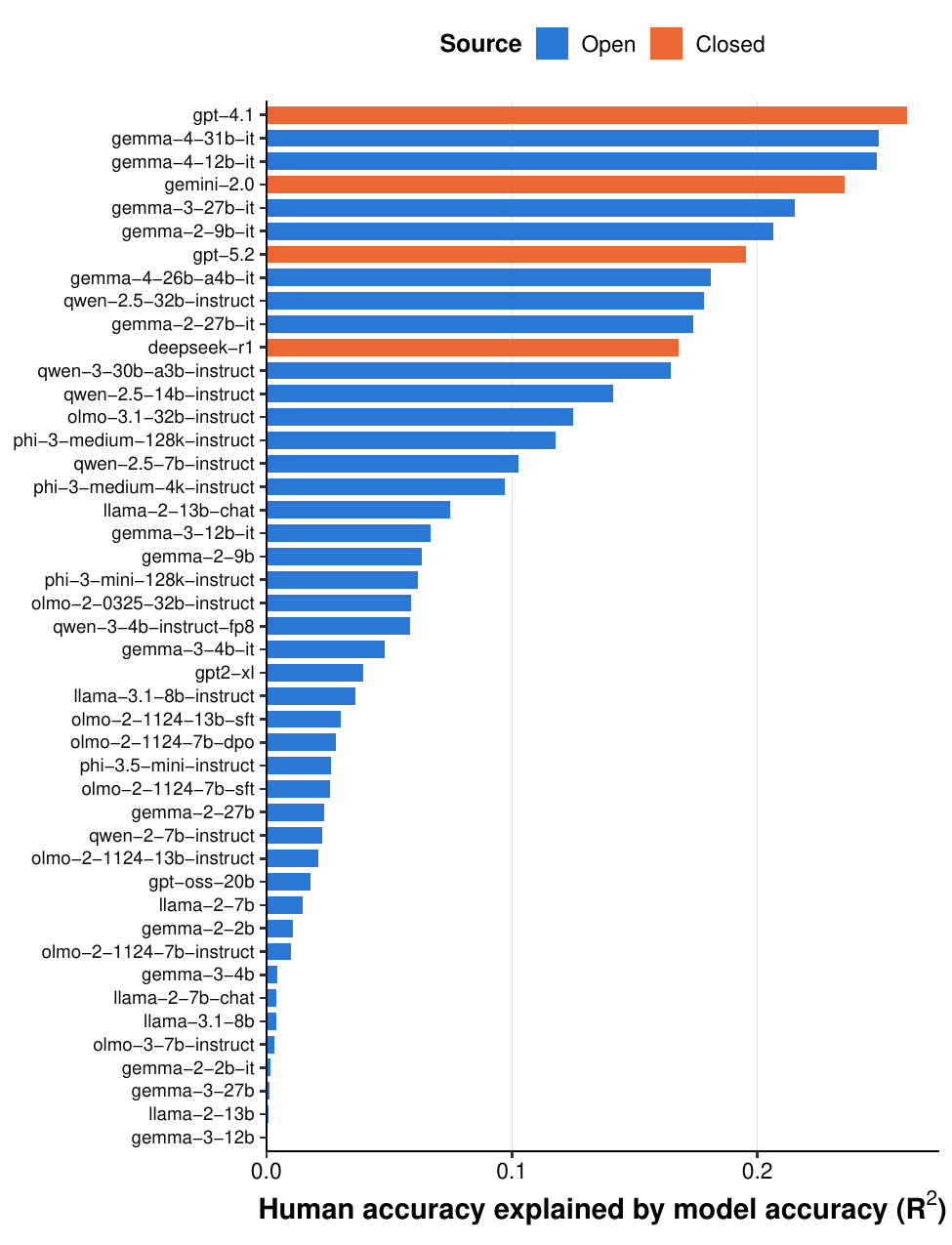}
    \caption{$R^2$ values for linear models predicting human accuracy from LLM generation accuracy.}

    \label{fig:model_acc_predict}

\end{figure}

\begin{figure}

    \centering
    \includegraphics[width=0.4\linewidth]{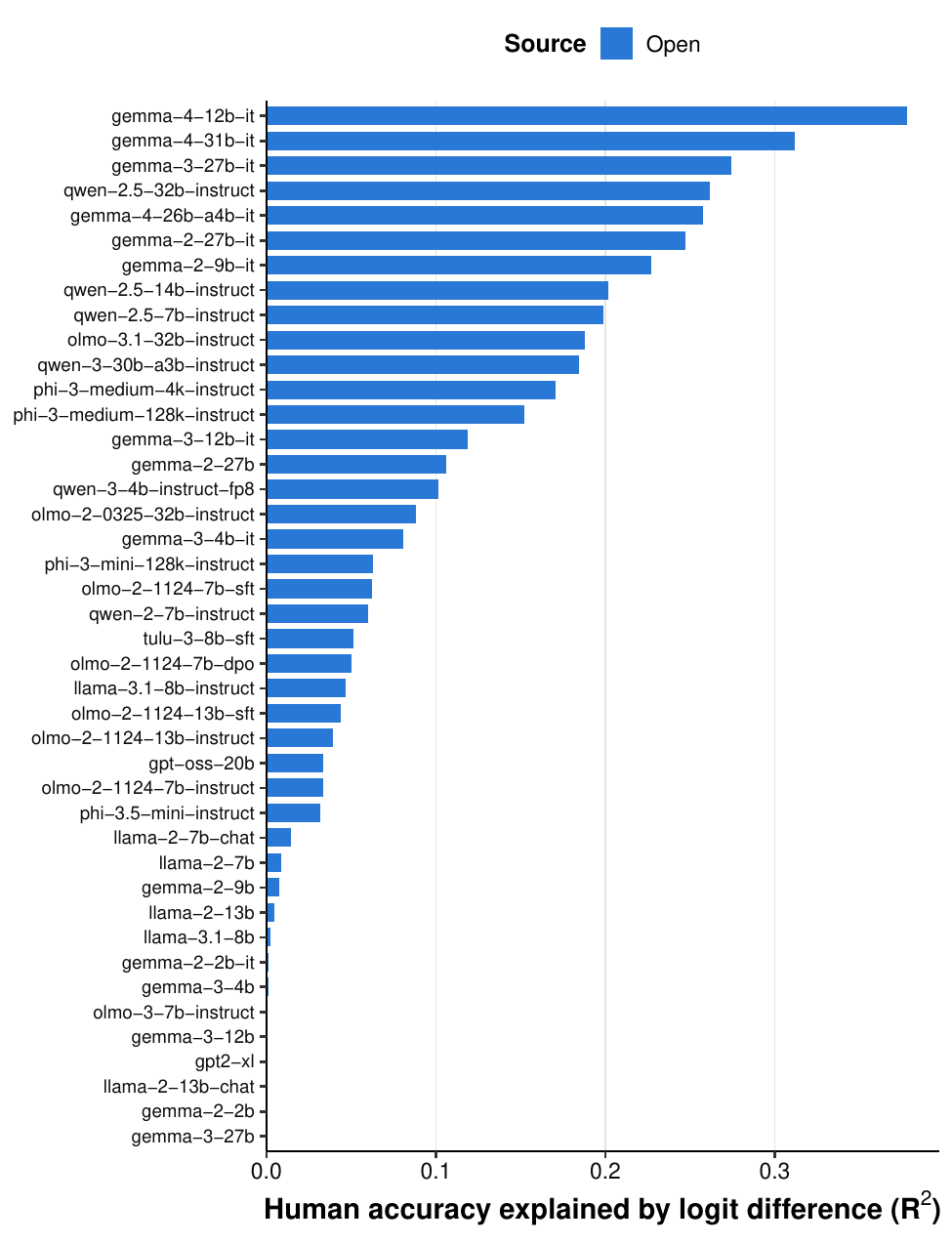}
    \caption{$R^2$ values for linear models predicting human accuracy from LLM logit differences.}

    \label{fig:model_logits_predict}

\end{figure}

\pagebreak

\subsection{Category level human-LLM alignment}
\label{app:cat-alignment}
\paragraph{Open-source model category alignment} We include additional category-level visualizations of human and model accuracy alignment, for the five open-source models most aligned with human accuracy (excluding \texttt{gemma-4-12b-it}, which is included in the main text). All visualizations here use \textit{logit difference} as a measure of accuracy.

\begin{figure}[H]
    \centering
    \includegraphics[width=0.6\linewidth]{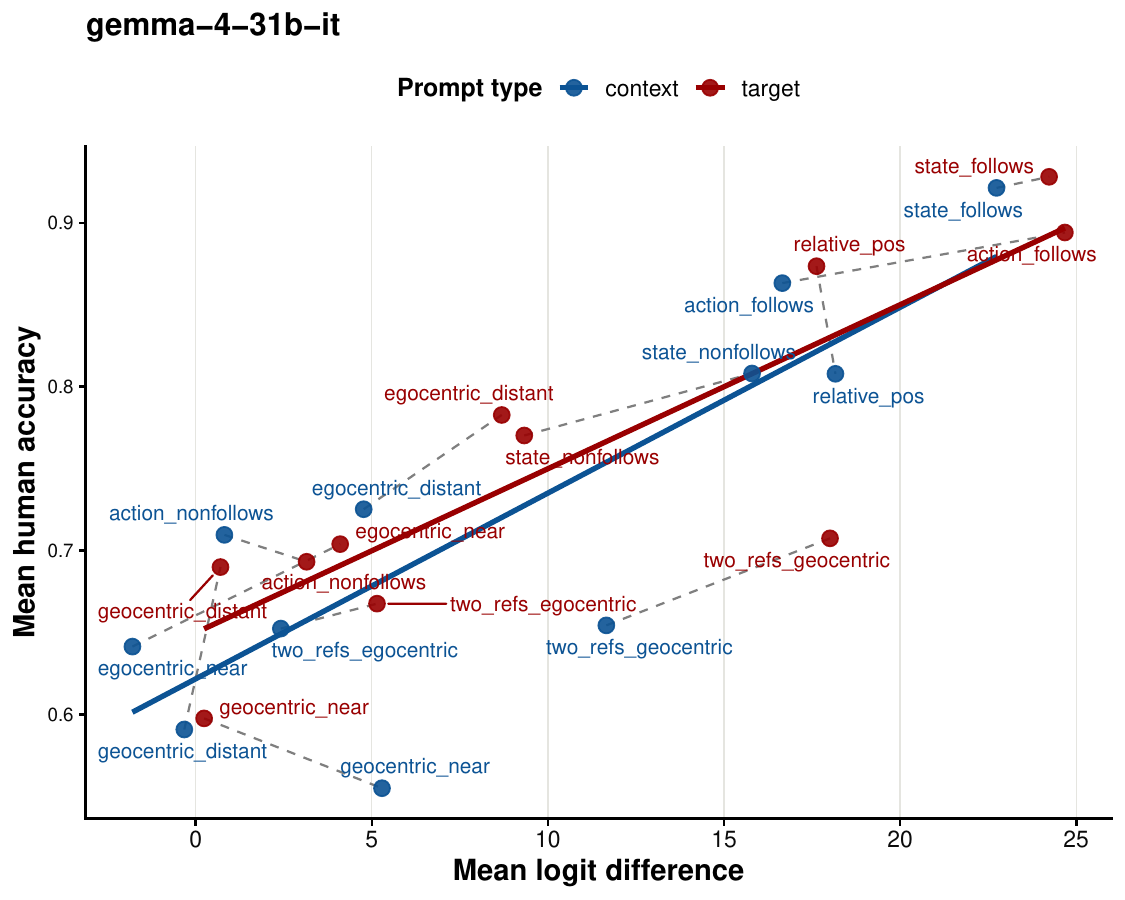}
    \caption{\textbf{Human accuracy versus logit difference for Gemma 4 31B IT} Category-level averages showing the relationship between model logit difference and human accuracy, with dashed lines connecting context and target prompt types within each category.}
\end{figure}

\begin{figure}[H]
    \centering
    \includegraphics[width=0.6\linewidth]{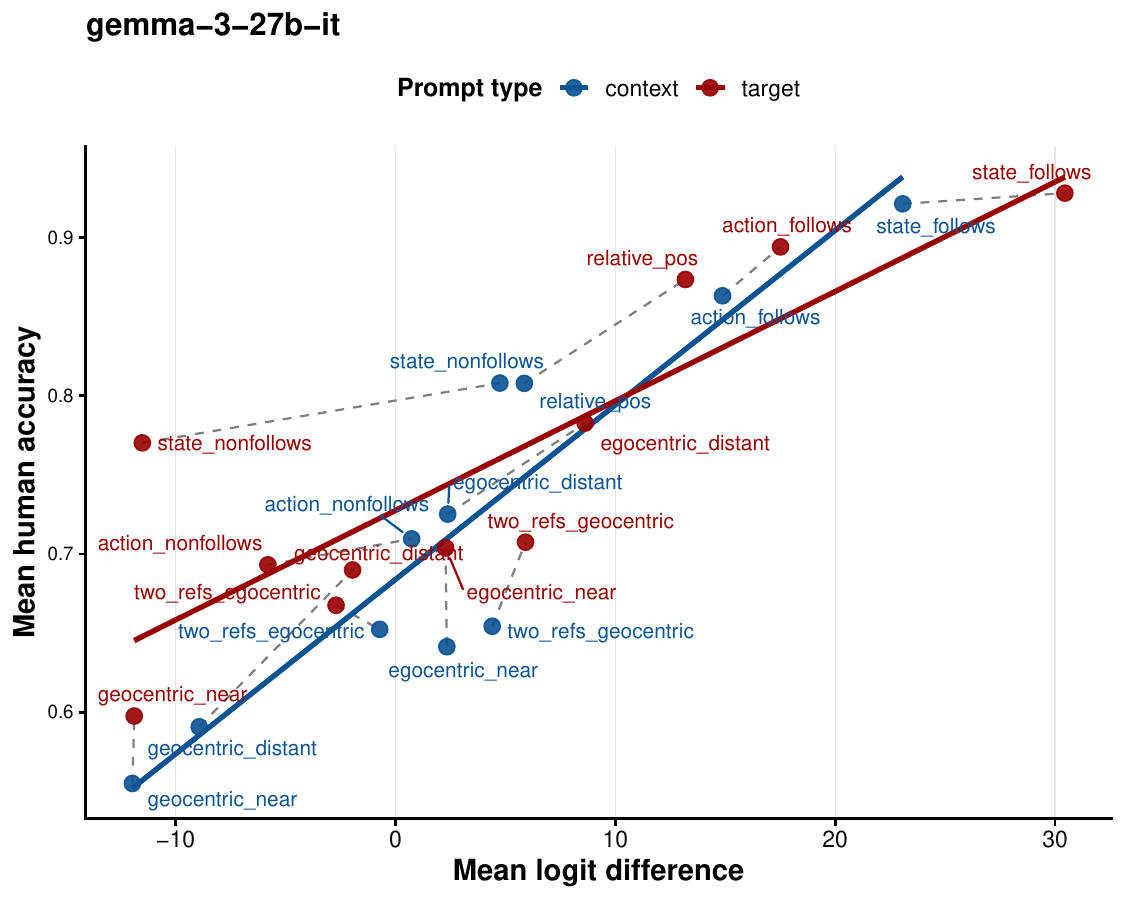}
    \caption{\textbf{Human accuracy versus logit difference for Gemma 3 27B IT} Category-level averages showing the relationship between model logit difference and human accuracy, with dashed lines connecting context and target prompt types within each category.}
\end{figure}

\begin{figure}[H]
    \centering
    \includegraphics[width=0.6\linewidth]{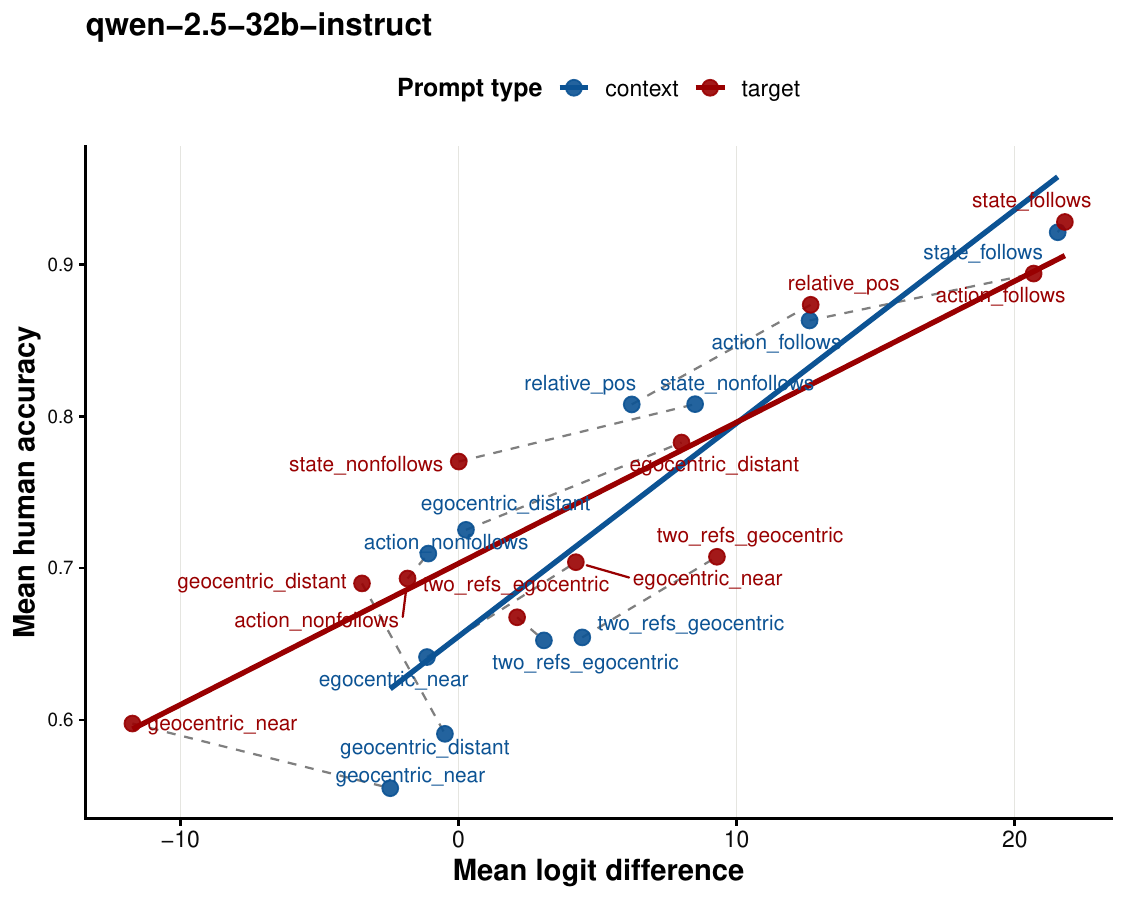}
    \caption{\textbf{Human accuracy versus logit difference for Qwen 2.5 32B Instruct} Category-level averages showing the relationship between model logit difference and human accuracy, with dashed lines connecting context and target prompt types within each category.}
\end{figure}

\begin{figure}[H]
    \centering
    \includegraphics[width=0.6\linewidth]{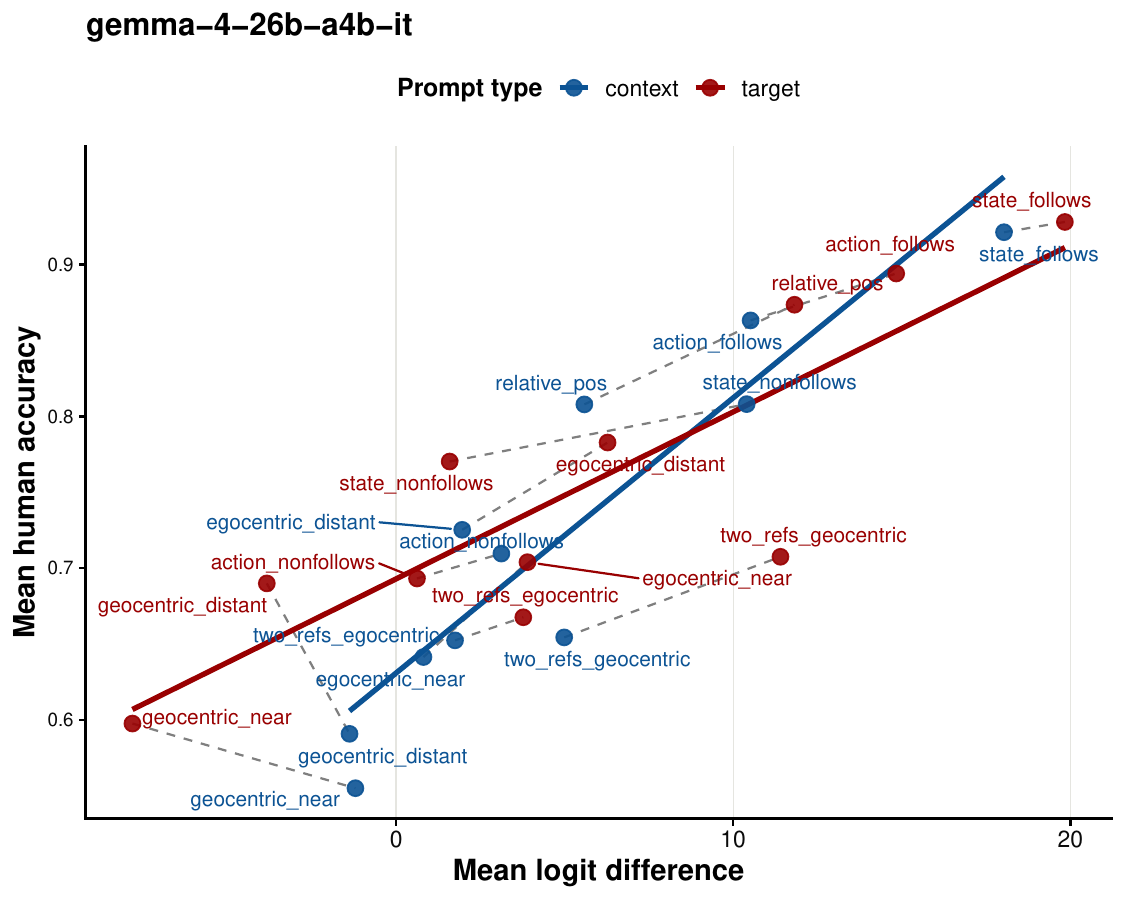}
    \caption{\textbf{Human accuracy versus logit difference for Gemma 4 26B-A4B IT} Category-level averages showing the relationship between model logit difference and human accuracy, with dashed lines connecting context and target prompt types within each category.}
\end{figure}

\begin{figure}[H]
    \centering
    \includegraphics[width=0.6\linewidth]{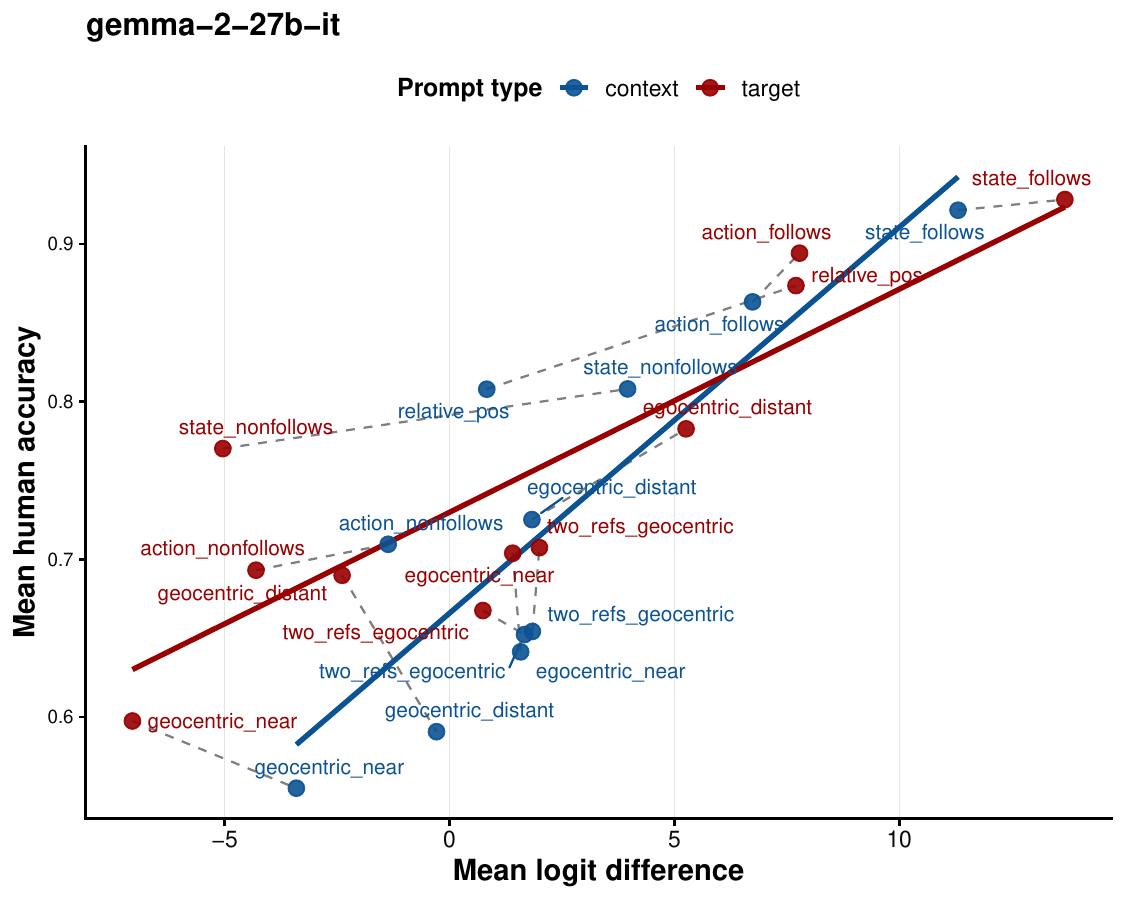}
    \caption{\textbf{Human accuracy versus logit difference for Gemma 2 27B IT} Category-level averages showing the relationship between model logit difference and human accuracy, with dashed lines connecting context and target prompt types within each category.}
\end{figure}

\newpage

\subsection{Mean category level alignment}
\label{app:mean-cat-alignment}
\paragraph{Open-source model category alignment}

\begin{figure}[H]
    \centering
    \includegraphics[width=0.6\linewidth]{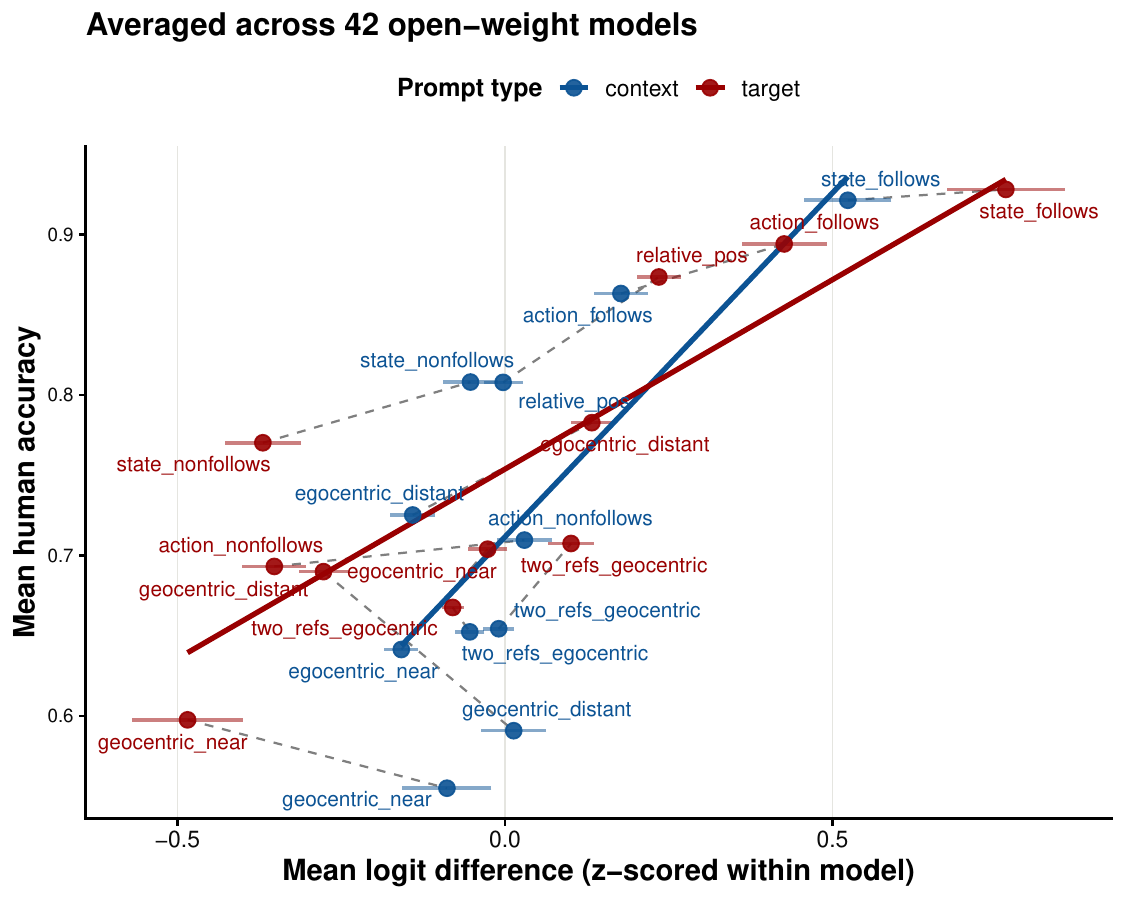}
    \caption{\textbf{Human accuracy versus logit difference averaged across models.} Category-level averages showing the relationship between model logit difference and human accuracy, with dashed lines connecting context and target prompt types within each category.}
\end{figure}
\newpage

\subsection{Item-level human-LLM alignment}
\label{app:item-align}
We reproduce the results shown in Figure \ref{fig:behavioral} for the five open-source models whose logit differences best predict human accuracy.

\begin{figure}[H]
    \centering
    \includegraphics[width=0.95\linewidth]{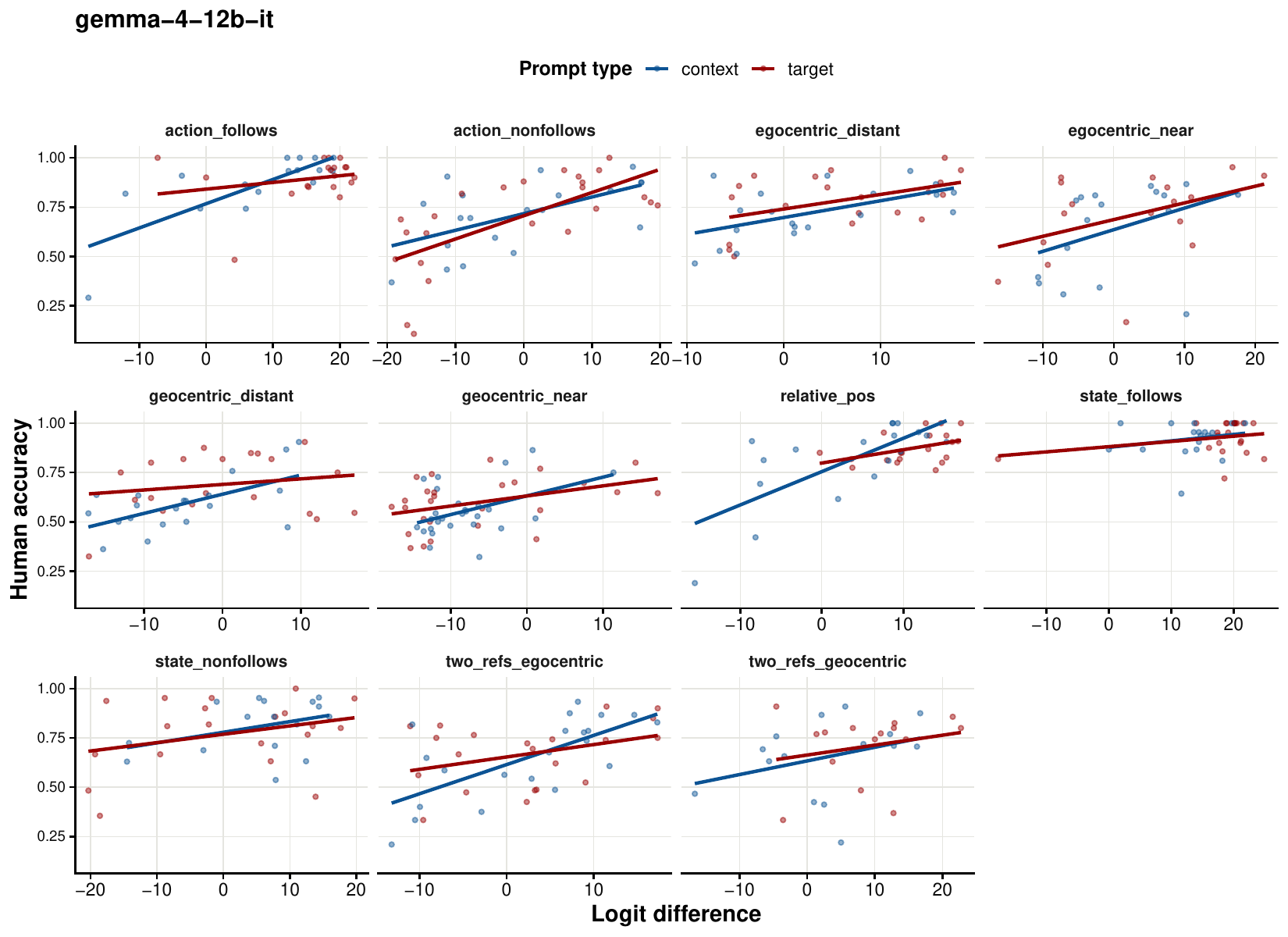}
    \caption{\textbf{Item-Level Human accuracy versus logit difference for Gemma 4 12B IT} Human accuracy plotted against model logit difference at the item level, faceted by category and colored by prompt type (context vs target), with least-squares fits per prompt type within each category.}
\end{figure}

\begin{figure}[H]
    \centering
    \includegraphics[width=0.95\linewidth]{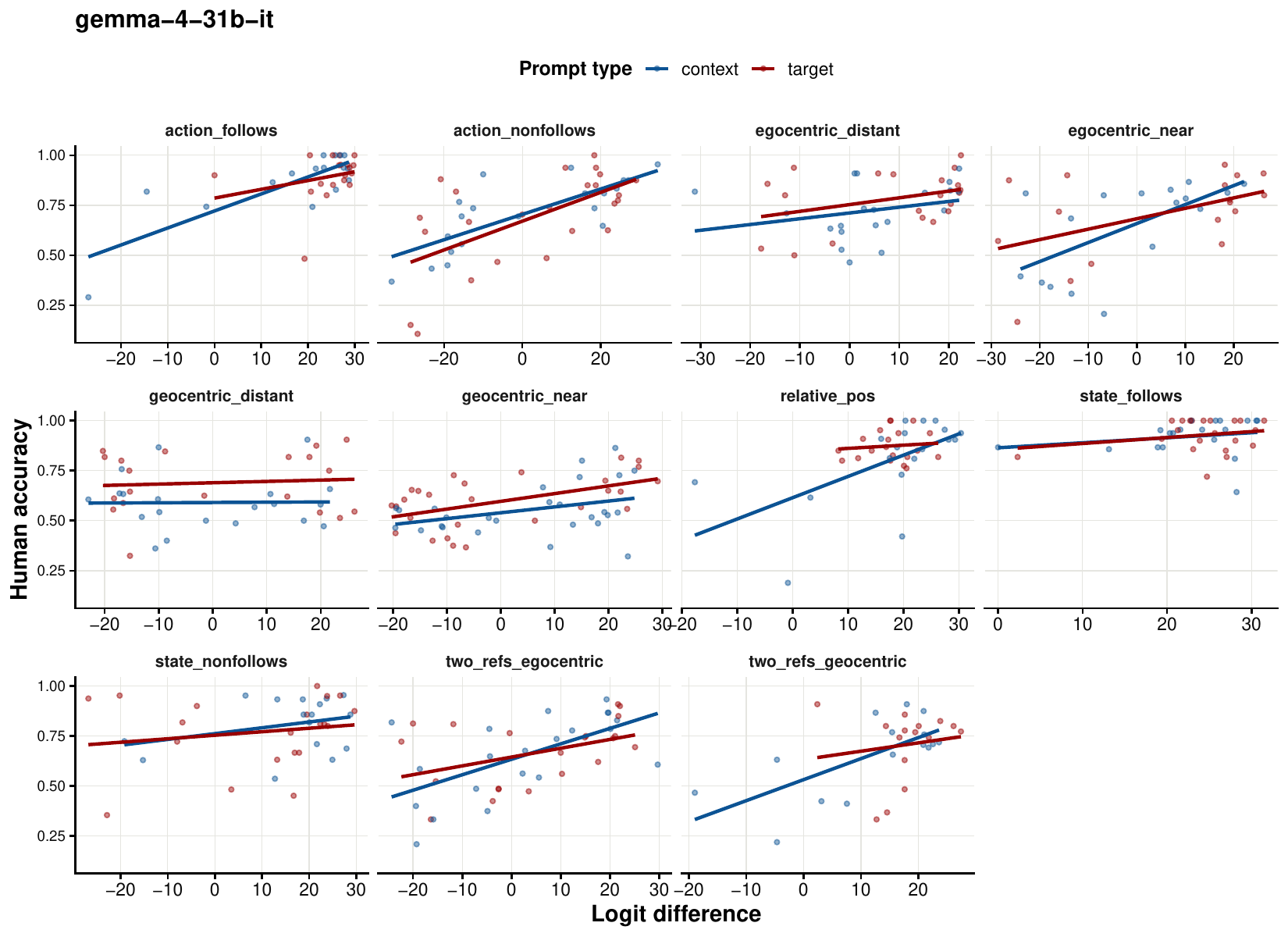}
    \caption{\textbf{Item-Level Human accuracy versus logit difference for Gemma 4 31B IT} Human accuracy plotted against model logit difference at the item level, faceted by category and colored by prompt type (context vs target), with least-squares fits per prompt type within each category.}
\end{figure}

\begin{figure}[H]
    \centering
    \includegraphics[width=0.95\linewidth]{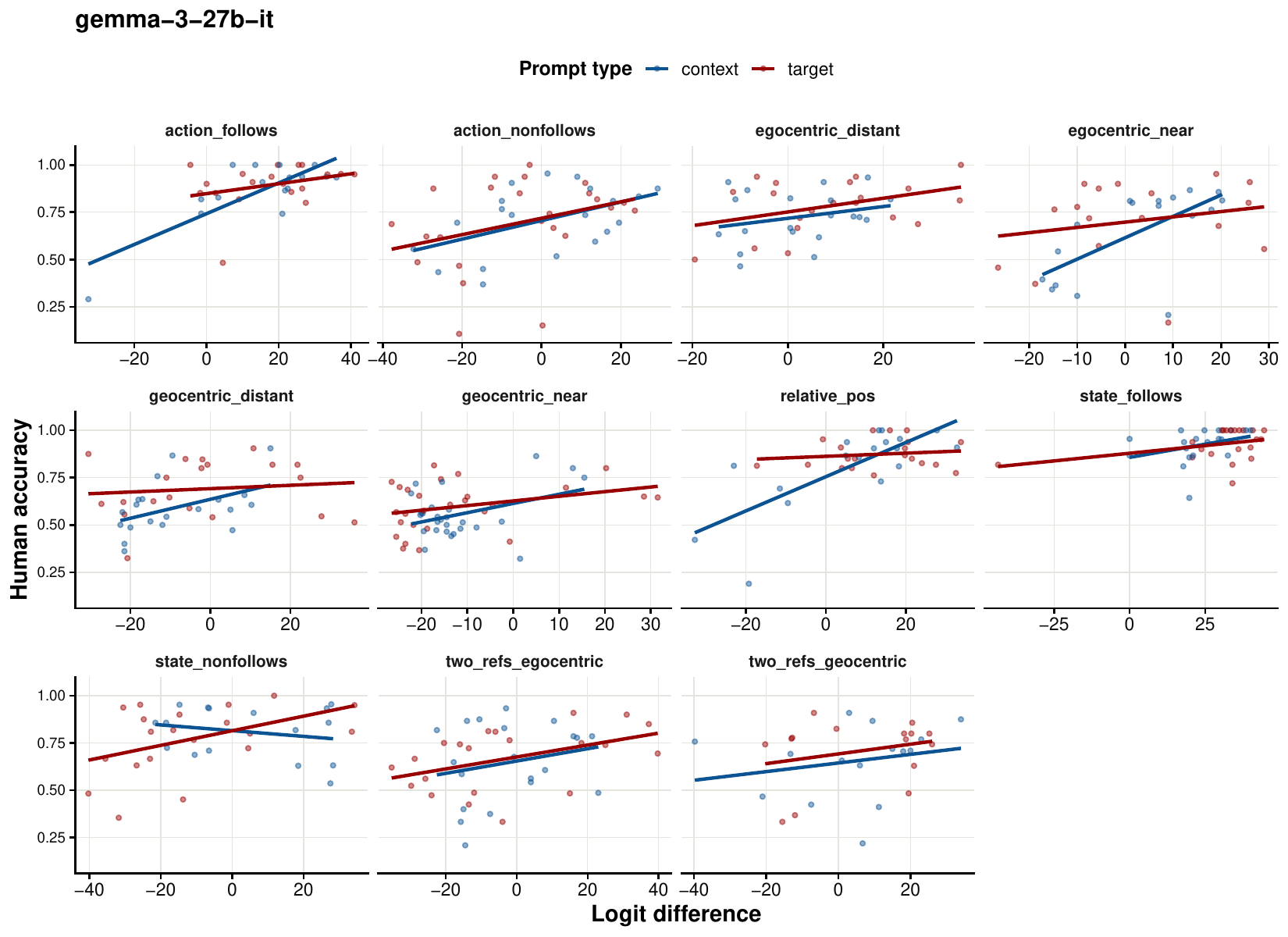}
    \caption{\textbf{Item-Level Human accuracy versus logit difference for Gemma 3 27B IT} Human accuracy plotted against model logit difference at the item level, faceted by category and colored by prompt type (context vs target), with least-squares fits per prompt type within each category.}
\end{figure}

\begin{figure}[H]
    \centering
    \includegraphics[width=0.95\linewidth]{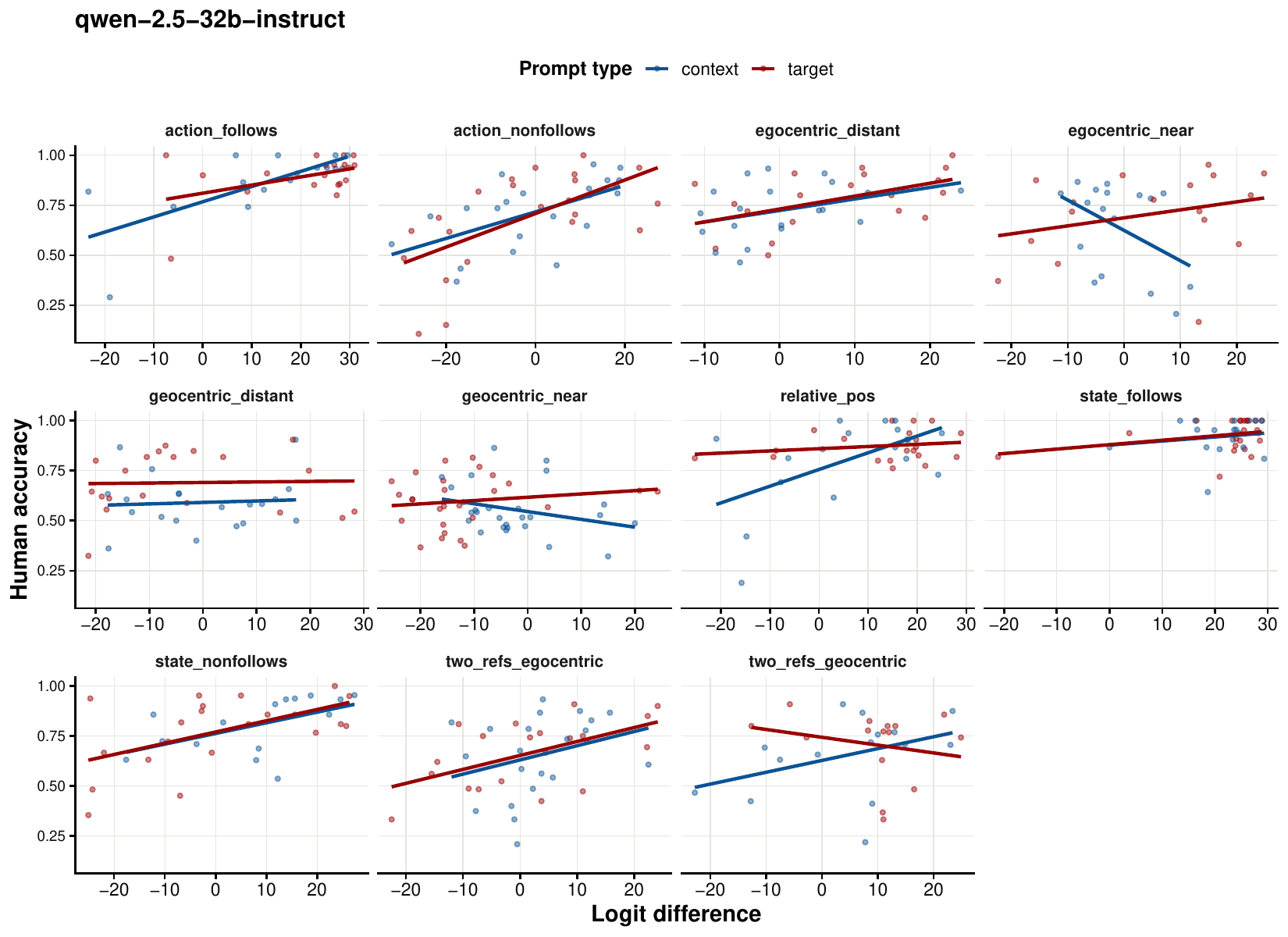}
    \caption{\textbf{Item-Level Human accuracy versus logit difference for Qwen 2.5 32B Instruct} Human accuracy plotted against model logit difference at the item level, faceted by category and colored by prompt type (context vs target), with least-squares fits per prompt type within each category.}
\end{figure}

\begin{figure}[H]
    \centering
    \includegraphics[width=0.95\linewidth]{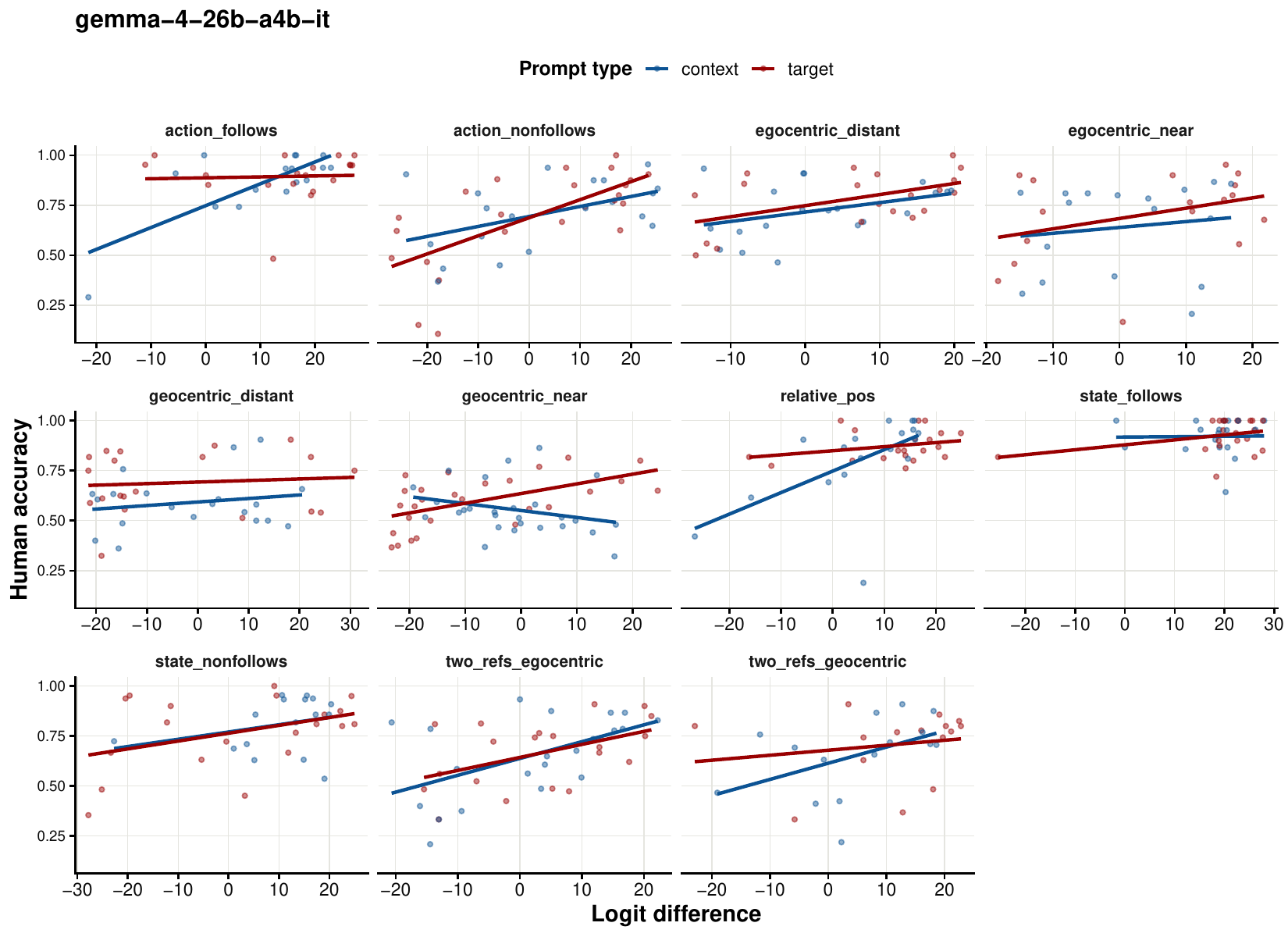}
    \caption{\textbf{Item-Level Human accuracy versus logit difference for Gemma 4 26B-A4B IT} Human accuracy plotted against model logit difference at the item level, faceted by category and colored by prompt type (context vs target), with least-squares fits per prompt type within each category.}
\end{figure}
\newpage
\subsection{Reasoning traces predict human accuracy}
\label{app:reasoning_traces_predict}

Following from \citet{de2025cost}, we asked whether the reasoning traces of \texttt{Deepseek-R1} are well-aligned with human accuracy, above category and prompt type. We include the accuracy(logit difference) of the most predictive open source model \texttt{gemma-3-27b-it} as an additional  control. We then evaluate whether \texttt{Deepseek-R1} spends more tokens thinking through the answers to problems where people perform worse. We observe a significant relationship between reasoning trace length and human accuracy and include these results below.

\begin{figure}[H]
    \centering
    \includegraphics[width=0.4\linewidth]{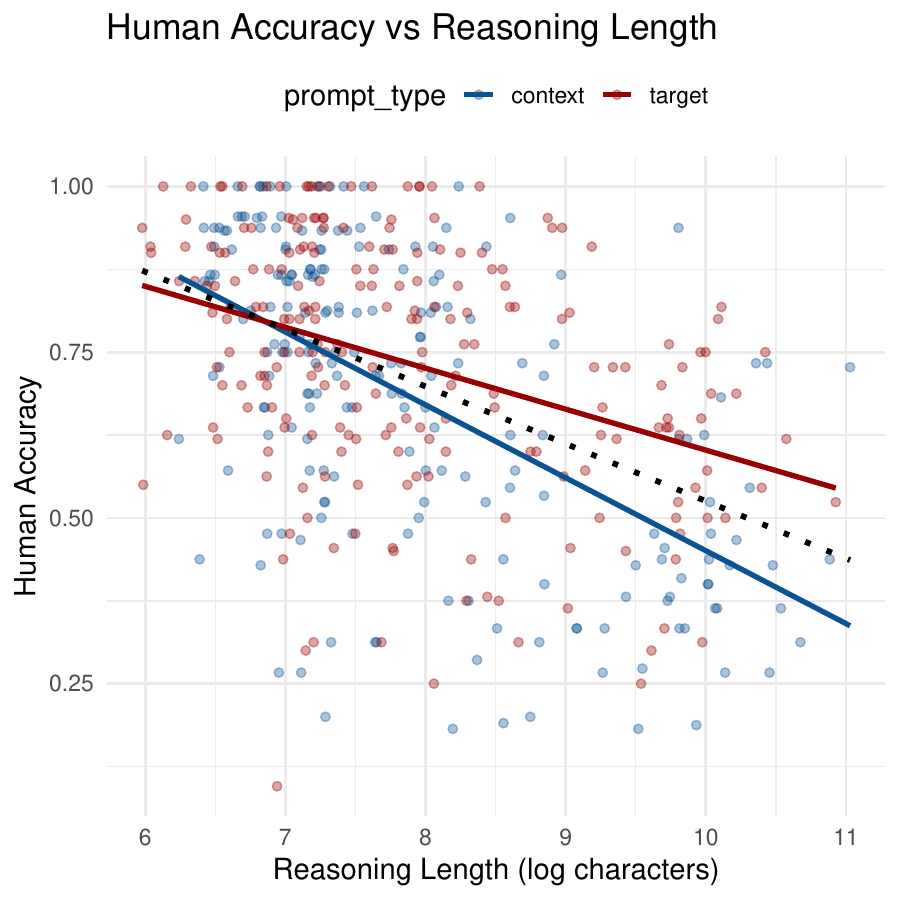}
    \caption{\textbf{Human Human accuracy vs Reasoning Length (DeepSeek R1)} Human accuracy plotted against log-transformed reasoning length from DeepSeek R1, showing separate regression lines for context (blue) and target (red) prompt types, with an overall trend line (black dotted).}
    \label{fig:reasoning_length}
\end{figure}

We fit a mixed effects model predicting human accuracy from reasoning length, gemma-3-27b-it logit difference, category, and prompt type, and find that reasoning length is a better predictor of human accuracy than model outputs, even when controlling for category and prompt type.

\begin{table}[H]
\centering
\caption{Main Regression: Human Accuracy Predicted by Reasoning Length with Controls}
\label{tab:reasoning_main_reg}
\begin{tabular}{lcccc}
\toprule
& Estimate & Std. Error & t value & p value \\
\midrule
(Intercept) & 0.987 & 0.070 & 14.125 & $< 0.001$*** \\
Reasoning Length (log) & $-0.027$ & 0.008 & $-3.313$ & 0.001** \\
Prompt Type: Target & 0.045 & 0.015 & 2.918 & 0.004** \\
Action (Non-follows) & $-0.105$ & 0.037 & $-2.825$ & 0.005** \\
Egocentric (Distant) & $-0.070$ & 0.037 & $-1.906$ & 0.057. \\
Egocentric (Near) & $-0.122$ & 0.039 & $-3.143$ & 0.002** \\
Geocentric (Distant) & $-0.141$ & 0.039 & $-3.635$ & $< 0.001$*** \\
Geocentric (Near) & $-0.221$ & 0.038 & $-5.849$ & $< 0.001$*** \\
Relative Position & $-0.040$ & 0.037 & $-1.087$ & 0.278 \\
State (Follows) & 0.007 & 0.037 & 0.189 & 0.850 \\
State (Non-follows) & $-0.025$ & 0.039 & $-0.647$ & 0.518 \\
Two Refs (Egocentric) & $-0.134$ & 0.037 & $-3.593$ & $< 0.001$*** \\
Two Refs (Geocentric) & $-0.188$ & 0.040 & $-4.744$ & $< 0.001$*** \\
Model Logit Diff (Gemma 3 27B IT) & 0.004 & 0.001 & 7.009 & $< 0.001$*** \\
\midrule
\multicolumn{5}{l}{$R^2 = 0.455$, Adjusted $R^2 = 0.438$} \\
\multicolumn{5}{l}{$F(13, 419) = 26.86$, $p < 0.001$} \\
\multicolumn{5}{l}{Residual SE = 0.158, df = 419, $n = 433$} \\
\bottomrule
\end{tabular}
\end{table}

\clearpage

\subsection{Regression details: human-model item-level alignment}
\label{app:regression-item-level}

\begin{table}[H]
\centering
\small
\caption{Regressions predicting item-level human accuracy from model logit difference ($z$-scored within model). Column 1 is an OLS regression for \texttt{gemma-4-12b-it} alone ($n = 433$ items); columns 2--3 are linear mixed-effects models across the 42 open-weight models, with a by-model random slope for logit difference. Column 3 adds distributional lexical baselines. Reference cell: \textit{action-follows}, \textit{context} framing. Standard errors in parentheses. $^{*}p<0.05$, $^{**}p<0.01$, $^{***}p<0.001$.}
\label{tab:lmm_item_level}
\begin{tabular}{lccc}
\toprule
& \textbf{gemma-4-12b-it} & \textbf{All models} & \textbf{All models} \\
& & & \textbf{+ baselines} \\
& Est. (SE) & Est. (SE) & Est. (SE) \\
\midrule
Intercept                      & 0.809 (0.033)$^{***}$ & 0.842 (0.004)$^{***}$ & 0.846 (0.004)$^{***}$ \\
Logit difference ($z$)         & 0.111 (0.029)$^{***}$ & 0.050 (0.005)$^{***}$ & 0.049 (0.005)$^{***}$ \\
Trigram frequency ($z$)        & ---                   & ---                   & 0.005 (0.001)$^{***}$ \\
GPT-2 surprisal ($z$)          & ---                   & ---                   & $-$0.009 (0.001)$^{***}$ \\
\midrule
\multicolumn{4}{l}{\textit{Category}} \\
\textit{action-nonfollows}     & $-$0.070 (0.039)      & $-$0.149 (0.006)$^{***}$ & $-$0.157 (0.006)$^{***}$ \\
\textit{egocentric-distant}    & $-$0.070 (0.039)      & $-$0.105 (0.006)$^{***}$ & $-$0.107 (0.006)$^{***}$ \\
\textit{egocentric-near}       & $-$0.122 (0.041)$^{**}$ & $-$0.184 (0.006)$^{***}$ & $-$0.189 (0.006)$^{***}$ \\
\textit{geocentric-distant}    & $-$0.133 (0.042)$^{**}$ & $-$0.215 (0.006)$^{***}$ & $-$0.215 (0.006)$^{***}$ \\
\textit{geocentric-near}       & $-$0.165 (0.045)$^{***}$ & $-$0.279 (0.006)$^{***}$ & $-$0.283 (0.006)$^{***}$ \\
\textit{relative-pos}          & $-$0.018 (0.041)      & $-$0.025 (0.006)$^{***}$ & $-$0.029 (0.006)$^{***}$ \\
\textit{state-follows}         & 0.066 (0.055)         & 0.054 (0.006)$^{***}$ & 0.054 (0.006)$^{***}$ \\
\textit{state-nonfollows}      & $-$0.029 (0.040)      & $-$0.066 (0.006)$^{***}$ & $-$0.066 (0.006)$^{***}$ \\
\textit{two-refs-egocentric}   & $-$0.147 (0.039)$^{***}$ & $-$0.197 (0.006)$^{***}$ & $-$0.201 (0.006)$^{***}$ \\
\textit{two-refs-geocentric}   & $-$0.146 (0.042)$^{***}$ & $-$0.181 (0.006)$^{***}$ & $-$0.182 (0.006)$^{***}$ \\
\midrule
\textit{target} framing        & 0.011 (0.014)         & 0.033 (0.002)$^{***}$ & 0.032 (0.002)$^{***}$ \\
\midrule
\multicolumn{4}{l}{\textit{Logit difference $\times$ Category}} \\
\textit{action-nonfollows}     & 0.031 (0.034)         & $-$0.003 (0.005)      & $-$0.004 (0.005) \\
\textit{egocentric-distant}    & 0.001 (0.041)         & $-$0.023 (0.006)$^{***}$ & $-$0.023 (0.006)$^{***}$ \\
\textit{egocentric-near}       & 0.020 (0.041)         & $-$0.023 (0.006)$^{***}$ & $-$0.022 (0.006)$^{***}$ \\
\textit{geocentric-distant}    & $-$0.018 (0.039)      & $-$0.045 (0.006)$^{***}$ & $-$0.045 (0.006)$^{***}$ \\
\textit{geocentric-near}       & $-$0.014 (0.039)      & $-$0.054 (0.005)$^{***}$ & $-$0.053 (0.005)$^{***}$ \\
\textit{relative-pos}          & 0.048 (0.043)         & $-$0.004 (0.006)      & $-$0.004 (0.006) \\
\textit{state-follows}         & $-$0.049 (0.044)      & $-$0.043 (0.005)$^{***}$ & $-$0.042 (0.005)$^{***}$ \\
\textit{state-nonfollows}      & $-$0.031 (0.036)      & $-$0.013 (0.005)$^{*}$ & $-$0.013 (0.005)$^{*}$ \\
\textit{two-refs-egocentric}   & 0.028 (0.039)         & $-$0.022 (0.006)$^{***}$ & $-$0.020 (0.006)$^{***}$ \\
\textit{two-refs-geocentric}   & $-$0.022 (0.043)      & $-$0.021 (0.006)$^{**}$ & $-$0.020 (0.006)$^{**}$ \\
\midrule
Logit difference $\times$ \textit{target} & $-$0.035 (0.015)$^{*}$ & $-$0.002 (0.002) & $-$0.001 (0.002) \\
\midrule
SD (logit difference $|$ model) & ---      & 0.019 & 0.020 \\
SD (residual)                   & 0.142    & 0.156 & 0.156 \\
Observations                    & 433      & 18{,}186 & 18{,}186 \\
Models                          & 1        & 42 & 42 \\
$R^2$                           & 0.487    & --- & --- \\
\bottomrule
\end{tabular}
\end{table}
\pagebreak

\subsection{Noise ceiling in predicting human responses}
\label{app:held-out-predict}
We ask whether the correlation (Pearson $r$) between each held out participant response and the mean participant response is significantly better than the correlation between each held out participant response and the \texttt{gemma-4-12b} response, within category and prompt type. Asterisks indicate significance values for a Williams' t test for each category x prompt type comparison.

\begin{figure}[H]
    \centering
    \includegraphics[width=0.6\linewidth]{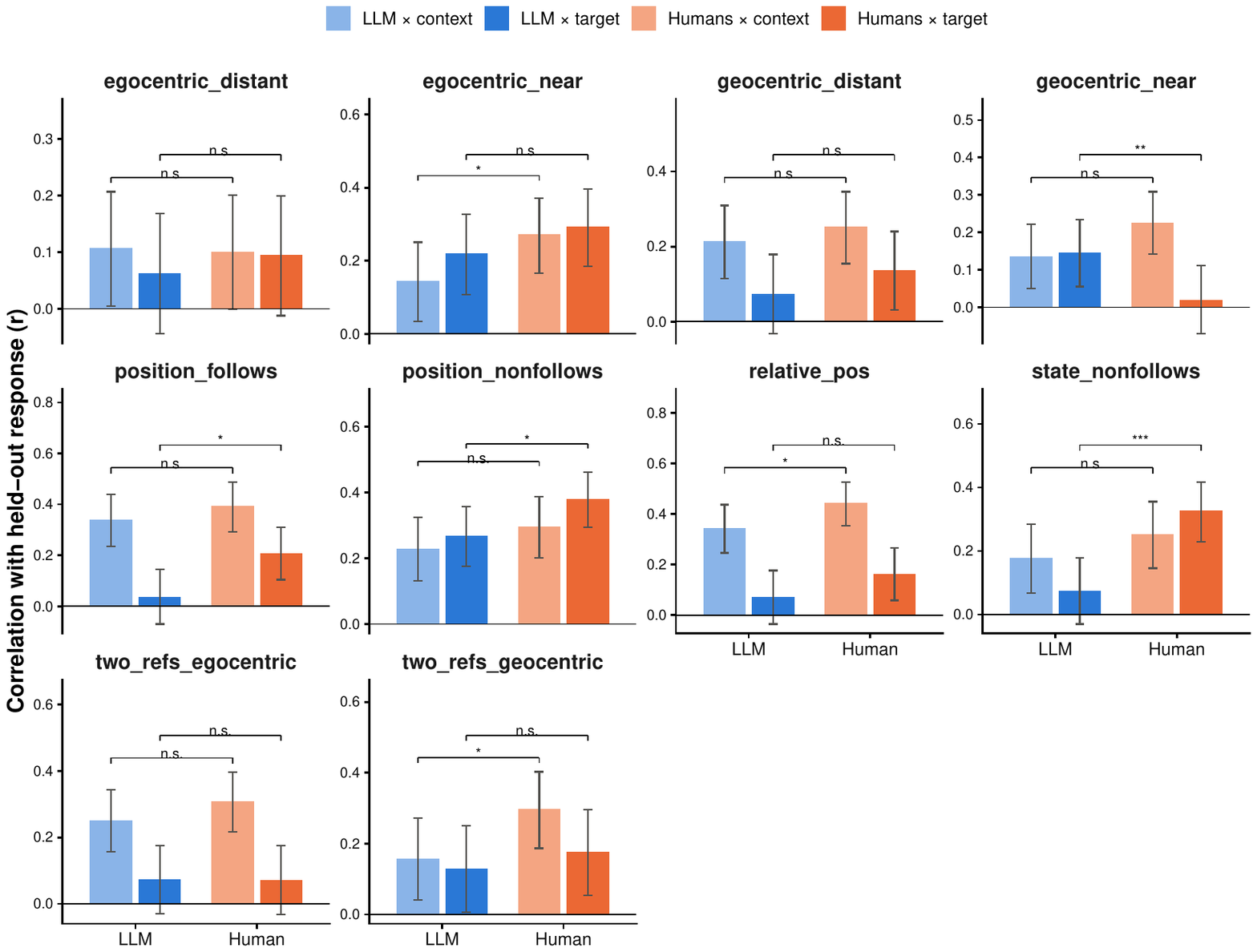}
    \caption{Differences in correlations for humans predicting other humans and \texttt{gemma-4-12b} predicting other humans at the item level.}
    \label{fig:heatmap_ablation_9b}
\end{figure}

\section{Mechanistic evaluation details}

We summarize the results of our post-hoc mechanistic interpretability analyses on the main causal reasoning stimuli below, and additionally include summaries of linear models connecting these analyses to human behavior.

\subsection{Head distributions}
\label{app:head-distributions}
Distributions of content-sensitive and abstraction-sensitive attention heads across model layers, for the top five aligned open-source models.

\begin{figure}[H]
    \centering
    \includegraphics[width=0.95\linewidth]{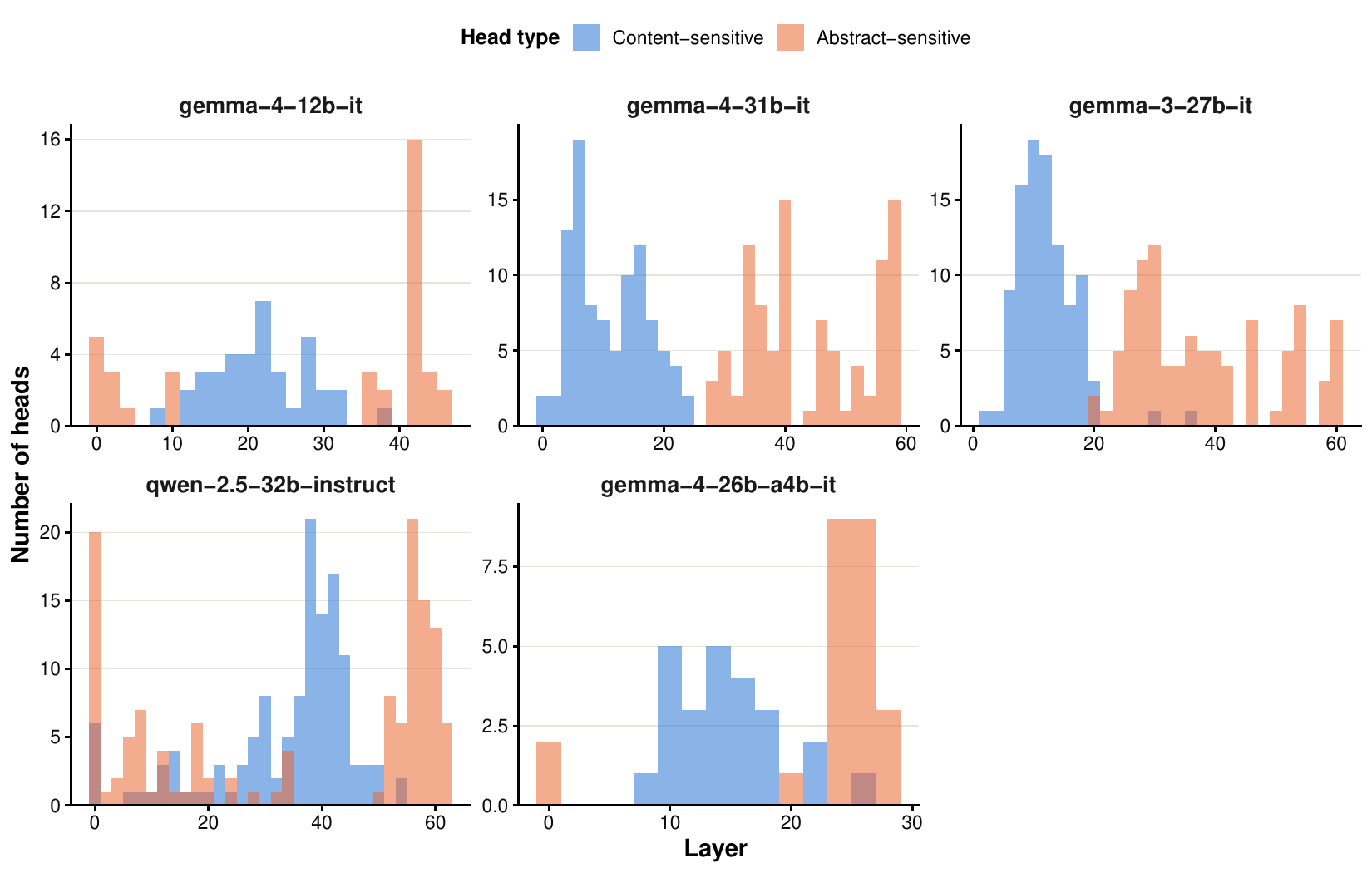}
    \caption{\textbf{Distribution of content-sensitive and abstraction-sensitive attention heads across layers.}}
    \label{fig:head_layer_distribution}
\end{figure}

\newpage

\subsection{Abstraction-sensitive and content-sensitive contrast stimuli}
\label{app:patching-stim}
To identify content-sensitive and abstraction-sensitive attention heads, we define three \textit{content relevant substitutions} per prompt (where substituting the new span has no effect on the ground-truth outcome) and three \textit{abstraction relevant substitutions} per prompt (where substituting the new span changes the ground truth correct response). We provide examples of these substitutions below, with two \textit{context} framing and two \textit{target} framing prompts per category.

\begin{table}[H]
\centering
\footnotesize
\caption{Example prompts and patching substitutions, position\_follows. \underline{Underlined} sections are swapped in content patches; \textbf{bolded} sections are swapped in critical patches.}
\begin{tabular}{@{}l p{0.36\textwidth} p{0.24\textwidth} p{0.24\textwidth}@{}}
\toprule
Framing & Prompt & Content substitutions (content relevant) & Critical substitutions (answer-flipping) \\
\midrule
\textit{context} & The \underline{wine glass} is on the BLANK. The wine glass falls over. The wine spills \textbf{on the table}. & \underline{wine glass} $\to$ \{beer mug, water cup, coffee mug\} & \textbf{on the table} $\to$ \{on the floor, on the shelf, on the counter\} \\
\addlinespace[2pt]
\textit{context} & The \underline{wine glass} is on the BLANK. Ali bumps the wine glass. The wine spills \textbf{on the floor}. & \underline{wine glass} $\to$ \{beer mug, water cup, coffee mug\} & \textbf{on the floor} $\to$ \{on the table, on the shelf, on the counter\} \\
\addlinespace[2pt]
\textit{target} & The \underline{wine} glass is \textbf{on the table}. The wine glass falls over. The wine spills on the BLANK. & \underline{wine} $\to$ \{juice, milk, broth\} & \textbf{on the table} $\to$ \{on the floor, on the shelf, on the counter\} \\
\addlinespace[2pt]
\textit{target} & The \underline{box} is in the bin. Ali \textbf{moves the bin}. The box BLANK. & \underline{box} $\to$ \{crate, chest, bin\} & \textbf{moves the bin} $\to$ \{moves nothing, touches the bin, taps the bin\} \\
\addlinespace[2pt]
\bottomrule
\end{tabular}
\end{table}

\begin{table}[H]
\centering
\footnotesize
\caption{Example prompts and patching substitutions, position\_nonfollows. \underline{Underlined} sections are swapped in content patches; \textbf{bolded} sections are swapped in critical patches.}
\begin{tabular}{@{}l p{0.36\textwidth} p{0.24\textwidth} p{0.24\textwidth}@{}}
\toprule
Framing & Prompt & Content substitutions (content relevant) & Critical substitutions (answer-flipping) \\
\midrule
\textit{context} & The \underline{wine glass} is on the table. Ali bumps the BLANK. The wine \textbf{doesn't spill}. & \underline{wine glass} $\to$ \{beer mug, water cup, coffee mug\} & \textbf{doesn't spill} $\to$ \{spills, spills over, tips over\} \\
\addlinespace[2pt]
\textit{context} & The \underline{box} is in the bin. Ali moves the BLANK. The bin \textbf{doesn't move}. & \underline{box} $\to$ \{crate, chest, bin\} & \textbf{doesn't move} $\to$ \{moves, rolls away, slides over\} \\
\addlinespace[2pt]
\textit{target} & The \underline{wine} glass is on the table. Ali \textbf{bumps the lamp}. The wine BLANK. & \underline{wine} $\to$ \{juice, milk, broth\} & \textbf{bumps the lamp} $\to$ \{bumps the glass, knocks the glass, tips the glass\} \\
\addlinespace[2pt]
\textit{target} & The \underline{box} is in the bin. Ali \textbf{moves the box}. The bin BLANK. & \underline{box} $\to$ \{crate, chest, bin\} & \textbf{moves the box} $\to$ \{moves the bin, lifts the bin, carries the bin\} \\
\addlinespace[2pt]
\bottomrule
\end{tabular}
\end{table}

\begin{table}[H]
\centering
\footnotesize
\caption{Example prompts and patching substitutions, egocentric\_near. \underline{Underlined} sections are swapped in content patches; \textbf{bolded} sections are swapped in critical patches.}
\begin{tabular}{@{}l p{0.36\textwidth} p{0.24\textwidth} p{0.24\textwidth}@{}}
\toprule
Framing & Prompt & Content substitutions (content relevant) & Critical substitutions (answer-flipping) \\
\midrule
\textit{context} & The \underline{painting} is BLANK Ali. Ali turns around. The painting is \textbf{behind} Ali. & \underline{painting} $\to$ \{vase, plant, statue\} & \textbf{behind} $\to$ \{in front of, left of, right of\} \\
\addlinespace[2pt]
\textit{context} & The \underline{painting} is BLANK Ali. Ali turns around. The painting is \textbf{in front of} Ali. & \underline{painting} $\to$ \{vase, plant, statue\} & \textbf{in front of} $\to$ \{behind, left of, right of\} \\
\addlinespace[2pt]
\textit{target} & The \underline{painting} is \textbf{in front of Ali}. Ali turns around. The painting is BLANK Ali. & \underline{painting} $\to$ \{vase, plant, statue\} & \textbf{in front of Ali} $\to$ \{behind Ali, left of Ali, right of Ali\} \\
\addlinespace[2pt]
\textit{target} & The \underline{painting} is \textbf{behind Ali}. Ali turns around. The painting is BLANK Ali. & \underline{painting} $\to$ \{vase, plant, statue\} & \textbf{behind Ali} $\to$ \{in front of Ali, left of Ali, right of Ali\} \\
\addlinespace[2pt]
\bottomrule
\end{tabular}
\end{table}

\begin{table}[H]
\centering
\footnotesize
\caption{Example prompts and patching substitutions, egocentric\_distant. \underline{Underlined} sections are swapped in content patches; \textbf{bolded} sections are swapped in critical patches.}
\begin{tabular}{@{}l p{0.36\textwidth} p{0.24\textwidth} p{0.24\textwidth}@{}}
\toprule
Framing & Prompt & Content substitutions (content relevant) & Critical substitutions (answer-flipping) \\
\midrule
\textit{context} & \underline{Chicago} is BLANK Ali. Ali turns around. Chicago is \textbf{behind} Ali. & \underline{Chicago} $\to$ \{Milwaukee, Seattle, Houston\} & \textbf{behind} $\to$ \{in front of, left of, right of\} \\
\addlinespace[2pt]
\textit{context} & \underline{Chicago} is BLANK Ali. Ali turns around. Chicago is \textbf{in front of} Ali. & \underline{Chicago} $\to$ \{Milwaukee, Seattle, Houston\} & \textbf{in front of} $\to$ \{behind, left of, right of\} \\
\addlinespace[2pt]
\textit{target} & \underline{Chicago} is \textbf{in front of Ali}. Ali turns around. Chicago is BLANK Ali. & \underline{Chicago} $\to$ \{Milwaukee, Seattle, Houston\} & \textbf{in front of Ali} $\to$ \{behind Ali, left of Ali, right of Ali\} \\
\addlinespace[2pt]
\textit{target} & \underline{Chicago} is \textbf{behind Ali}. Ali turns around. Chicago is BLANK Ali. & \underline{Chicago} $\to$ \{Milwaukee, Seattle, Houston\} & \textbf{behind Ali} $\to$ \{in front of Ali, left of Ali, right of Ali\} \\
\addlinespace[2pt]
\bottomrule
\end{tabular}
\end{table}

\begin{table}[H]
\centering
\footnotesize
\caption{Example prompts and patching substitutions, geocentric\_near. \underline{Underlined} sections are swapped in content patches; \textbf{bolded} sections are swapped in critical patches.}
\begin{tabular}{@{}l p{0.36\textwidth} p{0.24\textwidth} p{0.24\textwidth}@{}}
\toprule
Framing & Prompt & Content substitutions (content relevant) & Critical substitutions (answer-flipping) \\
\midrule
\textit{context} & The \underline{painting} is BLANK Ali. Ali turns around. The painting is \textbf{North of} Ali. & \underline{painting} $\to$ \{vase, plant, statue\} & \textbf{North of} $\to$ \{South of, East of, West of\} \\
\addlinespace[2pt]
\textit{context} & The \underline{painting} is BLANK Ali. Ali turns around. The painting is \textbf{South of} Ali. & \underline{painting} $\to$ \{vase, plant, statue\} & \textbf{South of} $\to$ \{North of, East of, West of\} \\
\addlinespace[2pt]
\textit{target} & The \underline{painting} is \textbf{North of Ali}. Ali turns around. The painting is BLANK Ali. & \underline{painting} $\to$ \{vase, plant, statue\} & \textbf{North of Ali} $\to$ \{South of Ali, East of Ali, West of Ali\} \\
\addlinespace[2pt]
\textit{target} & The \underline{painting} is \textbf{South of Ali}. Ali turns around. The painting is BLANK Ali. & \underline{painting} $\to$ \{vase, plant, statue\} & \textbf{South of Ali} $\to$ \{North of Ali, East of Ali, West of Ali\} \\
\addlinespace[2pt]
\bottomrule
\end{tabular}
\end{table}

\begin{table}[H]
\centering
\footnotesize
\caption{Example prompts and patching substitutions, geocentric\_distant. \underline{Underlined} sections are swapped in content patches; \textbf{bolded} sections are swapped in critical patches.}
\begin{tabular}{@{}l p{0.36\textwidth} p{0.24\textwidth} p{0.24\textwidth}@{}}
\toprule
Framing & Prompt & Content substitutions (content relevant) & Critical substitutions (answer-flipping) \\
\midrule
\textit{context} & \underline{Milwaukee} is BLANK Ali. Ali turns around. Milwaukee is \textbf{North of} Ali. & \underline{Milwaukee} $\to$ \{Chicago, Seattle, Houston\} & \textbf{North of} $\to$ \{South of, East of, West of\} \\
\addlinespace[2pt]
\textit{context} & \underline{Milwaukee} is BLANK Ali. Ali turns around. Milwaukee is \textbf{South of} Ali. & \underline{Milwaukee} $\to$ \{Chicago, Seattle, Houston\} & \textbf{South of} $\to$ \{North of, East of, West of\} \\
\addlinespace[2pt]
\textit{target} & \underline{Milwaukee} is \textbf{North of Ali}. Ali turns around. Milwaukee is BLANK Ali. & \underline{Milwaukee} $\to$ \{Chicago, Seattle, Houston\} & \textbf{North of Ali} $\to$ \{South of Ali, East of Ali, West of Ali\} \\
\addlinespace[2pt]
\textit{target} & \underline{Milwaukee} is \textbf{South of Ali}. Ali turns around. Milwaukee is BLANK Ali. & \underline{Milwaukee} $\to$ \{Chicago, Seattle, Houston\} & \textbf{South of Ali} $\to$ \{North of Ali, East of Ali, West of Ali\} \\
\addlinespace[2pt]
\bottomrule
\end{tabular}
\end{table}

\begin{table}[H]
\centering
\footnotesize
\caption{Example prompts and patching substitutions, relative\_pos. \underline{Underlined} sections are swapped in content patches; \textbf{bolded} sections are swapped in critical patches.}
\begin{tabular}{@{}l p{0.36\textwidth} p{0.24\textwidth} p{0.24\textwidth}@{}}
\toprule
Framing & Prompt & Content substitutions (content relevant) & Critical substitutions (answer-flipping) \\
\midrule
\textit{context} & The \underline{ball} and the box are on the table. The BLANK falls off the table. The box is \textbf{lower than} the ball. & \underline{ball} $\to$ \{cube, disc, stone\} & \textbf{lower than} $\to$ \{higher than, level with, far from\} \\
\addlinespace[2pt]
\textit{context} & The \underline{ball} and the box are on the table. Ali picks up the BLANK. The ball is \textbf{higher than} the box. & \underline{ball} $\to$ \{cube, disc, stone\} & \textbf{higher than} $\to$ \{lower than, level with, far from\} \\
\addlinespace[2pt]
\textit{target} & The \underline{ball} and the box are on the table. The box \textbf{falls off the table}. The box is BLANK the ball. & \underline{ball} $\to$ \{cube, disc, stone\} & \textbf{falls off the table} $\to$ \{is lifted up, is picked up, is raised up\} \\
\addlinespace[2pt]
\textit{target} & The \underline{ball} and the box are on the table. Ali \textbf{picks up the ball}. The ball is BLANK the box. & \underline{ball} $\to$ \{cube, disc, stone\} & \textbf{picks up the ball} $\to$ \{drops the ball, lowers the ball, pushes the ball\} \\
\addlinespace[2pt]
\bottomrule
\end{tabular}
\end{table}

\begin{table}[H]
\centering
\footnotesize
\caption{Example prompts and patching substitutions, state\_follows. \underline{Underlined} sections are swapped in content patches; \textbf{bolded} sections are swapped in critical patches.}
\begin{tabular}{@{}l p{0.36\textwidth} p{0.24\textwidth} p{0.24\textwidth}@{}}
\toprule
Framing & Prompt & Content substitutions (content relevant) & Critical substitutions (answer-flipping) \\
\midrule
\textit{context} & The \underline{glass} is on the table. Ali BLANK the glass. The table is \textbf{wet}. & \underline{glass} $\to$ \{mug, cup, flask\} & \textbf{wet} $\to$ \{dry, clean, spotless\} \\
\addlinespace[2pt]
\textit{context} & The \underline{room} is dark. Ali turns BLANK the light. The room \textbf{gets lighter}. & \underline{room} $\to$ \{hall, office, kitchen\} & \textbf{gets lighter} $\to$ \{gets darker, stays dark, stays dim\} \\
\addlinespace[2pt]
\textit{target} & The \underline{room} is dark. Ali \textbf{turns on the light}. The room gets BLANK. & \underline{room} $\to$ \{hall, office, kitchen\} & \textbf{turns on the light} $\to$ \{blows out the candle, turns off the light, unplugs the light\} \\
\addlinespace[2pt]
\textit{target} & The \underline{room} is dark. Ali \textbf{blows out the candle}. The room gets BLANK. & \underline{room} $\to$ \{hall, office, kitchen\} & \textbf{blows out the candle} $\to$ \{turns on the light, lights the candle, strikes a match\} \\
\addlinespace[2pt]
\bottomrule
\end{tabular}
\end{table}

\begin{table}[H]
\centering
\footnotesize
\caption{Example prompts and patching substitutions, state\_nonfollows. \underline{Underlined} sections are swapped in content patches; \textbf{bolded} sections are swapped in critical patches.}
\begin{tabular}{@{}l p{0.36\textwidth} p{0.24\textwidth} p{0.24\textwidth}@{}}
\toprule
Framing & Prompt & Content substitutions (content relevant) & Critical substitutions (answer-flipping) \\
\midrule
\textit{context} & The \underline{glass} is on the table. Ali fills the glass BLANK. The table is \textbf{dry}. & \underline{glass} $\to$ \{mug, cup, flask\} & \textbf{dry} $\to$ \{wet, soaked, damp\} \\
\addlinespace[2pt]
\textit{context} & The empty \underline{glass} is on the table. Ali fills the BLANK halfway. The glass is \textbf{dry}. & \underline{glass} $\to$ \{mug, cup, flask\} & \textbf{dry} $\to$ \{wet, soaked, damp\} \\
\addlinespace[2pt]
\textit{target} & The \underline{glass} is on the table. Ali \textbf{fills the glass halfway}. The table is BLANK. & \underline{glass} $\to$ \{mug, cup, flask\} & \textbf{fills the glass halfway} $\to$ \{overfills the glass, overflows the glass, floods the glass\} \\
\addlinespace[2pt]
\textit{target} & The empty \underline{glass} is on the table. Ali \textbf{fills the sink halfway}. The glass is BLANK. & \underline{glass} $\to$ \{mug, cup, flask\} & \textbf{fills the sink halfway} $\to$ \{overfills the glass, fills the glass fully, overflows the glass\} \\
\addlinespace[2pt]
\bottomrule
\end{tabular}
\end{table}

\begin{table}[H]
\centering
\footnotesize
\caption{Example prompts and patching substitutions, two\_refs\_egocentric. \underline{Underlined} sections are swapped in content patches; \textbf{bolded} sections are swapped in critical patches.}
\begin{tabular}{@{}l p{0.36\textwidth} p{0.24\textwidth} p{0.24\textwidth}@{}}
\toprule
Framing & Prompt & Content substitutions (content relevant) & Critical substitutions (answer-flipping) \\
\midrule
\textit{context} & \underline{Ali} is to Mark's BLANK. Ali turns around. Ali is \textbf{to Mark's right}. & \underline{Ali} $\to$ \{Sam, Jay, Lee\} & \textbf{to Mark's right} $\to$ \{to Mark's left, behind Mark, in front of Mark\} \\
\addlinespace[2pt]
\textit{context} & \underline{Ali} is to Mark's BLANK. Mark turns around. Ali is \textbf{to Mark's left}. & \underline{Ali} $\to$ \{Sam, Jay, Lee\} & \textbf{to Mark's left} $\to$ \{to Mark's right, behind Mark, in front of Mark\} \\
\addlinespace[2pt]
\textit{target} & \underline{Ali} is \textbf{to Mark's right}. Ali turns around. Ali is to Mark's BLANK. & \underline{Ali} $\to$ \{Sam, Jay, Lee\} & \textbf{to Mark's right} $\to$ \{to Mark's left, behind Mark, in front of Mark\} \\
\addlinespace[2pt]
\textit{target} & \underline{Ali} is \textbf{to Mark's right}. Mark turns around. Ali is to Mark's BLANK. & \underline{Ali} $\to$ \{Sam, Jay, Lee\} & \textbf{to Mark's right} $\to$ \{to Mark's left, behind Mark, in front of Mark\} \\
\addlinespace[2pt]
\bottomrule
\end{tabular}
\end{table}

\begin{table}[H]
\centering
\footnotesize
\caption{Example prompts and patching substitutions, two\_refs\_geocentric. \underline{Underlined} sections are swapped in content patches; \textbf{bolded} sections are swapped in critical patches.}
\begin{tabular}{@{}l p{0.36\textwidth} p{0.24\textwidth} p{0.24\textwidth}@{}}
\toprule
Framing & Prompt & Content substitutions (content relevant) & Critical substitutions (answer-flipping) \\
\midrule
\textit{context} & \underline{Mark} is facing BLANK and Ali is \textbf{facing West}. Mark turns around. Mark is facing West. & \underline{Mark} $\to$ \{Sam, Jay, Lee\} & \textbf{facing West} $\to$ \{facing East, facing North, facing South\} \\
\addlinespace[2pt]
\textit{context} & \underline{Mark} is directly BLANK of Milwaukee and Ali is directly South of Milwaukee. Mark is \textbf{North of Ali}. & \underline{Mark} $\to$ \{Sam, Jay, Lee\} & \textbf{North of Ali} $\to$ \{South of Ali, East of Ali, West of Ali\} \\
\addlinespace[2pt]
\textit{target} & \underline{Mark} is \textbf{facing East} and Ali is facing West. Mark turns around. Mark is facing BLANK. & \underline{Mark} $\to$ \{Sam, Jay, Lee\} & \textbf{facing East} $\to$ \{facing West, facing North, facing South\} \\
\addlinespace[2pt]
\textit{target} & \underline{Mark} is directly \textbf{North of Milwaukee} and Ali is directly South of Milwaukee. Mark is BLANK of Ali. & \underline{Mark} $\to$ \{Sam, Jay, Lee\} & \textbf{North of Milwaukee} $\to$ \{South of Milwaukee, East of Milwaukee, West of Milwaukee\} \\
\addlinespace[2pt]
\bottomrule
\end{tabular}
\end{table}

\pagebreak

\subsection{Partial correlations of attention head predictions against human accuracy}
\label{app:partial-corrs}
We first ask whether \textit{content} or \textit{abstraction} sensitive attention head outputs (specifically, the pre-projection attention output vectors), provide better fit to human responses. We evaluate whether $k$ content and abstraction sensitive heads at varying $k$ accurately model human responses alone, when controlling for category and prompt type, and when controlling for category, prompt type, mean trigram frequency (see Section \ref{sec:head-predictions}) and \texttt{GPT-2} surprisal.

We then compare the predictions of content sensitive and abstraction sensitive heads against \textit{eachother} using a non-parametric permutation test. On each permutation ($n=20000$), we randomly sample one squared error observation from the set of content head predictions, and one squared error observation from the set of abstraction head predictions. Across many permutations, the effect size estimates the difference in $R^2$ between the abstraction and content-sensitive distributions, where the null hypothesis is that the two distributions are equivalent.

\begin{figure}[H]
    \centering
    \includegraphics[width=1\linewidth]{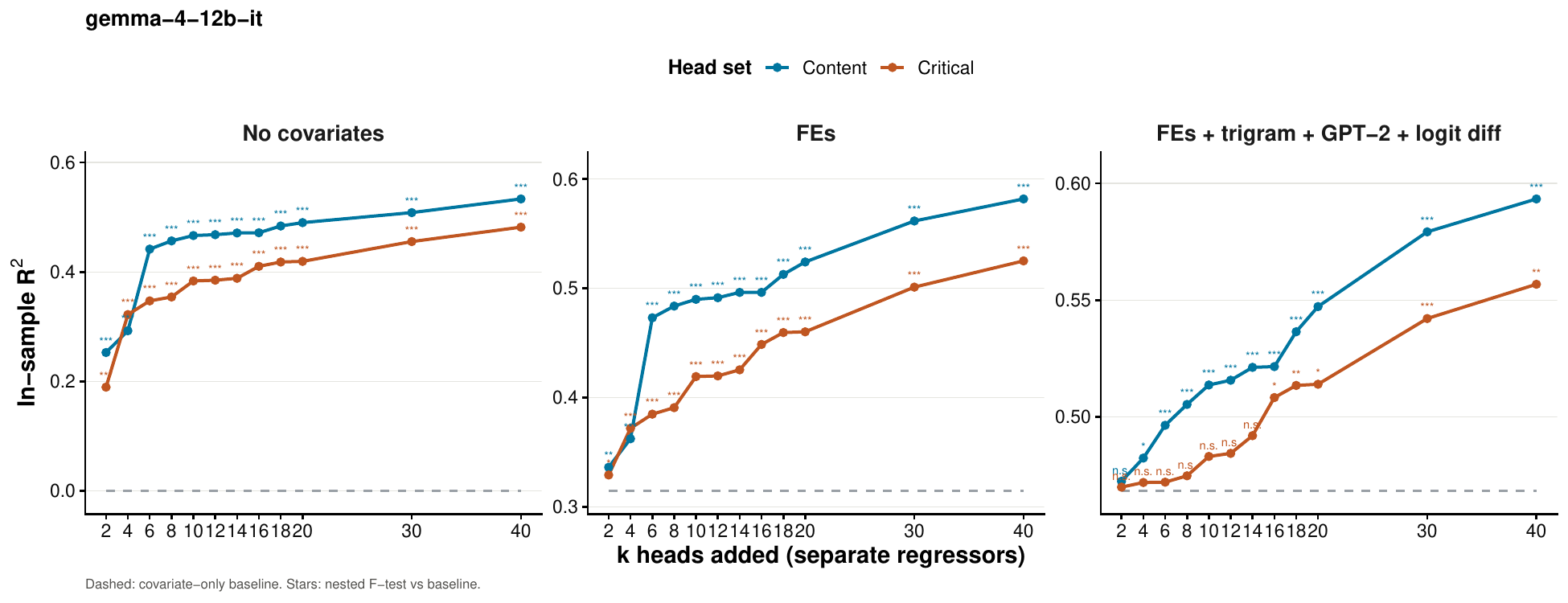}
    \caption{F-test nested model comparisons for \texttt{gemma-4-12b-it} content-sensitive and abstraction-sensitive attention heads, in-distribution $R^2$ at varying $k$. Asterisks indicate p-values for nested model comparisons against the base model. We compare nested model fit 1. with no covariates, 2. controlling for category and prompt type as fixed effects, and 3. controlling for FEs and distributional baselines.}
\end{figure}

\begin{figure}[H]
    \centering
    \includegraphics[width=0.8\linewidth]{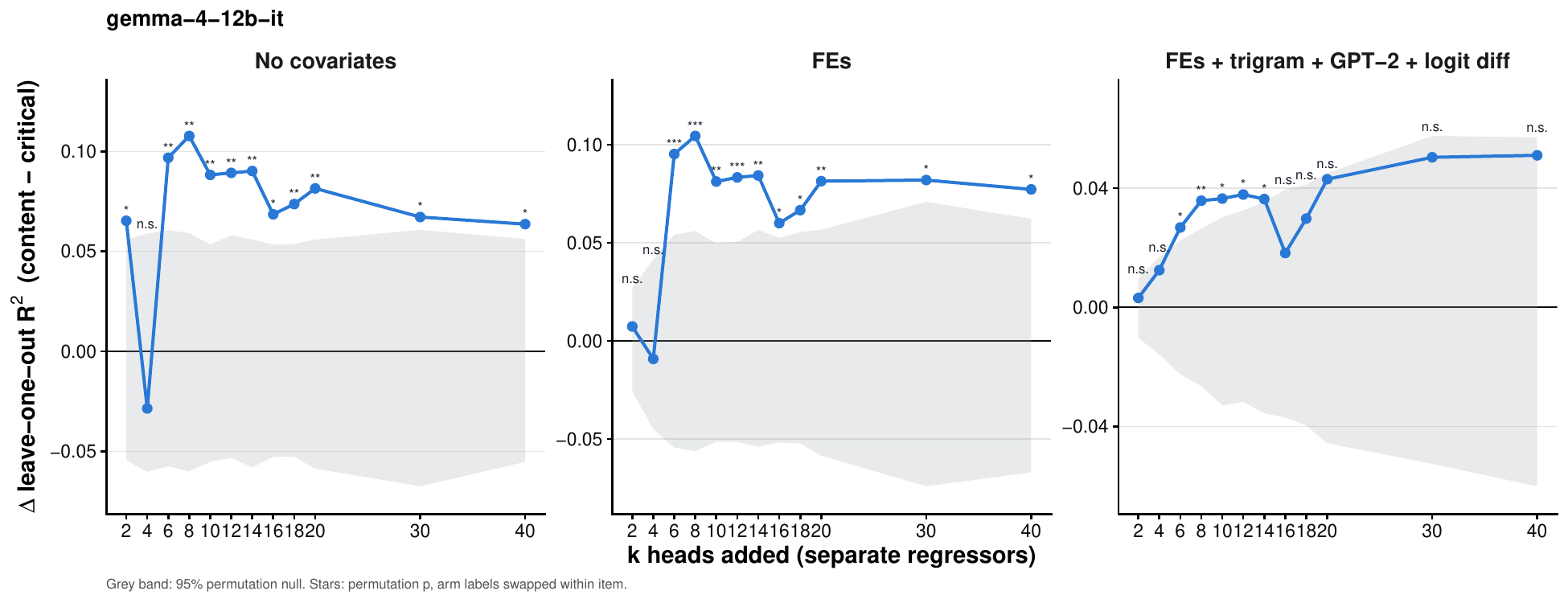}
    \caption{Permutation tests for \texttt{gemma-4-12b-it} content-sensitive and abstraction-sensitive attention heads, comparing out of distribution $R^2$ at varying $k$. Asterisks indicate p-values for nested model comparisons against the base model. We compare nested model fit 1. with no covariates, 2. controlling for category and prompt type as fixed effects, and 3. controlling for FEs and distributional baselines.}
\end{figure}
\pagebreak
\begin{figure}[H]
    \centering
    \includegraphics[width=0.8\linewidth]{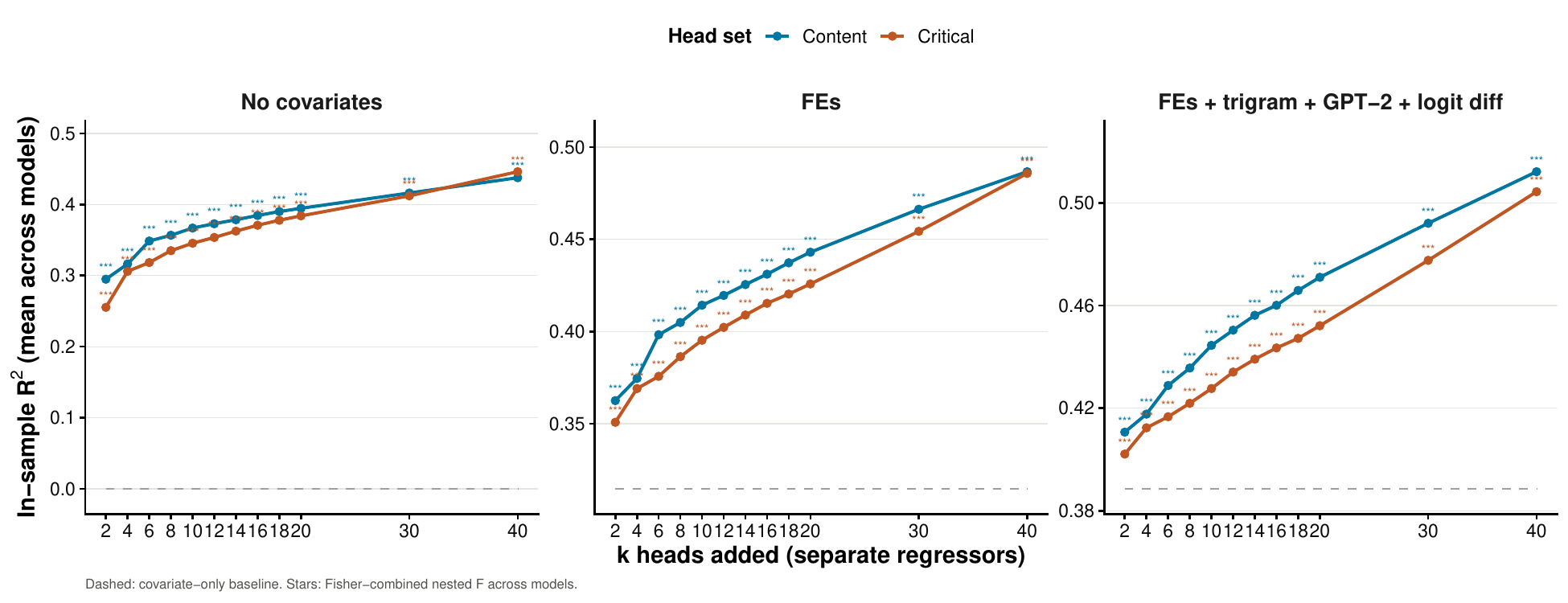}
    \caption{F-test nested model comparisons for all 15 LLMs with mechanistic results: content-sensitive and abstraction-sensitive attention heads, in-distribution $R^2$ at varying $k$. Asterisks indicate p-values for nested model comparisons against the base model. We compare nested model fit 1. with no covariates, 2. controlling for category and prompt type as fixed effects, and 3. controlling for FEs and distributional baselines.}
\end{figure}

\begin{figure}[H]
    \centering
    \includegraphics[width=1\linewidth]{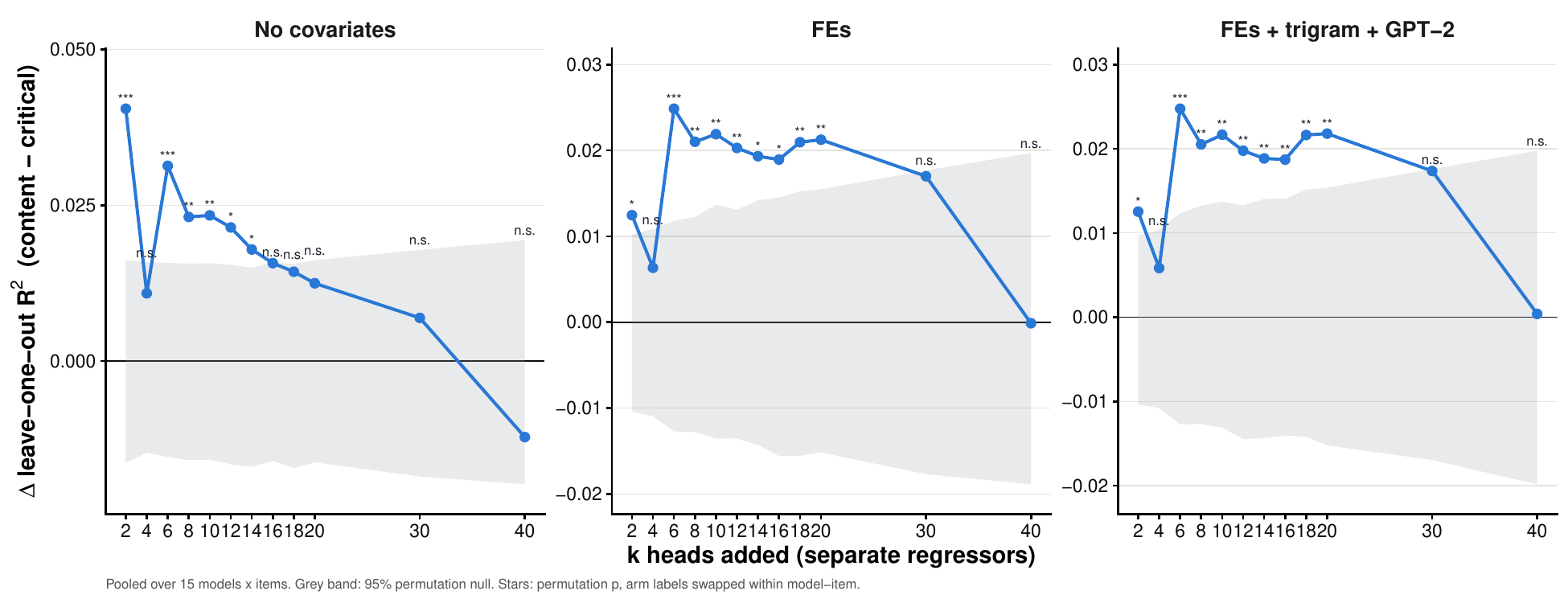}
    \caption{Permutation tests for all 15 LLM content-sensitive and abstraction-sensitive attention heads, comparing out of distribution $R^2$ at varying $k$. Asterisks indicate p-values for nested model comparisons against the base model. We compare nested model fit 1. with no covariates, 2. controlling for category and prompt type as fixed effects, and 3. controlling for FEs and distributional baselines.}
\end{figure}

\pagebreak

\newpage
\subsection{Per-category breakdown of head predictiveness}
\label{app:per-category-breakdown}
We include per-category predictions of human accuracy below from the top-10 content-sensitive attention heads in \texttt{gemma-4-12b-it}. Pearson r values indicate correlation of leave-one-out predictions against actual human accuracy. 

\begin{figure}[H]
    \centering
    \includegraphics[width=0.8\linewidth]{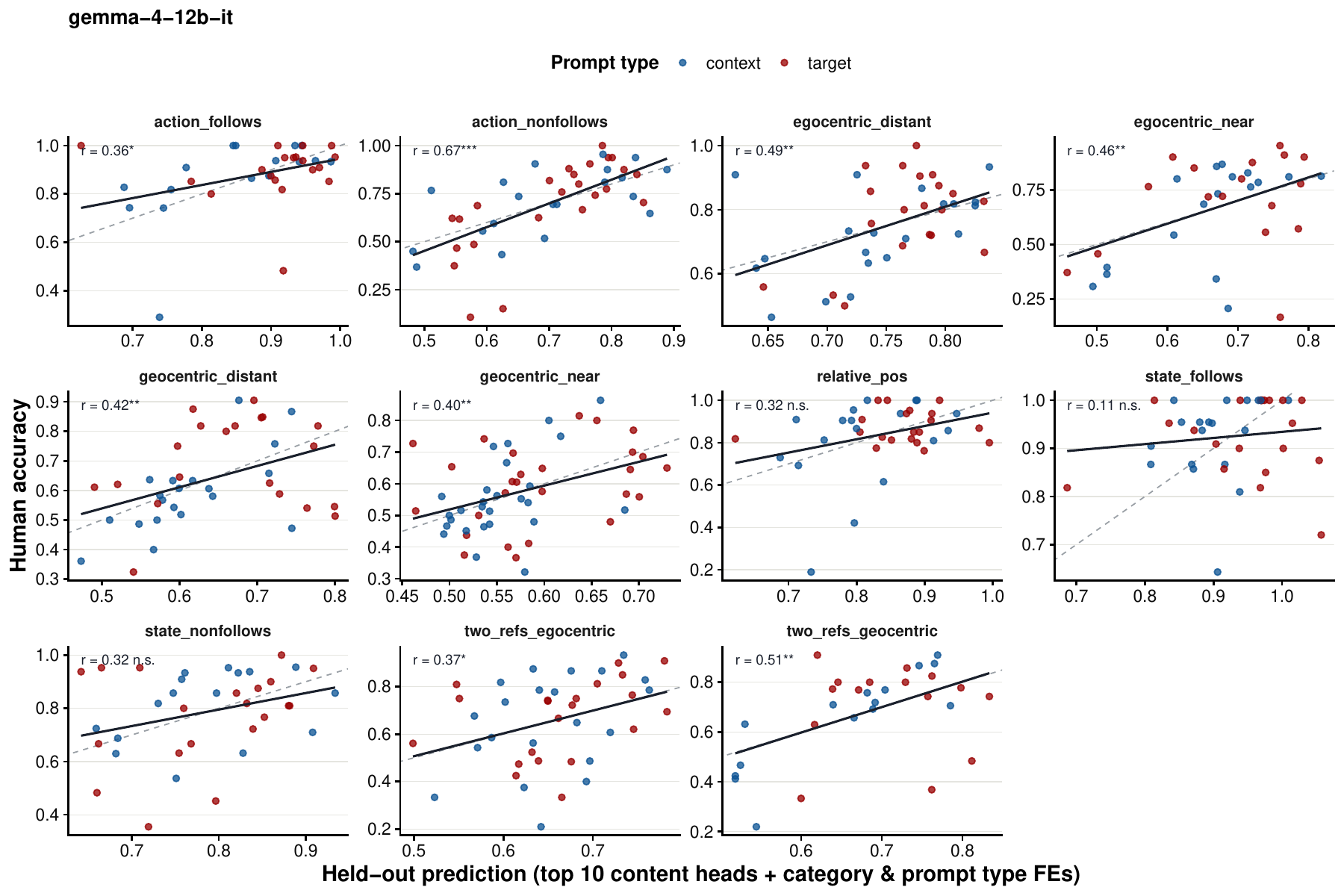}
    \caption{\textbf{Per-scenario predictions of human acccuracy.} Y-axis indicates ground-truth mean human accuracy. X-axis shows predicted human accuracy from the top ten content-sensitive attention head activations.}
\end{figure}

\subsection{Head ablation results}
\label{app:head-ablation-general}

\subsubsection{Correlation effects}
\begin{table}[H]
\centering
\small
\caption{Summary of linear mixed effects model for human-\texttt{gemma-4-12b} correlations }
\label{tab:general_eval_corr_model}
\begin{tabular}{lccc}
\toprule
\textbf{Fixed effect} & \textbf{Est.} & \textbf{SE} & \textbf{\textit{t}} \\
\midrule
Intercept                        & 0.191    & 0.014 & 14.08 \\
Baseline correlation ($z$)       & 0.092    & 0.007 & 13.08 \\
Abstraction vs.\ content         & 0.201    & 0.015 & 13.85 \\
Content vs.\ random              & $-$0.214 & 0.015 & $-$14.69 \\
\bottomrule
\end{tabular}
\end{table}

\begin{figure}[H]
    \centering
    \includegraphics[width=1\linewidth]{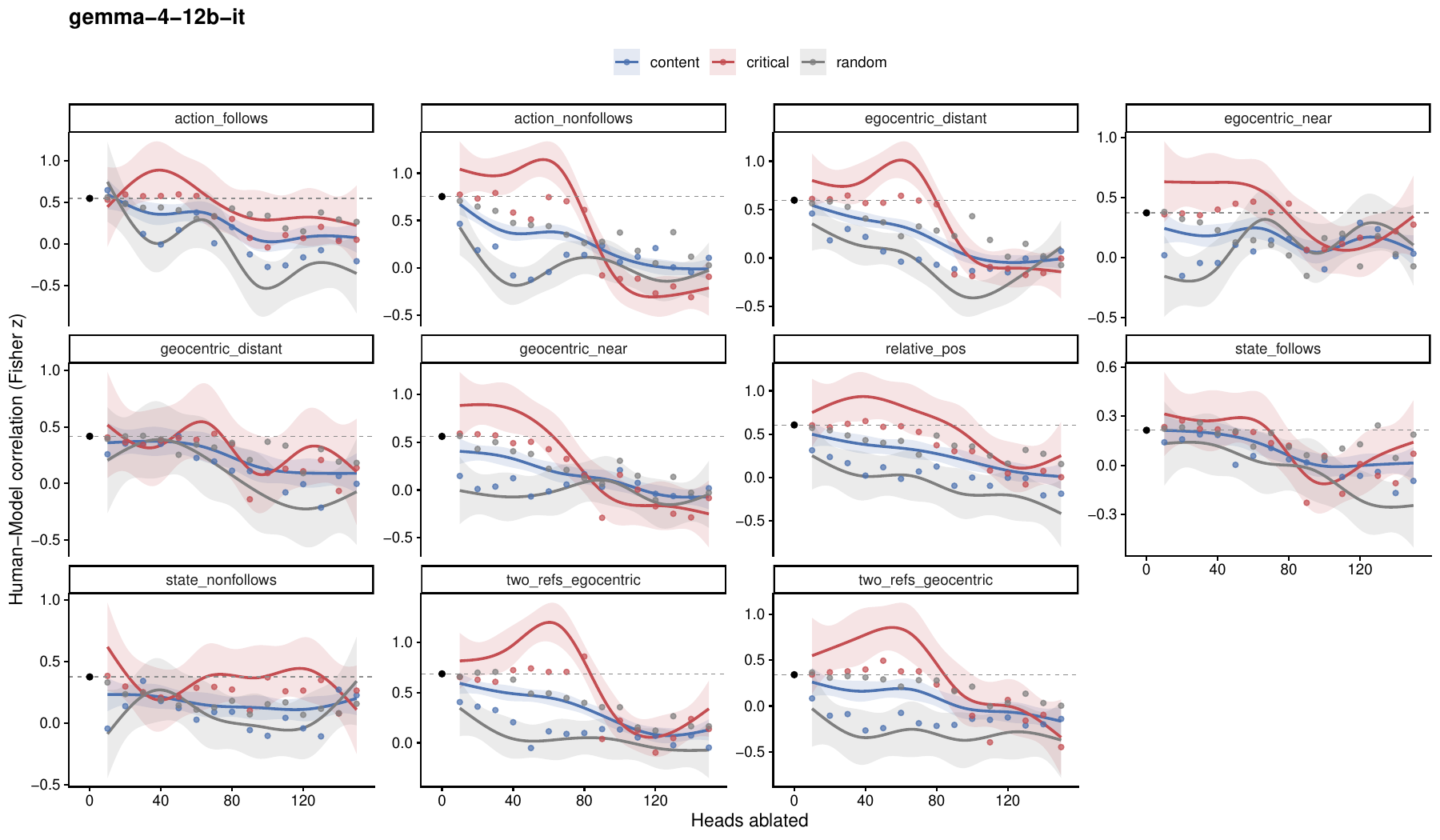}
    \caption{Per-category correlation effects of ablations for \texttt{gemma-4-12b-it}.}
\end{figure}

\subsubsection{Accuracy effects}

\begin{figure}[H]
    \centering
    \includegraphics[width=0.6\linewidth]{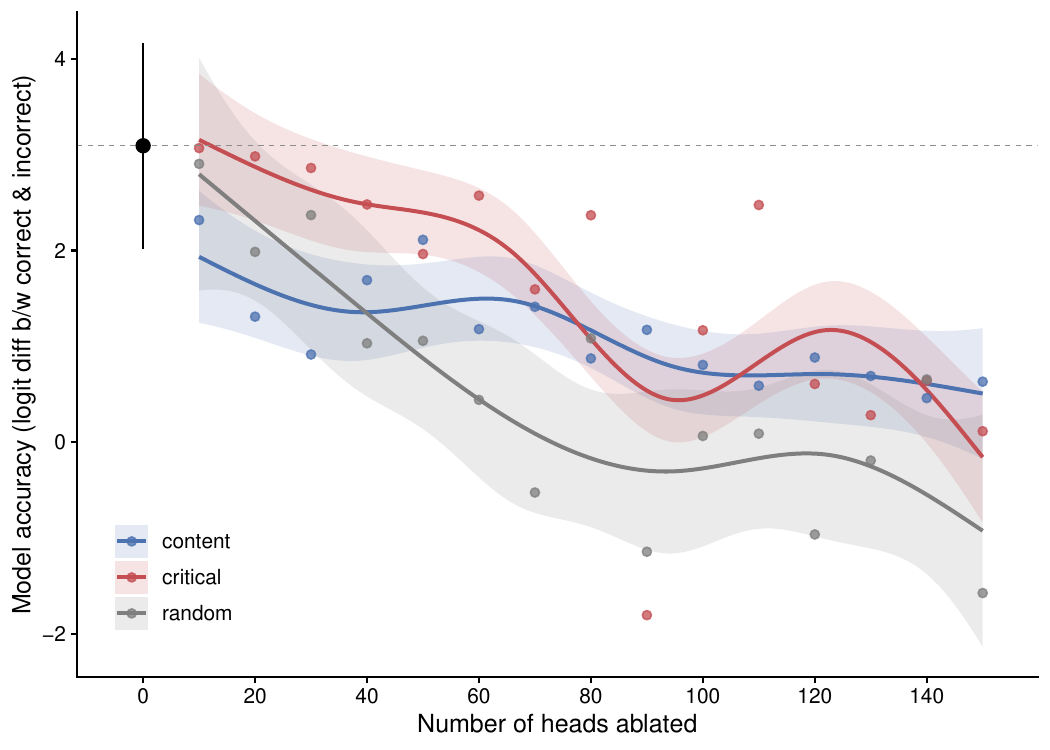}
    \caption{Effect of abstraction-sensitive, content-sensitive, and random heads on overall model \textit{accuracy}.}
\end{figure}

\begin{table}[H]
\centering
\small
\caption{Summary of linear mixed effects model for \texttt{gemma-4-12b-it} accuracy (logit difference)}
\label{tab:general_eval_acc_model}
\begin{tabular}{lccc}
\toprule
\textbf{Fixed effect} & \textbf{Est.} & \textbf{SE} & \textbf{\textit{t}} \\
\midrule
Intercept                        & 1.038 & 0.161 & 6.47  \\
Baseline logit difference ($z$)  & 4.288 & 0.092 & 46.71 \\
Abstraction vs.\ content         & 0.422 & 0.086 & 4.89  \\
Content vs.\ random              & 0.651 & 0.166 & 3.91  \\
\bottomrule
\end{tabular}
\end{table}

\begin{figure}[H]
    \centering
    \includegraphics[width=1\linewidth]{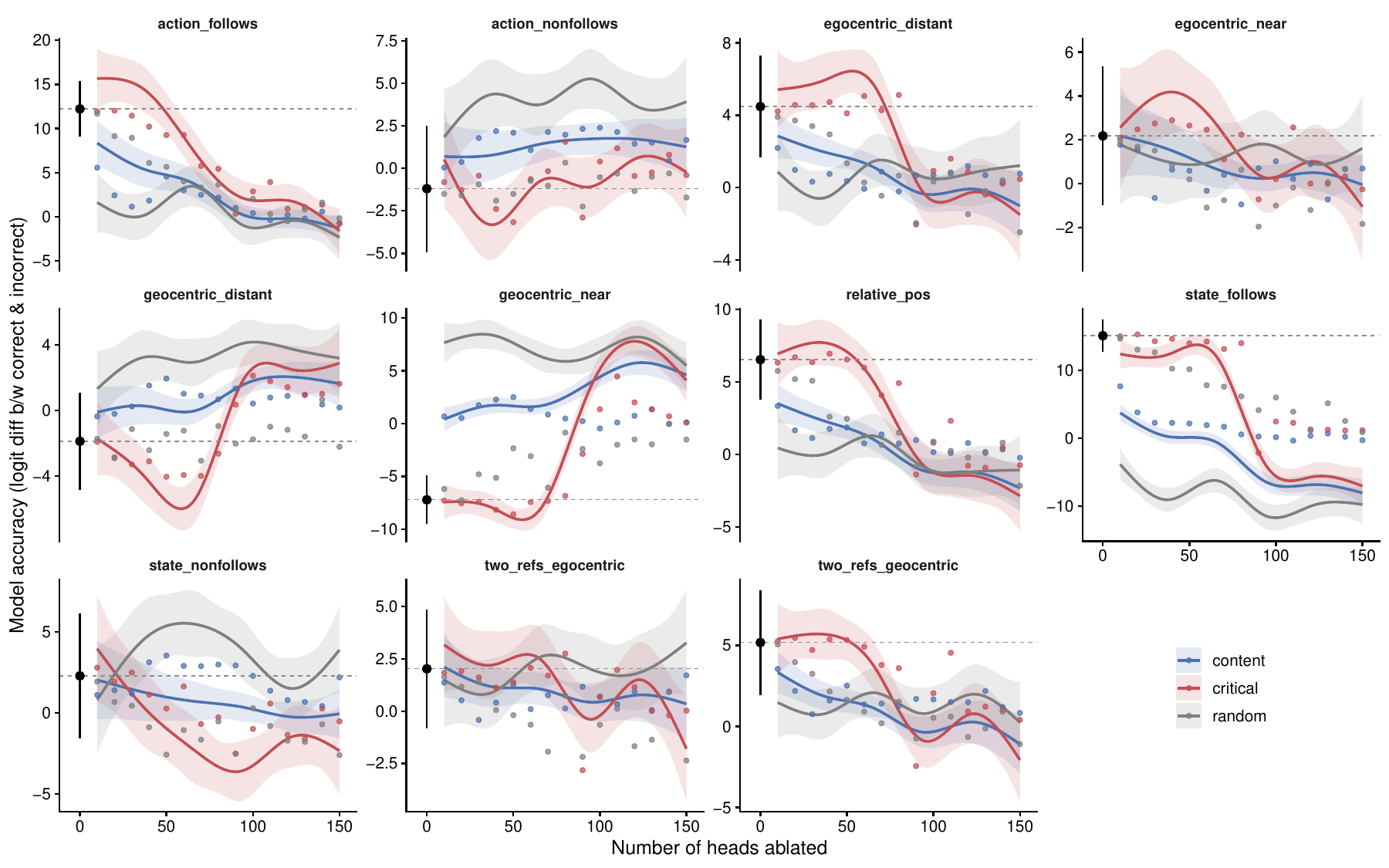}
    \caption{Per-category model accuracy effects of ablations for \texttt{gemma-4-12b-it}.}
\end{figure}

\subsubsection{Aggregate correlation and accuracy effects across 15 LLMs}

\begin{figure}[H]
    \centering
    \includegraphics[width=0.8\linewidth]{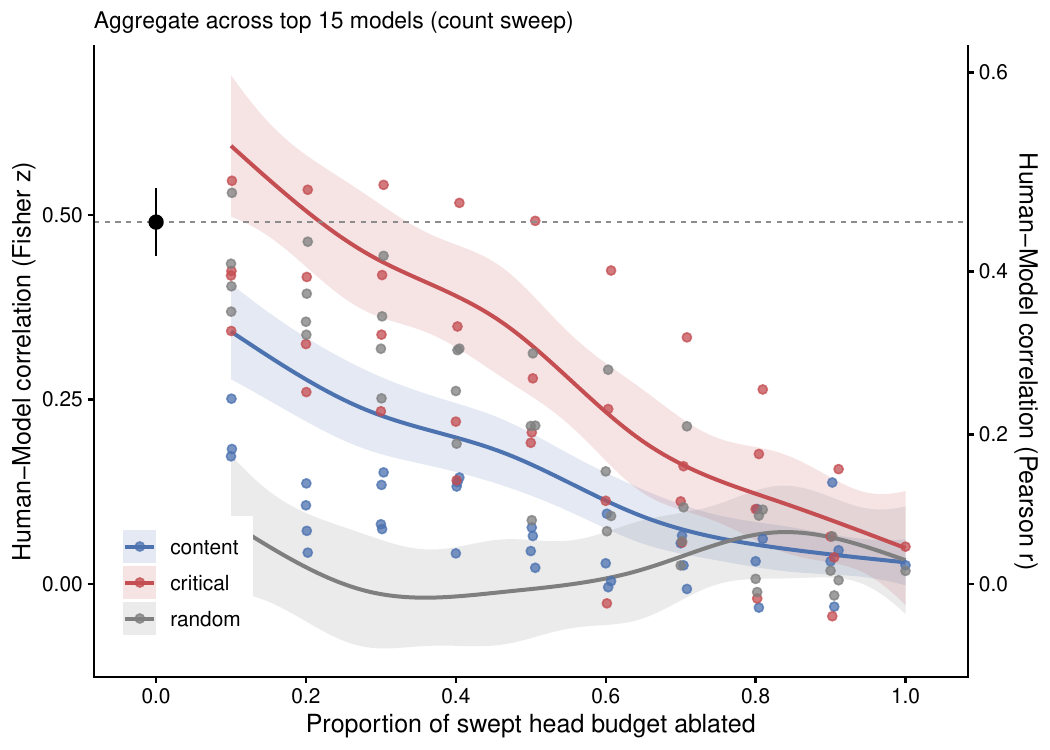}
    \caption{Model correlation ablation effects aggregated across models.}
\end{figure}

\begin{table}[H]
\centering
\small
\caption{Summary of linear mixed effects model for human--LLM correlations, aggregated across 15 models}
\label{tab:aggregate_corr_model}
\begin{tabular}{lccc}
\toprule
\textbf{Fixed effect} & \textbf{Est.} & \textbf{SE} & \textbf{\textit{t}} \\
\midrule
Intercept                        & 0.156    & 0.018 & 8.56 \\
Baseline correlation ($z$)       & $-$0.013 & 0.014 & $-$0.92 \\
Abstraction vs.\ content         & 0.138    & 0.012 & 11.73 \\
Content vs.\ random              & $-$0.124 & 0.009 & $-$13.58 \\
\bottomrule
\end{tabular}
\end{table}

\begin{figure}[H]
    \centering
    \includegraphics[width=0.8\linewidth]{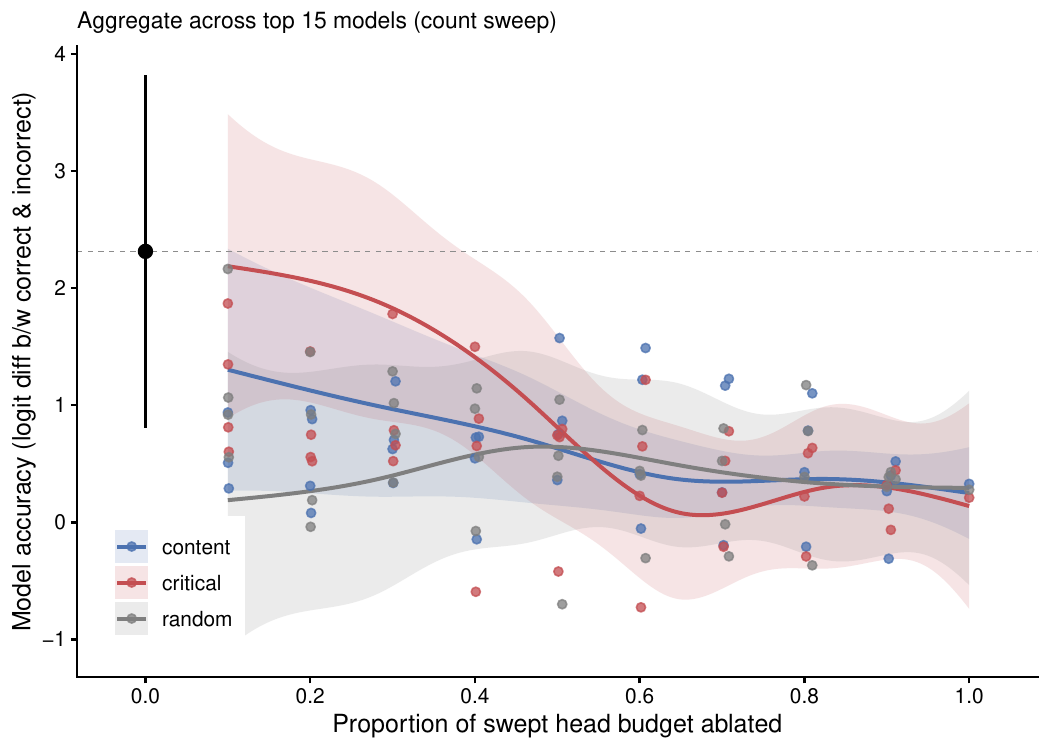}
    \caption{Model accuracy ablation effects aggregated across models.}
\end{figure}

\begin{table}[H]
\centering
\small
\caption{Summary of linear mixed effects model for LLM accuracy (logit difference), aggregated across 15 models}
\label{tab:aggregate_acc_model}
\begin{tabular}{lccc}
\toprule
\textbf{Fixed effect} & \textbf{Est.} & \textbf{SE} & \textbf{\textit{t}} \\
\midrule
Intercept                        & 0.644    & 0.186 & 3.46  \\
Baseline logit difference ($z$)  & 0.180    & 0.052 & 3.48  \\
Abstraction vs.\ content         & 0.271    & 0.131 & 2.07  \\
Content vs.\ random              & $-$0.258 & 0.101 & $-$2.54 \\
\bottomrule
\end{tabular}
\end{table}

\subsection{DLA versus direct ablation convergence}
\label{app:dla-effect}
Although direct logit attribution is often used in interpretability analyses \citep{wang2022interpretability}, we confirm that--for our evaluation stimuli-- actual head ablations have effects consistent with DLA. Empirically, we observe a strong degree of alignment between these methods when estimating the ablation effect for individual heads.

\begin{figure}[H]
    \centering
    \includegraphics[width=0.8\linewidth]{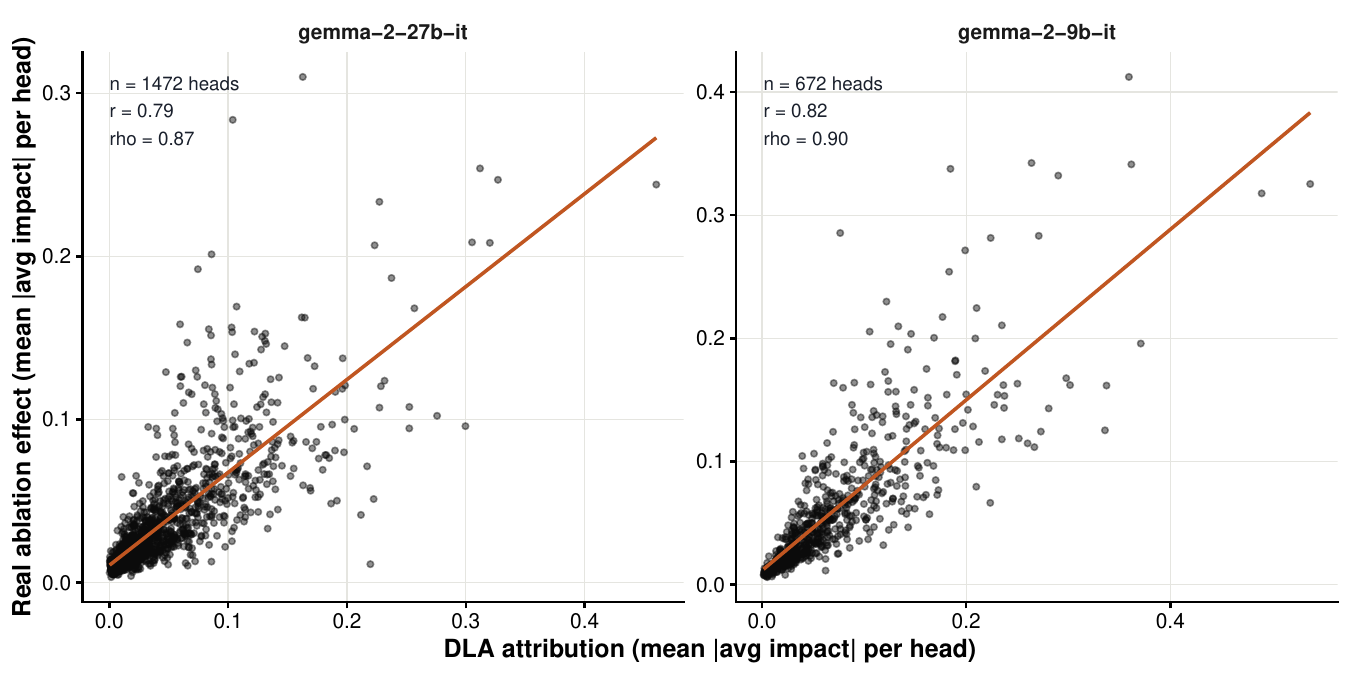}
    \caption{Correlation (Pearson $r$) for estimated causal impacts of the same heads when calculated with DLA and with direct ablation.}
\end{figure}

\pagebreak

\section{Content-effects stimuli and evaluation details}
\label{app:content-fx-stim}

\subsection{Stimuli overview}
\begin{table}[H]
\centering
\footnotesize
\caption{ prompts for the \texttt{north\_turns\_around} scenario (16 prompts).}
\begin{tabular}{@{}p{0.66\textwidth} p{0.12\textwidth} p{0.12\textwidth}@{}}
\toprule
Prompt & Correct & Incorrect \\
\midrule
The painting is North of Ali. Ali turns around. The painting is BLANK Ali. & North of & South of \\
The statue is North of Ali. Ali turns around. The statue is BLANK Ali. & North of & South of \\
The museum is North of Ali. Ali turns around. The museum is BLANK Ali. & North of & South of \\
The dog is North of Ali. Ali turns around. The dog is BLANK Ali. & North of & South of \\
The ocean is North of Ali. Ali turns around. The ocean is BLANK Ali. & North of & South of \\
The city is North of Ali. Ali turns around. The city is BLANK Ali. & North of & South of \\
The vase is North of Ali. Ali turns around. The vase is BLANK Ali. & North of & South of \\
The man is North of Ali. Ali turns around. The man is BLANK Ali. & North of & South of \\
The house is North of Ali. Ali turns around. The house is BLANK Ali. & North of & South of \\
The mountain is North of Ali. Ali turns around. The mountain is BLANK Ali. & North of & South of \\
The book is North of Ali. Ali turns around. The book is BLANK Ali. & North of & South of \\
The whiteboard is North of Ali. Ali turns around. The whiteboard is BLANK Ali. & North of & South of \\
The computer is North of Ali. Ali turns around. The computer is BLANK Ali. & North of & South of \\
The street is North of Ali. Ali turns around. The street is BLANK Ali. & North of & South of \\
The lake is North of Ali. Ali turns around. The lake is BLANK Ali. & North of & South of \\
The tree is North of Ali. Ali turns around. The tree is BLANK Ali. & North of & South of \\
\bottomrule
\end{tabular}
\end{table}

\begin{table}[H]
\centering
\footnotesize
\caption{ prompts for the \texttt{bumps\_breaks} scenario (16 prompts).}
\begin{tabular}{@{}p{0.66\textwidth} p{0.12\textwidth} p{0.12\textwidth}@{}}
\toprule
Prompt & Correct & Incorrect \\
\midrule
The glass is on the table. Ali bumps the lamp. The glass BLANK. & doesn't break & breaks \\
The vase is on the table. Ali bumps the lamp. The vase BLANK. & doesn't break & breaks \\
The cup is on the table. Ali bumps the lamp. The cup BLANK. & doesn't break & breaks \\
The plate is on the table. Ali bumps the lamp. The plate BLANK. & doesn't break & breaks \\
The bottle is on the table. Ali bumps the lamp. The bottle BLANK. & doesn't break & breaks \\
The phone is on the table. Ali bumps the lamp. The phone BLANK. & doesn't break & breaks \\
The mug is on the table. Ali bumps the lamp. The mug BLANK. & doesn't break & breaks \\
The jar is on the table. Ali bumps the lamp. The jar BLANK. & doesn't break & breaks \\
The ornament is on the table. Ali bumps the lamp. The ornament BLANK. & doesn't break & breaks \\
The tablet is on the table. Ali bumps the lamp. The tablet BLANK. & doesn't break & breaks \\
The portrait is on the table. Ali bumps the lamp. The portrait BLANK. & doesn't break & breaks \\
The sculpture is on the table. Ali bumps the lamp. The sculpture BLANK. & doesn't break & breaks \\
The candle is on the table. Ali bumps the lamp. The candle BLANK. & doesn't break & breaks \\
The clock is on the table. Ali bumps the lamp. The clock BLANK. & doesn't break & breaks \\
The mirror is on the table. Ali bumps the lamp. The mirror BLANK. & doesn't break & breaks \\
The television is on the table. Ali bumps the lamp. The television BLANK. & doesn't break & breaks \\
\bottomrule
\end{tabular}
\end{table}

\begin{table}[H]
\centering
\footnotesize
\caption{ prompts for the \texttt{empty\_fills} scenario (16 prompts).}
\begin{tabular}{@{}p{0.66\textwidth} p{0.12\textwidth} p{0.12\textwidth}@{}}
\toprule
Prompt & Correct & Incorrect \\
\midrule
The empty glass is on the table. Ali fills the sink. The glass is BLANK. & dry & wet \\
The empty cup is on the table. Ali fills the sink. The cup is BLANK. & dry & wet \\
The empty bowl is on the table. Ali fills the sink. The bowl is BLANK. & dry & wet \\
The empty pitcher is on the table. Ali fills the sink. The pitcher is BLANK. & dry & wet \\
The empty vase is on the table. Ali fills the sink. The vase is BLANK. & dry & wet \\
The empty bottle is on the table. Ali fills the sink. The bottle is BLANK. & dry & wet \\
The empty jug is on the table. Ali fills the sink. The jug is BLANK. & dry & wet \\
The empty mug is on the table. Ali fills the sink. The mug is BLANK. & dry & wet \\
The empty flask is on the table. Ali fills the sink. The flask is BLANK. & dry & wet \\
The empty can is on the table. Ali fills the sink. The can is BLANK. & dry & wet \\
The empty jar is on the table. Ali fills the sink. The jar is BLANK. & dry & wet \\
The empty tub is on the table. Ali fills the sink. The tub is BLANK. & dry & wet \\
The empty kettle is on the table. Ali fills the sink. The kettle is BLANK. & dry & wet \\
The empty thermos is on the table. Ali fills the sink. The thermos is BLANK. & dry & wet \\
The empty container is on the table. Ali fills the sink. The container is BLANK. & dry & wet \\
The empty box is on the table. Ali fills the sink. The box is BLANK. & dry & wet \\
\bottomrule
\end{tabular}
\end{table}

\begin{table}[H]
\centering
\footnotesize
\caption{ prompts for the \texttt{push\_no\_break} scenario (16 prompts).}
\begin{tabular}{@{}p{0.66\textwidth} p{0.12\textwidth} p{0.12\textwidth}@{}}
\toprule
Prompt & Correct & Incorrect \\
\midrule
The glass is on the floor. Ali bumps the table. The glass BLANK. & doesn't break & breaks \\
The vase is on the floor. Ali bumps the table. The vase BLANK. & doesn't break & breaks \\
The cup is on the floor. Ali bumps the table. The cup BLANK. & doesn't break & breaks \\
The plate is on the floor. Ali bumps the table. The plate BLANK. & doesn't break & breaks \\
The mug is on the floor. Ali bumps the table. The mug BLANK. & doesn't break & breaks \\
The phone is on the floor. Ali bumps the table. The phone BLANK. & doesn't break & breaks \\
The laptop is on the floor. Ali bumps the table. The laptop BLANK. & doesn't break & breaks \\
The clock is on the floor. Ali bumps the table. The clock BLANK. & doesn't break & breaks \\
The lamp is on the floor. Ali bumps the table. The lamp BLANK. & doesn't break & breaks \\
The sculpture is on the floor. Ali bumps the table. The sculpture BLANK. & doesn't break & breaks \\
The ornament is on the floor. Ali bumps the table. The ornament BLANK. & doesn't break & breaks \\
The television is on the floor. Ali bumps the table. The television BLANK. & doesn't break & breaks \\
The tablet is on the floor. Ali bumps the table. The tablet BLANK. & doesn't break & breaks \\
The mirror is on the floor. Ali bumps the table. The mirror BLANK. & doesn't break & breaks \\
The pot is on the floor. Ali bumps the table. The pot BLANK. & doesn't break & breaks \\
The jar is on the floor. Ali bumps the table. The jar BLANK. & doesn't break & breaks \\
\bottomrule
\end{tabular}
\end{table}

\begin{table}[H]
\centering
\footnotesize
\caption{ prompts for the \texttt{table\_stove} scenario (16 prompts).}
\begin{tabular}{@{}p{0.66\textwidth} p{0.12\textwidth} p{0.12\textwidth}@{}}
\toprule
Prompt & Correct & Incorrect \\
\midrule
The soup is on the table. Ali turns on the stove. The soup gets BLANK. & colder & warmer \\
The porridge is on the table. Ali turns on the stove. The porridge gets BLANK. & colder & warmer \\
The water is on the table. Ali turns on the stove. The water gets BLANK. & colder & warmer \\
The tea is on the table. Ali turns on the stove. The tea gets BLANK. & colder & warmer \\
The coffee is on the table. Ali turns on the stove. The coffee gets BLANK. & colder & warmer \\
The stew is on the table. Ali turns on the stove. The stew gets BLANK. & colder & warmer \\
The sauce is on the table. Ali turns on the stove. The sauce gets BLANK. & colder & warmer \\
The broth is on the table. Ali turns on the stove. The broth gets BLANK. & colder & warmer \\
The chili is on the table. Ali turns on the stove. The chili gets BLANK. & colder & warmer \\
The gumbo is on the table. Ali turns on the stove. The gumbo gets BLANK. & colder & warmer \\
The curry is on the table. Ali turns on the stove. The curry gets BLANK. & colder & warmer \\
The steak is on the table. Ali turns on the stove. The steak gets BLANK. & colder & warmer \\
The chicken is on the table. Ali turns on the stove. The chicken gets BLANK. & colder & warmer \\
The fish is on the table. Ali turns on the stove. The fish gets BLANK. & colder & warmer \\
The rice is on the table. Ali turns on the stove. The rice gets BLANK. & colder & warmer \\
The pasta is on the table. Ali turns on the stove. The pasta gets BLANK. & colder & warmer \\
\bottomrule
\end{tabular}
\end{table}

\begin{table}[H]
\centering
\footnotesize
\caption{ prompts for the \texttt{empty\_dry} scenario (16 prompts).}
\begin{tabular}{@{}p{0.66\textwidth} p{0.12\textwidth} p{0.12\textwidth}@{}}
\toprule
Prompt & Correct & Incorrect \\
\midrule
The empty glass is on the table. Ali knocks the glass over. The table is BLANK. & dry & wet \\
The empty mug is on the table. Ali knocks the mug over. The table is BLANK. & dry & wet \\
The empty wineglass is on the table. Ali knocks the wineglass over. The table is BLANK. & dry & wet \\
The empty cup is on the table. Ali knocks the cup over. The table is BLANK. & dry & wet \\
The empty can is on the table. Ali knocks the can over. The table is BLANK. & dry & wet \\
The empty bottle is on the table. Ali knocks the bottle over. The table is BLANK. & dry & wet \\
The empty carton is on the table. Ali knocks the carton over. The table is BLANK. & dry & wet \\
The empty thermos is on the table. Ali knocks the thermos over. The table is BLANK. & dry & wet \\
The empty jar is on the table. Ali knocks the jar over. The table is BLANK. & dry & wet \\
The empty pitcher is on the table. Ali knocks the pitcher over. The table is BLANK. & dry & wet \\
The empty bowl is on the table. Ali knocks the bowl over. The table is BLANK. & dry & wet \\
The empty vase is on the table. Ali knocks the vase over. The table is BLANK. & dry & wet \\
The empty flask is on the table. Ali knocks the flask over. The table is BLANK. & dry & wet \\
The empty jug is on the table. Ali knocks the jug over. The table is BLANK. & dry & wet \\
The empty pot is on the table. Ali knocks the pot over. The table is BLANK. & dry & wet \\
The empty teapot is on the table. Ali knocks the teapot over. The table is BLANK. & dry & wet \\
\bottomrule
\end{tabular}
\end{table}

\subsection{Abstraction and content contrast stimuli (minimal-difference evaluation)}
\label{app:abstract-content-min-diff-stim}

\begin{table}[H]
\centering
\footnotesize
\caption{As defined in Section \ref{sec:content-effects-min-diff}, we define content and abstraction contrasts for our minimal difference stimuli. }
\begin{tabular}{@{}l p{0.30\textwidth} p{0.30\textwidth} p{0.20\textwidth}@{}}
\toprule
Scenario & Critical item & Matched non-critical item & Content nouns \\
\midrule
\textit{bumps\_breaks} & The glass is on the table. Ali bumps the \textbf{lamp.} The glass BLANK. \newline \textit{(doesn't break / breaks)} & The glass is on the table. Ali bumps the \textbf{glass.} The glass BLANK. \newline \textit{(breaks / doesn't break)} & glass, vase, cup, plate, bottle, phone, \ldots \\
\addlinespace[3pt]
\textit{empty\_dry} & The \textbf{empty} glass is on the table. Ali knocks the glass over. The table is BLANK. \newline \textit{(dry / wet)} & The \textbf{full} glass is on the table. Ali knocks the glass over. The table is BLANK. \newline \textit{(wet / dry)} & glass, mug, wineglass, cup, can, bottle, \ldots \\
\addlinespace[3pt]
\textit{empty\_fills} & The empty glass is on the table. Ali fills the \textbf{sink.} The glass is BLANK. \newline \textit{(dry / wet)} & The empty glass is on the table. Ali fills the \textbf{glass.} The glass is BLANK. \newline \textit{(wet / dry)} & glass, cup, bowl, pitcher, vase, bottle, \ldots \\
\addlinespace[3pt]
\textit{north\_turns\_around} & The painting is \textbf{North} of Ali. Ali turns around. The painting is BLANK Ali. \newline \textit{(North of / South of)} & The painting is \textbf{in front} of Ali. Ali turns around. The painting is BLANK Ali. \newline \textit{(behind / in front of)} & painting, statue, museum, dog, ocean, city, \ldots \\
\addlinespace[3pt]
\textit{push\_no\_break} & The glass is on the floor. Ali bumps the \textbf{table.} The glass BLANK. \newline \textit{(doesn't break / breaks)} & The glass is on the floor. Ali bumps the \textbf{glass.} The glass BLANK. \newline \textit{(breaks / doesn't break)} & glass, vase, cup, plate, mug, phone, \ldots \\
\addlinespace[3pt]
\textit{table\_stove} & The soup is on the \textbf{table.} Ali turns on the stove. The soup gets BLANK. \newline \textit{(colder / warmer)} & The soup is on the \textbf{stove.} Ali turns on the stove. The soup gets BLANK. \newline \textit{(warmer / colder)} & soup, porridge, water, tea, coffee, stew, \ldots \\
\addlinespace[3pt]
\bottomrule
\end{tabular}
\end{table}

\newpage

\subsection{Partial correlations of attention head predictions (minimal-difference evaluation)}
\label{app:correlation-pred-min-diff}

\begin{figure}[H]
    \centering
    \includegraphics[width=1\linewidth]{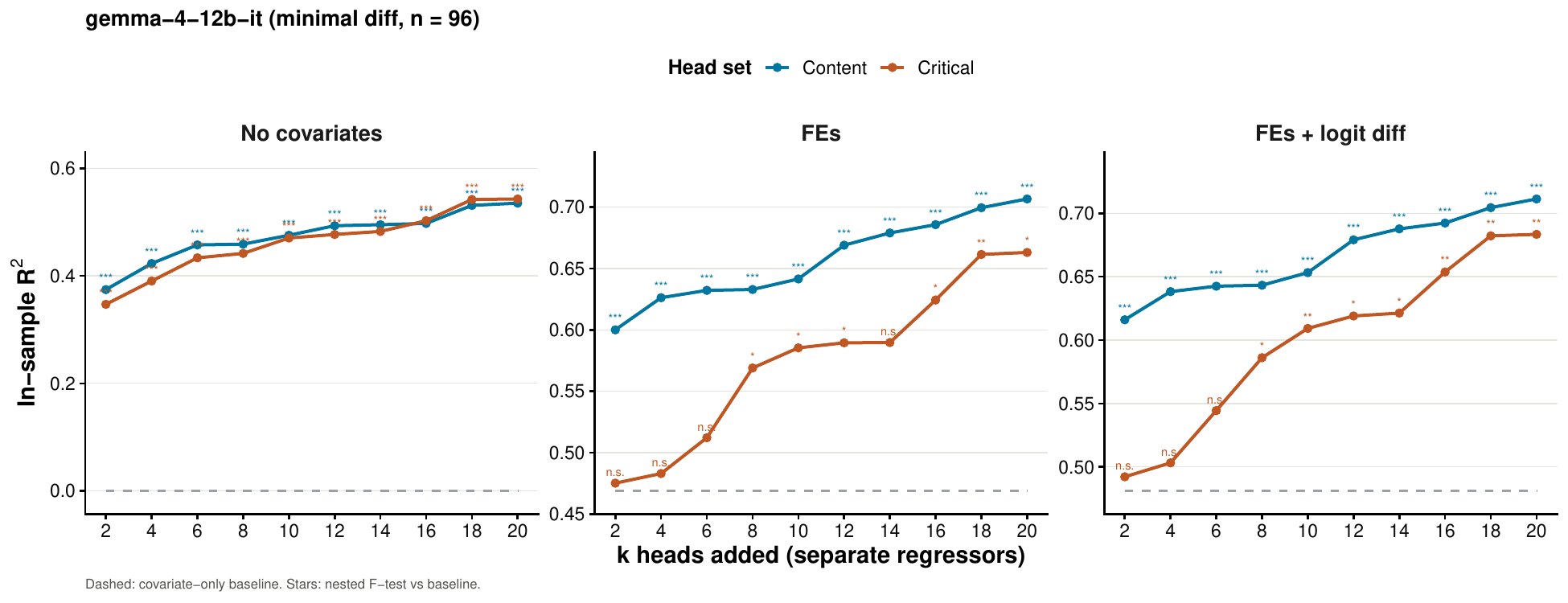}
    \caption{F-test nested model comparisons for \texttt{gemma-4-12b-it} content-sensitive and abstraction-sensitive attention heads, in-distribution $R^2$ at varying $k$. Asterisks indicate p-values for nested model comparisons against the base model. We compare nested model fit 1. with no covariates, 2. controlling for category and prompt type as fixed effects, and 3. controlling for FEs and distributional baselines.}
\end{figure}

\begin{figure}[H]
    \centering
    \includegraphics[width=0.8\linewidth]{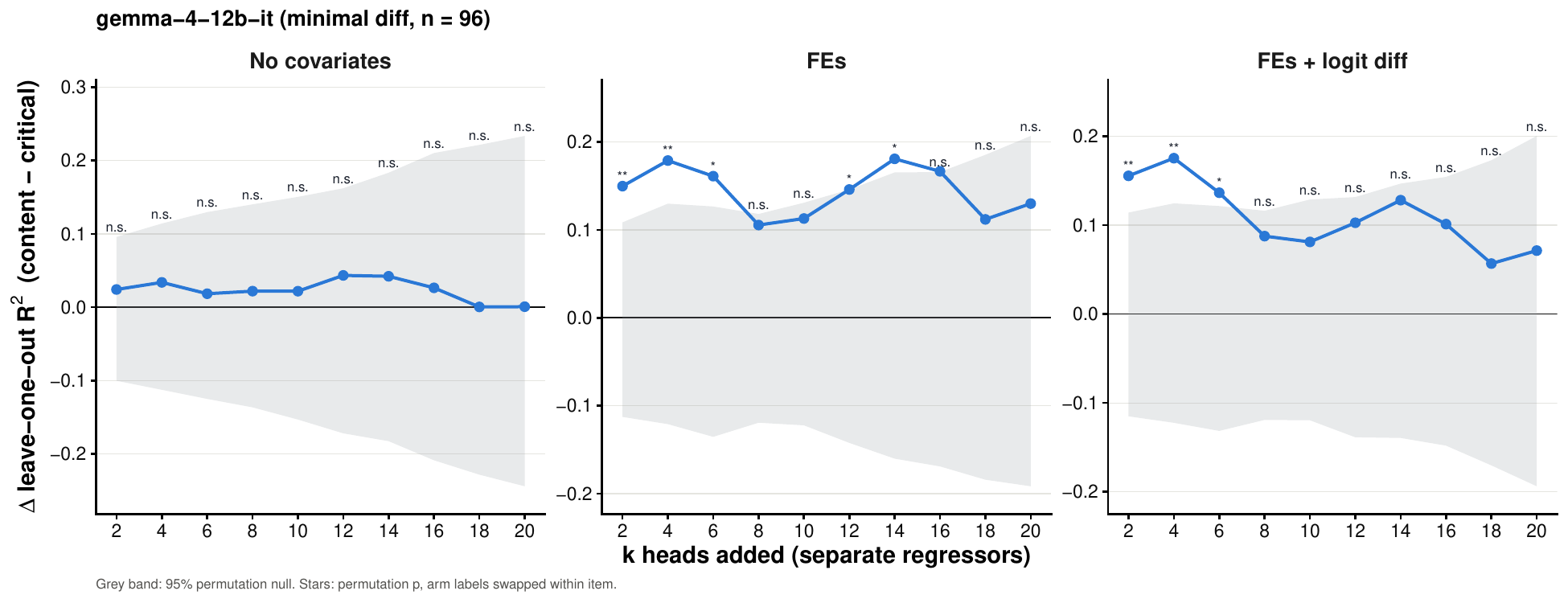}
    \caption{Permutation tests for \texttt{gemma-4-12b-it} content-sensitive and abstraction-sensitive attention heads, comparing out of distribution $R^2$ at varying $k$. Asterisks indicate p-values for nested model comparisons against the base model. We compare nested model fit 1. with no covariates, 2. controlling for category and prompt type as fixed effects, and 3. controlling for FEs and distributional baselines.}
\end{figure}
\pagebreak
\begin{figure}[H]
    \centering
    \includegraphics[width=0.8\linewidth]{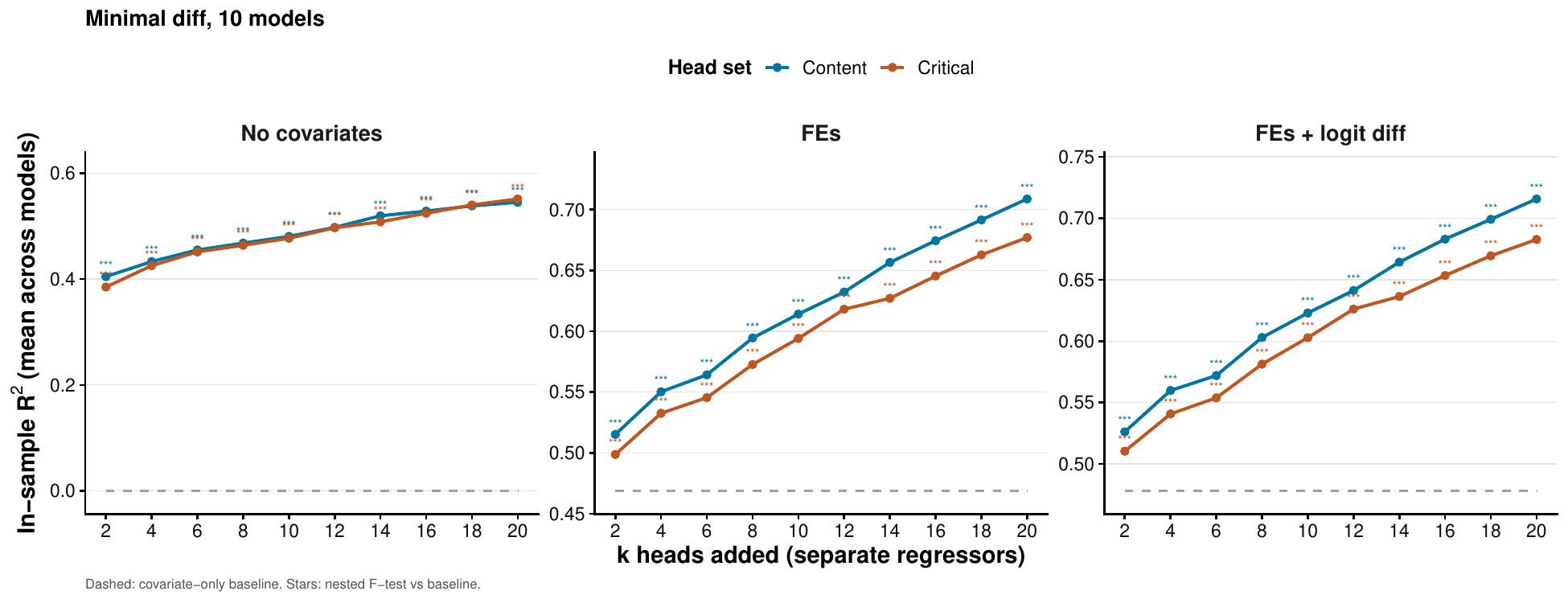}
    \caption{F-test nested model comparisons for all 15 LLMs with mechanistic results: content-sensitive and abstraction-sensitive attention heads, in-distribution $R^2$ at varying $k$. Asterisks indicate p-values for nested model comparisons against the base model. We compare nested model fit 1. with no covariates, 2. controlling for category and prompt type as fixed effects, and 3. controlling for FEs and distributional baselines.}
\end{figure}

\begin{figure}[H]
    \centering
    \includegraphics[width=1\linewidth]{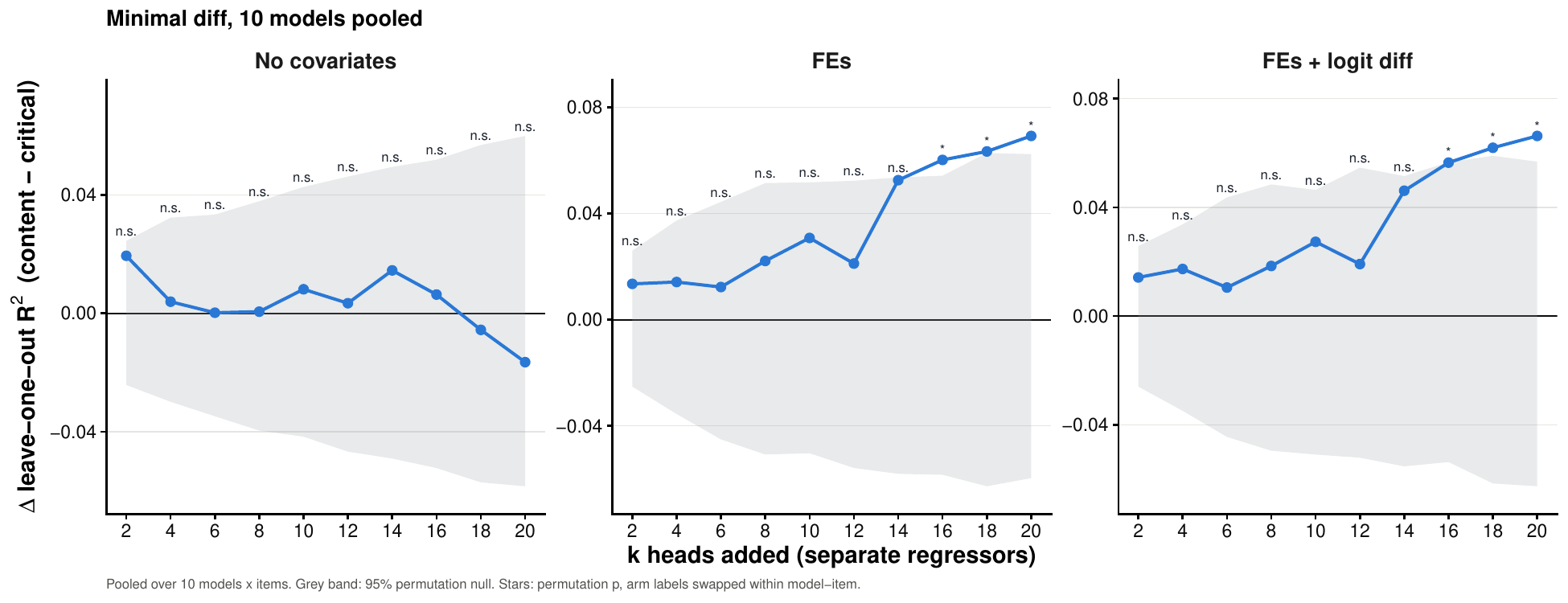}
    \caption{Permutation tests for all 15 LLM content-sensitive and abstraction-sensitive attention heads, comparing out of distribution $R^2$ at varying $k$. Asterisks indicate p-values for nested model comparisons against the base model. We compare nested model fit 1. with no covariates, 2. controlling for category and prompt type as fixed effects, and 3. controlling for FEs and distributional baselines.}
\end{figure}

\newpage
\subsection{Minimal-difference head distributions}
\label{app:min-diff-head-distributions}
Distributions of content-sensitive and abstraction-sensitive attention heads across model layers, for the top five aligned open-source models.

\begin{figure}[H]
    \centering
    \includegraphics[width=0.95\linewidth]{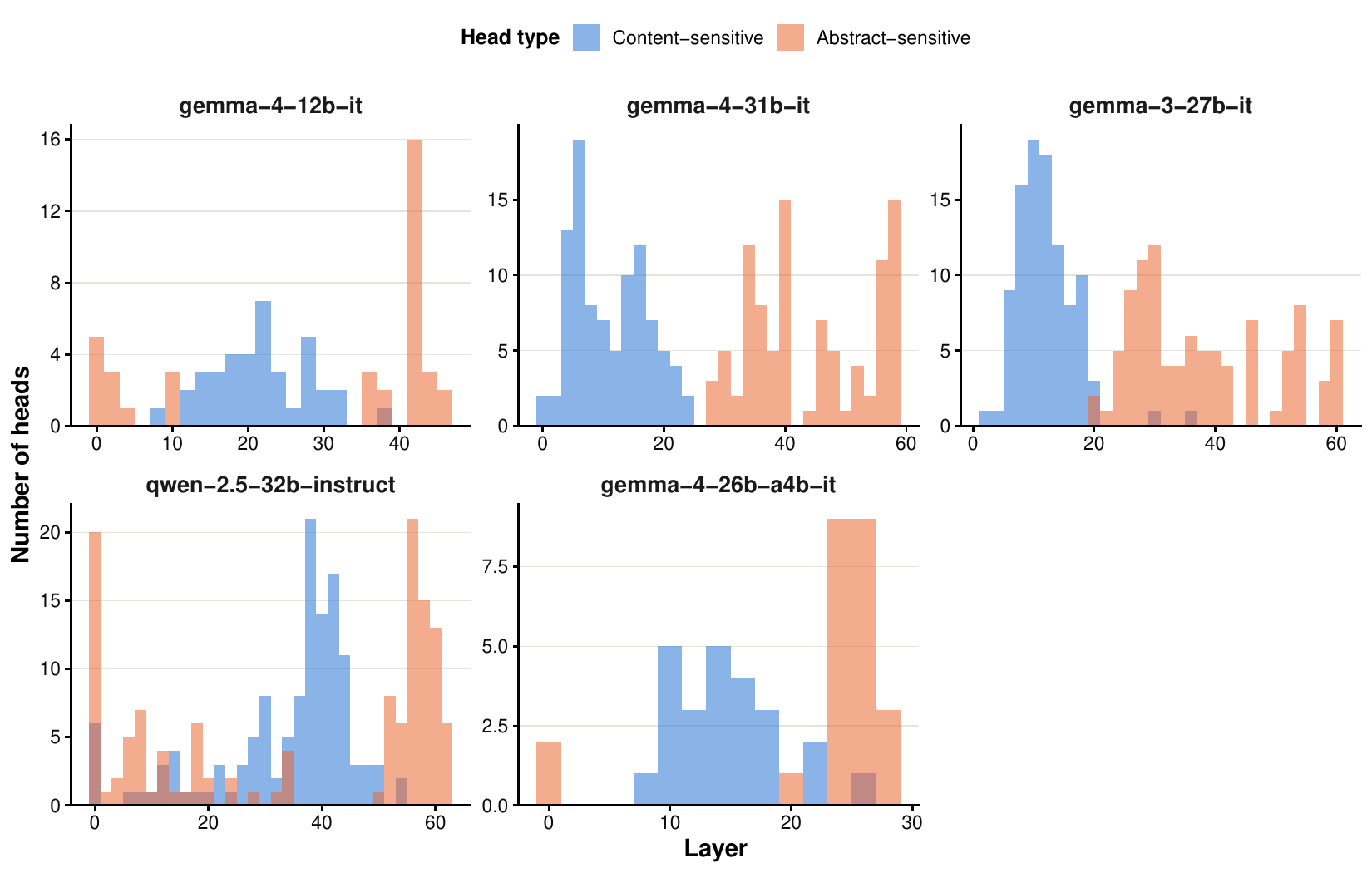}
    \caption{\textbf{Distribution of content-sensitive and abstraction-sensitive attention heads across layers.}}
    \label{fig:head_layer_distribution}
\end{figure}

\newpage

\subsection{Head ablation results (minimal-difference evaluation)}
\label{app:head-ablation-min-diff}

\subsubsection{Correlation effects}
\begin{table}[H]
\centering
\small
\caption{Summary of linear model for human-\texttt{gemma-4-12b} correlations.}
\label{tab:min_diff_corr_model}
\begin{tabular}{lccc}
\toprule
\textbf{Fixed effect} & \textbf{Est.} & \textbf{SE} & \textbf{\textit{t}} \\
\midrule
Intercept                        & 0.020    & 0.029 & 0.71     \\
Abstraction vs.\ content         & 0.452    & 0.083 & 5.48     \\
Content vs.\ random              & $-$0.608 & 0.064 & $-$9.51  \\
\bottomrule
\end{tabular}
\end{table}

\begin{figure}[H]
    \centering
    \includegraphics[width=0.6\linewidth]{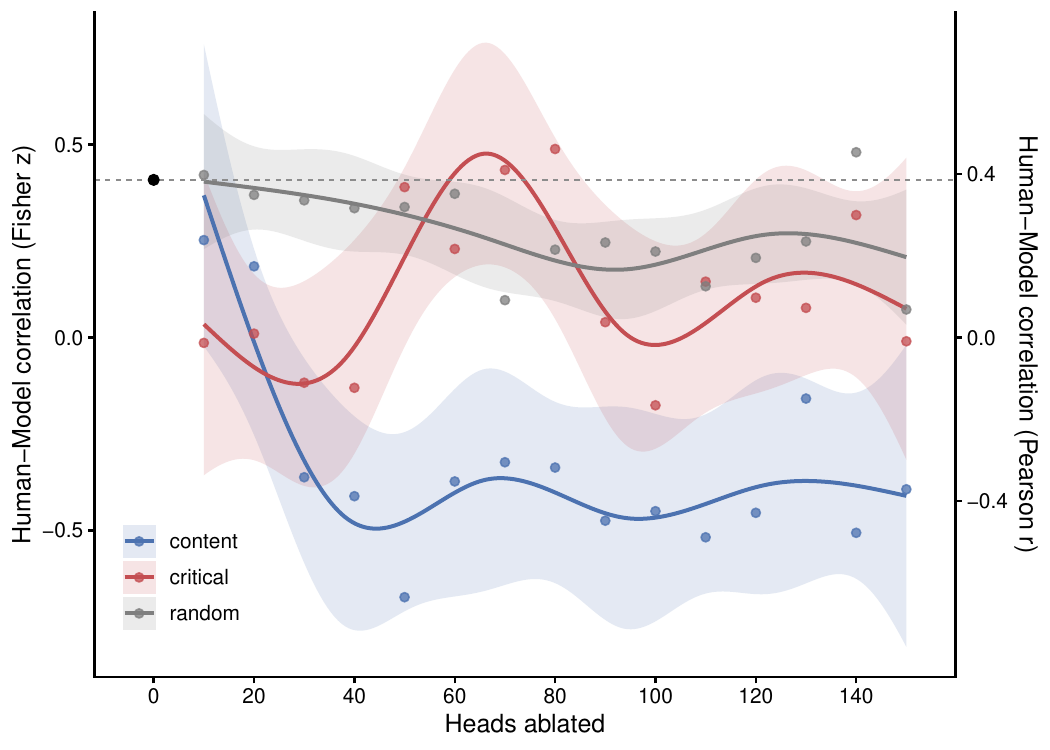}
    \caption{Effect of abstraction-sensitive, content-sensitive, and random head ablations on
    human--model correlation, minimal-difference evaluation.}
\end{figure}

\subsubsection{Accuracy effects}

\begin{figure}[H]
    \centering
    \includegraphics[width=0.6\linewidth]{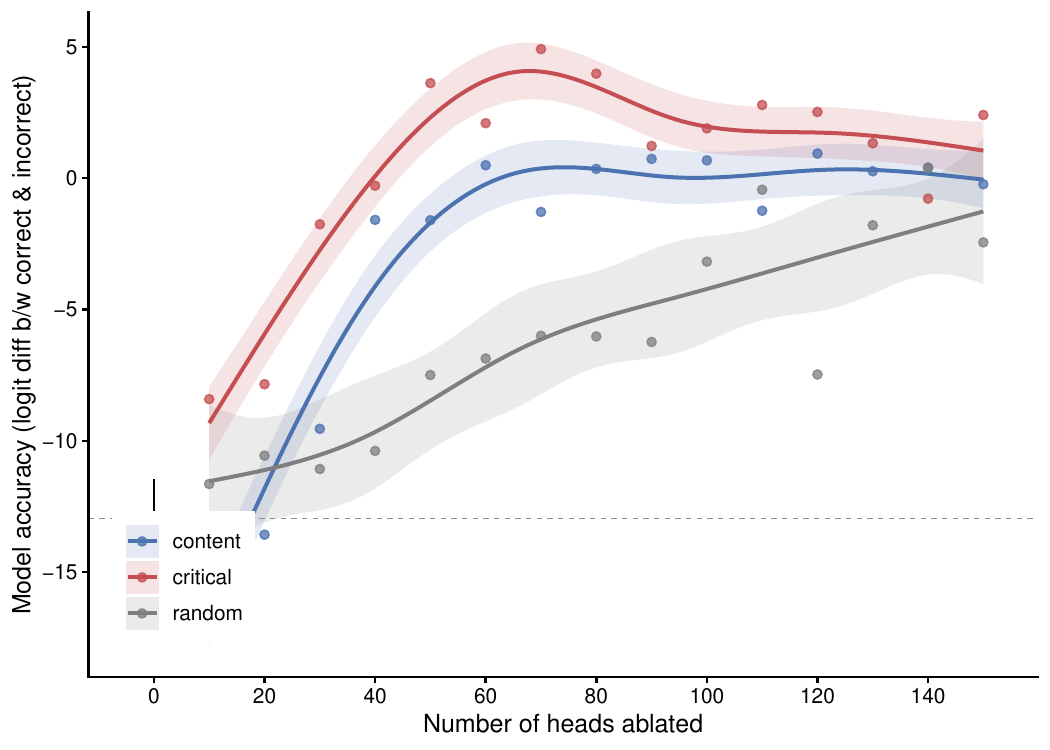}
    \caption{Effect of abstraction-sensitive, content-sensitive, and random heads on overall
    model \textit{accuracy}, minimal-difference evaluation.}
\end{figure}

\begin{table}[H]
\centering
\small
\caption{Summary of linear mixed effects model for \texttt{gemma-4-12b-it} accuracy (logit
difference), minimal-difference evaluation ($N = 10{,}080$ trials, 96 items, 6 scenarios).}
\label{tab:min_diff_acc_model}
\begin{tabular}{lccc}
\toprule
\textbf{Fixed effect} & \textbf{Est.} & \textbf{SE} & \textbf{\textit{t}} \\
\midrule
Intercept                        & $-$2.753 & 0.449 & $-$6.13 \\
Baseline logit difference ($z$)  & 0.228    & 0.105 & 2.18    \\
Abstraction vs.\ content         & 3.190    & 0.148 & 21.53   \\
Content vs.\ random              & 3.404    & 0.392 & 8.68    \\
\bottomrule
\end{tabular}
\end{table}

\pagebreak
\section{Additional causal reasoning evaluation results}

As a more thorough qualitative overview of our stimuli as well as human and model responses, we include (for each category) the five prompts with highest human and \texttt{gemma-3-27b-it} accuracy, highest human and lowest \texttt{gemma-3-27b-it} accuracy, lowest human and lowest \texttt{gemma-3-27b-it} accuracy, and highest \texttt{gemma-3-27b-it} and lowest human accuracy. \texttt{gemma-3-27b-it} accuracy for all prompts below is given as the logit difference measure. 

\begin{table}[H]
\centering
\small
\caption{ action\_follows (n=35)}
\begin{tabular}{@{}p{0.18\textwidth} p{0.49\textwidth} p{0.06\textwidth} p{0.08\textwidth} p{0.15\textwidth}@{}}
\toprule
Quadrant & Prompt & Hum. & Gemma $\Delta$ & A / B (correct) \\
\midrule
\multirow{5}{*}{\shortstack[l]{High human\\High gemma-3-27b}}
& [ctx] The book is on the shelf. Ali bumps the BLANK. The book falls off the shelf. & 1.000 & 32.00 & shelf / lamp (\checkmark) \\
& [tgt] The box is floating in the water. Ali puts a bowling ball in the box. The box BLANK. & 1.000 & 30.50 & sinks / floats (\checkmark) \\
& [tgt] The ball is on the table. Ali tilts the table. The ball BLANK. & 0.952 & 38.50 & rolls / stays still (\checkmark) \\
& [tgt] Ali is behind the wagon. Ali pushes on the wagon. The wagon BLANK. & 1.000 & 29.75 & moves / doesn't move (\checkmark) \\
& [tgt] The wagon is on top of the hill. Ali pushes the wagon. The wagon BLANK. & 0.950 & 34.25 & rolls / stays still (\checkmark) \\
\midrule
\multirow{5}{*}{\shortstack[l]{High gemma-3-27b\\Low human}}
& [ctx] There is a bowling ball on the right side of the seesaw. Ali BLANK the bowling ball up. The right side goes up. & 0.909 & 38.00 & picks / doesn't pick (\checkmark) \\
& [ctx] The wagon is on the BLANK of the hill. Ali pushes the wagon. The wagon rolls. & 0.762 & 24.50 & top / bottom (\checkmark) \\
& [tgt] The book is on the shelf. Ali bumps the shelf. The book BLANK the shelf. & 0.857 & 26.25 & falls off / stays on (\checkmark) \\
& [tgt] The book is on the table. Ali moves the book. The table BLANK. & 0.800 & 23.25 & doesn't move / moves (\checkmark) \\
& [tgt] The ball is on the table. Ali tilts the table. The ball BLANK. & 0.952 & 38.50 & rolls / stays still (\checkmark) \\
\midrule
\multirow{5}{*}{\shortstack[l]{Low gemma-3-27b\\High human}}
& [ctx] The ball is in the box. Ali moves the BLANK. The ball moves. & 1.000 & 8.00 & box / ball (\checkmark) \\
& [tgt] The ball is in the box. Ali moves the box. The ball BLANK. & 0.952 & $-1.50$ & moves / doesn't move ($\times$) \\
& [tgt] The ball is on top of the box. Ali pushes the box. The ball BLANK. & 1.000 & 15.75 & moves / doesn't move (\checkmark) \\
& [tgt] The tennis ball is next to the bowling ball. Ali throws the tennis ball. The tennis ball is BLANK the bowling ball. & 1.000 & 17.50 & far from / close to (\checkmark) \\
& [ctx] The tennis ball is next to the bowling ball. Ali BLANK the tennis ball. The tennis ball is far from the bowling ball. & 1.000 & 21.25 & throws / doesn't throw (\checkmark) \\
\midrule
\multirow{5}{*}{\shortstack[l]{Low human\\Low gemma-3-27b}}
& [ctx] The wine glass is on the BLANK. Ali bumps the wine glass. The wine spills on the floor. & 0.267 & $-33.00$ & floor / table ($\times$) \\
& [tgt] The wine glass is on the table. The wine glass falls over. The wine spills on the BLANK. & 0.455 & $-17.50$ & table / floor ($\times$) \\
& [tgt] The book is on the table. Ali moves the table. The book BLANK. & 0.727 & $-13.00$ & moves / doesn't move ($\times$) \\
& [ctx] Ali is BLANK the wagon. Ali pushes on the wagon. The wagon moves. & 0.733 & $-4.00$ & behind / in front of (--) \\
& [ctx] Ali is BLANK the wagon. Ali pulls on the wagon. The wagon moves. & 0.714 & $-1.75$ & in front of / behind (--) \\
\bottomrule
\end{tabular}
\end{table}

\begin{table}[H]
\centering
\small
\caption{ action\_nonfollows (n=44)}
\begin{tabular}{@{}p{0.18\textwidth} p{0.49\textwidth} p{0.06\textwidth} p{0.08\textwidth} p{0.15\textwidth}@{}}
\toprule
Quadrant & Prompt & Hum. & Gemma $\Delta$ & A / B (correct) \\
\midrule
\multirow{5}{*}{\shortstack[l]{High human\\High gemma-3-27b}}
& [ctx] The box is BLANK the tunnel. Ali goes under the tunnel. The box is above the tunnel. & 0.938 & 21.25 & above / below (\checkmark) \\
& [ctx] The box is in the bin. Ali moves the BLANK. The bin doesn't move. & 0.875 & 26.50 & box / bin (\checkmark) \\
& [ctx] The box is floating in the water. Ali puts a BLANK in the box. The box floats. & 0.955 & 11.00 & tennis ball / bowling ball (\checkmark) \\
& [ctx] There is a bowling ball on the right side of the seesaw. Ali puts a BLANK on the left side of the seesaw. The left side stays up. & 0.875 & 15.25 & tennis ball / bowling ball (\checkmark) \\
& [tgt] The box is below the bridge. Ali goes over the bridge. The box is BLANK the bridge. & 0.905 & 10.50 & below / above (\checkmark) \\
\midrule
\multirow{5}{*}{\shortstack[l]{High gemma-3-27b\\Low human}}
& [ctx] Ali is BLANK the wagon. Ali pulls on the wagon. The wagon doesn't move. & 0.533 & 12.00 & in / outside (\checkmark) \\
& [tgt] The book is under the table. Ali moves the book. The table BLANK. & 0.625 & 17.25 & doesn't move / moves (\checkmark) \\
& [tgt] The ball is resting on top of the box. Ali pushes the box forward. The ball is BLANK the box. & 0.650 & 17.25 & on top of / below (\checkmark) \\
& [tgt] The bowling ball is close to Ali. Ali throws the bowling ball. The bowling ball is BLANK Ali. & 0.095 & $-3.25$ & close to / far from ($\times$) \\
& [ctx] The ball is BLANK the box. The ball moves right. The ball is higher than the box. & 0.714 & 26.00 & above / under (\checkmark) \\
\midrule
\multirow{5}{*}{\shortstack[l]{Low gemma-3-27b\\High human}}
& [tgt] There is a bowling ball on the right side of the seesaw. Ali puts a tennis ball on the left side of the seesaw. The left side BLANK. & 0.875 & $-16.50$ & stays up / goes down ($\times$) \\
& [ctx] The ball is BLANK the box. Ali pushes the box. The ball doesn't move. & 0.905 & $-9.00$ & far from / touching ($\times$) \\
& [tgt] The ball is under the box. The ball moves right. The ball is BLANK the box. & 0.938 & $-6.75$ & lower than / higher than (--) \\
& [tgt] The ball is above the box. The ball moves right. The ball is BLANK the box. & 1.000 & $-3.00$ & higher than / lower than ($\times$) \\
& [tgt] The box is floating in the water. Ali puts a tennis ball in the box. The box BLANK. & 0.727 & $-21.25$ & floats / sinks ($\times$) \\
\midrule
\multirow{5}{*}{\shortstack[l]{Low human\\Low gemma-3-27b}}
& [tgt] Ali is in the wagon. Ali pushes on the wagon. The wagon BLANK. & 0.250 & $-28.25$ & doesn't move / moves ($\times$) \\
& [ctx] Ali is BLANK. Ali turns around. Ali is behind the statue. & 0.333 & $-22.50$ & behind the statue / in front of the statue ($\times$) \\
& [ctx] The ball is close to Ali. Ali throws the BLANK. The ball is close to Ali. & 0.267 & $-21.25$ & box / ball ($\times$) \\
& [ctx] The wine glass is on the table. Ali bumps the BLANK. The wine doesn't spill. & 0.524 & $-26.50$ & lamp / table ($\times$) \\
& [tgt] Ali is behind the statue. Ali turns around. Ali is BLANK. & 0.312 & $-19.50$ & behind the statue / in front of the statue ($\times$) \\
\bottomrule
\end{tabular}
\end{table}

\begin{table}[H]
\centering
\small
\caption{ egocentric\_distant (n=40)}
\begin{tabular}{@{}p{0.18\textwidth} p{0.49\textwidth} p{0.06\textwidth} p{0.08\textwidth} p{0.15\textwidth}@{}}
\toprule
Quadrant & Prompt & Hum. & Gemma $\Delta$ & A / B (correct) \\
\midrule
\multirow{5}{*}{\shortstack[l]{High human\\High gemma-3-27b}}
& [tgt] Ali is facing away from Milwaukee. Ali turns around. Ali is facing BLANK Milwaukee. & 1.000 & 29.00 & towards / away from (\checkmark) \\
& [tgt] Ali is facing towards Milwaukee. Ali turns around. Ali is facing BLANK Milwaukee. & 0.938 & 21.75 & away from / towards (\checkmark) \\
& [ctx] Chicago is BLANK Ali. Ali turns around. Chicago is in front of Ali. & 0.933 & 16.50 & behind / in front of (\checkmark) \\
& [tgt] Chicago is in front of Ali. Ali turns around. Chicago is BLANK Ali. & 0.875 & 23.50 & behind / in front of (\checkmark) \\
& [tgt] Chicago is behind Ali. Ali turns around. Chicago is BLANK Ali. & 0.909 & 15.50 & in front of / behind (\checkmark) \\
\midrule
\multirow{5}{*}{\shortstack[l]{High gemma-3-27b\\Low human}}
& [tgt] Ali is facing away from Chicago. Ali turns around. Chicago is BLANK Ali. & 0.636 & 17.50 & in front of / behind (\checkmark) \\
& [tgt] Ali is facing away from Milwaukee. Ali turns 180 degrees. Ali is facing BLANK Milwaukee. & 0.714 & 28.75 & towards / away from (\checkmark) \\
& [tgt] Chicago is left of Ali. Ali turns around. Chicago is BLANK Ali. & 0.700 & 22.75 & right of / left of (\checkmark) \\
& [ctx] Chicago is BLANK Ali. Ali turns around. Chicago is right of Ali. & 0.625 & 12.00 & left of / right of (\checkmark) \\
& [tgt] Chicago is right of Ali. Ali turns around. Chicago is BLANK Ali. & 0.714 & 18.50 & left of / right of (\checkmark) \\
\midrule
\multirow{5}{*}{\shortstack[l]{Low gemma-3-27b\\High human}}
& [tgt] Chicago is to Ali's left. Ali turns right. Ali is facing BLANK Chicago. & 0.938 & $-14.00$ & away from / towards ($\times$) \\
& [tgt] Milwaukee is to Ali's right. Ali turns left. Ali is facing BLANK Milwaukee. & 0.905 & $-11.00$ & away from / towards ($\times$) \\
& [ctx] Ali is facing BLANK Chicago. Ali turns right. Chicago is right of Ali. & 0.909 & $-9.00$ & away from / towards ($\times$) \\
& [ctx] Chicago is BLANK Ali. Ali turns around. Chicago is behind Ali. & 0.818 & $-13.00$ & in front of / behind ($\times$) \\
& [tgt] Chicago is in front of Ali. Ali turns left. Chicago is BLANK Ali. & 0.850 & $-5.50$ & right of / left of ($\times$) \\
\midrule
\multirow{5}{*}{\shortstack[l]{Low human\\Low gemma-3-27b}}
& [tgt] Ali is facing away from Chicago. Ali turns left. Chicago is BLANK Ali. & 0.600 & $-14.50$ & left of / right of ($\times$) \\
& [ctx] Ali is facing BLANK Chicago. Ali turns left. Chicago is left of Ali. & 0.500 & $-9.50$ & away from / towards ($\times$) \\
& [ctx] Chicago is BLANK Ali. Ali turns left. Chicago is right of Ali. & 0.562 & $-10.50$ & in front of / behind ($\times$) \\
& [tgt] Chicago is behind Ali. Ali turns right. Chicago is BLANK Ali. & 0.600 & $-8.50$ & right of / left of ($\times$) \\
& [ctx] Chicago is BLANK Ali. Ali turns right. Chicago is left of Ali. & 0.636 & $-10.00$ & in front of / behind ($\times$) \\
\bottomrule
\end{tabular}
\end{table}

\begin{table}[H]
\centering
\small
\caption{ egocentric\_near (n=34)}
\begin{tabular}{@{}p{0.18\textwidth} p{0.49\textwidth} p{0.06\textwidth} p{0.08\textwidth} p{0.15\textwidth}@{}}
\toprule
Quadrant & Prompt & Hum. & Gemma $\Delta$ & A / B (correct) \\
\midrule
\multirow{5}{*}{\shortstack[l]{High human\\High gemma-3-27b}}
& [tgt] The painting is in front of Ali. Ali turns around. The painting is BLANK Ali. & 0.909 & 25.50 & behind / in front of (\checkmark) \\
& [tgt] The painting is behind Ali. Ali turns around. The painting is BLANK Ali. & 0.952 & 18.00 & in front of / behind (\checkmark) \\
& [tgt] The painting is behind Ali. Ali turns right. The painting is BLANK Ali. & 0.900 & 14.00 & right of / left of (\checkmark) \\
& [ctx] Ali is BLANK the painting. Ali turns around. The painting is behind Ali. & 0.857 & 14.50 & in front of / behind (\checkmark) \\
& [tgt] Ali is in front of the painting. Ali turns around. The painting is BLANK Ali. & 0.800 & 22.50 & behind / in front of (\checkmark) \\
\midrule
\multirow{5}{*}{\shortstack[l]{High gemma-3-27b\\Low human}}
& [tgt] The painting is left of Ali. Ali turns around. The painting is BLANK Ali. & 0.500 & 25.25 & right of / left of (\checkmark) \\
& [ctx] The painting is to Ali's BLANK. The painting is moved left. The painting is to Ali's right. & 0.200 & 14.25 & left / right (\checkmark) \\
& [tgt] The painting is right of Ali. Ali turns around. The painting is BLANK Ali. & 0.700 & 24.50 & left of / right of (\checkmark) \\
& [tgt] The painting is right of Ali. Ali turns left. The painting is BLANK Ali. & 0.636 & 17.00 & behind / in front of (\checkmark) \\
& [tgt] The painting is to Ali's left. The painting is moved left. The painting is to Ali's BLANK. & 0.300 & 0.50 & right / left (\checkmark) \\
\midrule
\multirow{5}{*}{\shortstack[l]{Low gemma-3-27b\\High human}}
& [tgt] The painting is in front of Ali. Ali turns left. The painting is BLANK Ali. & 0.900 & $-6.50$ & right of / left of ($\times$) \\
& [tgt] The painting is left of Ali. Ali turns left. The painting is BLANK Ali. & 0.875 & $-9.50$ & in front of / behind ($\times$) \\
& [tgt] The painting is to Ali's right. The painting is moved right. The painting is to Ali's BLANK. & 0.750 & $-15.25$ & right / left ($\times$) \\
& [tgt] The painting is left of Ali. Ali turns right. The painting is BLANK Ali. & 0.850 & $-2.50$ & behind / in front of ($\times$) \\
& [ctx] The painting is BLANK of Ali. Ali turns left. The painting is behind Ali. & 0.810 & $-3.50$ & right / left ($\times$) \\
\midrule
\multirow{5}{*}{\shortstack[l]{Low human\\Low gemma-3-27b}}
& [tgt] Ali is behind the painting. Ali turns around. Ali is BLANK the painting. & 0.250 & $-20.25$ & behind / in front of ($\times$) \\
& [ctx] The painting is to Ali's BLANK. The painting is moved right. The painting is to Ali's right. & 0.545 & $-17.00$ & right / left ($\times$) \\
& [ctx] The painting is BLANK Ali. Ali turns left. The painting is left of Ali. & 0.500 & $-14.00$ & behind / in front of ($\times$) \\
& [tgt] Ali is right of the painting. Ali turns around. Ali is BLANK the painting. & 0.571 & $-15.50$ & right of / left of ($\times$) \\
& [ctx] Ali is BLANK the painting. Ali turns around. Ali is right of the painting. & 0.381 & $-12.50$ & right of / left of ($\times$) \\
\bottomrule
\end{tabular}
\end{table}

\begin{table}[H]
\centering
\small
\caption{ geocentric\_distant (n=40)}
\begin{tabular}{@{}p{0.18\textwidth} p{0.49\textwidth} p{0.06\textwidth} p{0.08\textwidth} p{0.15\textwidth}@{}}
\toprule
Quadrant & Prompt & Hum. & Gemma $\Delta$ & A / B (correct) \\
\midrule
\multirow{5}{*}{\shortstack[l]{High human\\High gemma-3-27b}}
& [tgt] Ali is directly South of Milwaukee. Ali walks West. Ali is BLANK Milwaukee. & 0.905 & 16.25 & Southwest / Northeast (\checkmark) \\
& [tgt] Milwaukee is West of Ali. Ali walks West. Ali is getting BLANK Milwaukee. & 0.818 & 20.75 & closer to / further from (\checkmark) \\
& [ctx] Milwaukee is BLANK of Ali. Ali walks East. Ali is getting further from Milwaukee. & 0.905 & 8.50 & West / East (\checkmark) \\
& [tgt] Ali is South of Milwaukee. Ali walks South. Ali is getting BLANK Milwaukee. & 0.818 & 1.00 & further from / closer to (\checkmark) \\
& [tgt] Ali is directly North of Milwaukee. Ali walks East. Ali is BLANK Milwaukee. & 0.700 & 9.50 & Northeast / Southwest (--) \\
\midrule
\multirow{5}{*}{\shortstack[l]{High gemma-3-27b\\Low human}}
& [tgt] Ali is South of Milwaukee. Ali walks North. Ali is getting BLANK Milwaukee. & 0.562 & 35.25 & closer to / further from (\checkmark) \\
& [ctx] Milwaukee is BLANK Ali. Ali turns left. Milwaukee is West of Ali. & 0.375 & $-4.00$ & West of / South of ($\times$) \\
& [tgt] Ali is East of Milwaukee. Ali walks East. Ali is getting BLANK Milwaukee. & 0.550 & 16.00 & further from / closer to (\checkmark) \\
& [tgt] Milwaukee is West of Ali. Ali turns left. Milwaukee is BLANK Ali. & 0.438 & $-2.00$ & West of / South of (--) \\
& [tgt] Ali is East of Milwaukee. Ali walks West. Ali is getting BLANK Milwaukee. & 0.619 & 27.00 & closer to / further from (\checkmark) \\
\midrule
\multirow{5}{*}{\shortstack[l]{Low gemma-3-27b\\High human}}
& [tgt] Milwaukee is West of Ali. Ali walks East. Ali is getting BLANK Milwaukee. & 0.875 & $-21.25$ & further from / closer to ($\times$) \\
& [tgt] Milwaukee is South of Ali. Ali turns around. Milwaukee is BLANK Ali. & 0.636 & $-23.00$ & South of / North of ($\times$) \\
& [ctx] Milwaukee is BLANK of Ali. Ali walks West. Ali is getting closer to Milwaukee. & 0.750 & $-16.50$ & West / East ($\times$) \\
& [ctx] Ali is BLANK of Milwaukee. Ali walks South. Ali is getting further from Milwaukee. & 0.750 & $-15.00$ & South / North ($\times$) \\
& [ctx] Ali is directly BLANK Milwaukee. Ali walks East. Ali is Northeast of Milwaukee. & 0.867 & $-6.50$ & North of / South of ($\times$) \\
\midrule
\multirow{5}{*}{\shortstack[l]{Low human\\Low gemma-3-27b}}
& [ctx] Milwaukee is BLANK Ali. Ali turns around. Milwaukee is East of Ali. & 0.182 & $-21.50$ & East of / West of ($\times$) \\
& [ctx] Milwaukee is BLANK Ali. Ali turns around. Milwaukee is West of Ali. & 0.267 & $-21.25$ & West of / East of ($\times$) \\
& [ctx] Milwaukee is BLANK Ali. Ali turns around. Milwaukee is South of Ali. & 0.312 & $-21.25$ & South of / North of ($\times$) \\
& [tgt] Milwaukee is North of Ali. Ali turns around. Milwaukee is BLANK Ali. & 0.500 & $-24.75$ & North of / South of ($\times$) \\
& [tgt] Milwaukee is East of Ali. Ali turns around. Milwaukee is BLANK Ali. & 0.333 & $-19.75$ & East of / West of ($\times$) \\
\bottomrule
\end{tabular}
\end{table}

\begin{table}[H]
\centering
\small
\caption{ geocentric\_near (n=54)}
\begin{tabular}{@{}p{0.18\textwidth} p{0.49\textwidth} p{0.06\textwidth} p{0.08\textwidth} p{0.15\textwidth}@{}}
\toprule
Quadrant & Prompt & Hum. & Gemma $\Delta$ & A / B (correct) \\
\midrule
\multirow{5}{*}{\shortstack[l]{High human\\High gemma-3-27b}}
& [tgt] Mark is facing West. Mark turns 90 degrees to his right. Mark is facing BLANK. & 0.800 & 19.25 & North / South (\checkmark) \\
& [ctx] Mark is facing BLANK. Mark turns 90 degrees to his right. Mark is facing South. & 0.800 & 5.00 & East / West (--) \\
& [ctx] Mark is facing BLANK. Mark turns 90 degrees to his right. Mark is facing East. & 0.864 & 0.00 & North / South (--) \\
& [tgt] Mark is facing East. Mark turns 90 degrees to his right. Mark is facing BLANK. & 0.727 & 26.00 & South / North (\checkmark) \\
& [ctx] Mark is facing BLANK. Mark turns 90 degrees to his right. Mark is facing North. & 0.750 & 0.00 & West / East (--) \\
\midrule
\multirow{5}{*}{\shortstack[l]{High gemma-3-27b\\Low human}}
& [ctx] The painting is BLANK Ali. Ali turns left. The painting is South of Ali. & 0.267 & $-5.50$ & South of / East of (--) \\
& [ctx] The painting is BLANK Ali. Ali turns left. The painting is West of Ali. & 0.188 & $-7.50$ & West of / South of (--) \\
& [ctx] The painting is BLANK Ali. Ali turns right. The painting is South of Ali. & 0.267 & $-8.00$ & South of / West of (--) \\
& [ctx] The painting is BLANK Ali. Ali turns left. The painting is North of Ali. & 0.333 & $-7.00$ & North of / West of (--) \\
& [ctx] The painting is BLANK Ali. Ali turns around. The painting is South of Ali. & 0.312 & $-11.25$ & South of / North of ($\times$) \\
\midrule
\multirow{5}{*}{\shortstack[l]{Low gemma-3-27b\\High human}}
& [ctx] The statue is BLANK Ali. Ali walks South. The statue is East of Ali. & 0.733 & $-23.00$ & East of / North of ($\times$) \\
& [tgt] The statue is South of Ali. Ali walks East. The statue is BLANK of Ali. & 0.650 & $-28.00$ & South of / West of ($\times$) \\
& [tgt] The box is North of Ali. Ali turns around. The box is BLANK Ali. & 0.650 & $-25.25$ & North of / South of ($\times$) \\
& [ctx] The statue is BLANK Ali. Ali walks North. The statue is East of Ali. & 0.682 & $-21.50$ & East of / South of ($\times$) \\
& [tgt] The statue is West of Ali. Ali walks South. The statue is BLANK of Ali. & 0.700 & $-20.00$ & West of / North of ($\times$) \\
\midrule
\multirow{5}{*}{\shortstack[l]{Low human\\Low gemma-3-27b}}
& [ctx] The box is BLANK Ali. Ali turns around. The box is East of Ali. & 0.182 & $-22.75$ & East of / West of ($\times$) \\
& [tgt] The painting is East of Ali. Ali turns around. The painting is BLANK Ali. & 0.312 & $-23.25$ & East of / West of ($\times$) \\
& [tgt] The box is East of Ali. Ali turns around. The box is BLANK Ali. & 0.364 & $-25.00$ & East of / West of ($\times$) \\
& [tgt] The box is South of Ali. Ali turns around. The box is BLANK Ali. & 0.375 & $-26.00$ & South of / North of ($\times$) \\
& [ctx] The box is BLANK Ali. Ali turns around. The box is South of Ali. & 0.312 & $-21.25$ & South of / North of ($\times$) \\
\bottomrule
\end{tabular}
\end{table}

\begin{table}[H]
\centering
\small
\caption{ relative\_pos (n=38)}
\begin{tabular}{@{}p{0.18\textwidth} p{0.49\textwidth} p{0.06\textwidth} p{0.08\textwidth} p{0.15\textwidth}@{}}
\toprule
Quadrant & Prompt & Hum. & Gemma $\Delta$ & A / B (correct) \\
\midrule
\multirow{5}{*}{\shortstack[l]{High human\\High gemma-3-27b}}
& [tgt] The box is next to the desk. Ali stacks the box on the desk. The box is BLANK the desk. & 1.000 & 28.75 & above / below (\checkmark) \\
& [tgt] The tennis ball is at the top of the ramp and the bowling ball is at the bottom of the ramp. Ali rolls the tennis ball down the ramp. The tennis ball is getting BLANK the bowling ball. & 0.938 & 39.75 & closer to / further from (\checkmark) \\
& [ctx] The ball is next to the box. Ali puts the BLANK. The ball is smaller than the box. & 1.000 & 22.75 & ball inside the box / box inside the ball (\checkmark) \\
& [ctx] The glass is on the floor. Ali sets the glass on the BLANK. The glass is above the table. & 0.955 & 23.50 & table / floor (\checkmark) \\
& [ctx] The tennis ball is BLANK the bowling ball. Ali rolls the bowling ball. The tennis ball is closer to the bowling ball. & 0.909 & 37.50 & far from / touching (\checkmark) \\
\midrule
\multirow{5}{*}{\shortstack[l]{High gemma-3-27b\\Low human}}
& [tgt] The ball is inside the box. Ali picks up the ball. The ball is BLANK the box. & 0.636 & 28.00 & not touching / touching (\checkmark) \\
& [ctx] Ali is holding the tennis ball and the ping pong ball. Ali BLANK the ping pong ball. The ping pong ball is further from the tennis ball. & 0.667 & 22.50 & drops / picks up (\checkmark) \\
& [tgt] The ball and the box are on top of the ramp. The ball rolls down the ramp. The ball is BLANK the box. & 0.750 & 23.25 & lower than / higher than (\checkmark) \\
& [tgt] The tennis ball is at the top of the ramp and the bowling ball is at the bottom of the ramp. Ali rolls the bowling ball up the ramp. The bowling ball is getting BLANK the tennis ball. & 0.818 & 34.25 & closer to / further from (\checkmark) \\
& [tgt] Ali is holding the tennis ball and the ping pong ball is on the floor. Ali picks up the ping pong ball. The ping pong ball is BLANK the tennis ball. & 0.667 & 20.00 & closer to / further from (\checkmark) \\
\midrule
\multirow{5}{*}{\shortstack[l]{Low gemma-3-27b\\High human}}
& [tgt] The box is in front of the ball. Ali pushes the box into the ball. The ball is BLANK the box. & 1.000 & 7.00 & touching / not touching (\checkmark) \\
& [tgt] The ball is next to the box. Ali puts the ball inside the box. The ball is BLANK the box. & 0.952 & $-0.75$ & smaller than / larger than (--) \\
& [ctx] The box is next to the desk. Ali stacks the BLANK. The box is above the desk. & 1.000 & 13.00 & box on the desk / desk on the box (\checkmark) \\
& [ctx] The ball and the box are on the floor. Ali picks up the BLANK. The ball is higher than the box. & 1.000 & 18.75 & ball / box (\checkmark) \\
& [ctx] The box is next to the ball. Ali puts the BLANK. The ball is touching the box. & 0.938 & 11.50 & ball inside the box / box inside the ball (\checkmark) \\
\midrule
\multirow{5}{*}{\shortstack[l]{Low human\\Low gemma-3-27b}}
& [ctx] The tennis ball is BLANK the bowling ball. Ali rolls the bowling ball. The bowling ball is further from the tennis ball. & 0.438 & $-35.25$ & touching / far from ($\times$) \\
& [ctx] Ali is holding the tennis ball and the ping pong ball is on the floor. Ali BLANK the ping pong ball. The ping pong ball is closer to the tennis ball. & 0.200 & $-9.50$ & drops / picks up ($\times$) \\
& [ctx] The box is next to the ball. Ali puts the BLANK. the ball is larger than the box. & 0.476 & $-4.50$ & box inside the ball / ball inside the box ($\times$) \\
& [ctx] The box is next to the desk. Ali stacks the BLANK. The box is below the desk. & 0.571 & $-5.75$ & desk on the box / box on the desk ($\times$) \\
& [tgt] The box is next to the ball. Ali puts the ball inside the box. The ball is BLANK the box. & 0.688 & $-7.00$ & touching / not touching (--) \\
\bottomrule
\end{tabular}
\end{table}

\begin{table}[H]
\centering
\small
\caption{ state\_follows (n=40)}
\begin{tabular}{@{}p{0.18\textwidth} p{0.49\textwidth} p{0.06\textwidth} p{0.08\textwidth} p{0.15\textwidth}@{}}
\toprule
Quadrant & Prompt & Hum. & Gemma $\Delta$ & A / B (correct) \\
\midrule
\multirow{5}{*}{\shortstack[l]{High human\\High gemma-3-27b}}
& [tgt] The paper is in one piece. Ali rips the paper. The paper is in BLANK. & 1.000 & 43.00 & two pieces / one piece (\checkmark) \\
& [tgt] The stove is on high heat. Ali puts his hand on the stove. Ali's hand is BLANK. & 1.000 & 41.00 & burnt / not burnt (\checkmark) \\
& [tgt] The ice cream is on the table. Ali puts the ice cream in the freezer. The ice cream gets BLANK. & 1.000 & 39.50 & colder / warmer (\checkmark) \\
& [ctx] The paper is in two pieces. Ali BLANK the pieces. The paper is in one piece. & 1.000 & 37.00 & tapes / rips (\checkmark) \\
& [ctx] The cup is empty. Ali pours BLANK in the cup. The cup is wet. & 1.000 & 36.50 & water / sand (\checkmark) \\
\midrule
\multirow{5}{*}{\shortstack[l]{High gemma-3-27b\\Low human}}
& [tgt] The soup is on the stove. Ali turns on the stove. The soup gets BLANK. & 0.850 & 42.75 & warmer / colder (\checkmark) \\
& [tgt] The glass is on the table. Ali bumps the table. The glass BLANK. & 0.750 & 32.00 & breaks / doesn't break (\checkmark) \\
& [tgt] The icecube is in the freezer. Ali puts the icecube in the frying pan. The icecube BLANK. & 0.818 & 34.00 & melts / doesn't melt (\checkmark) \\
& [tgt] The room is dark. Ali turns on the light. The room gets BLANK. & 0.909 & 36.75 & lighter / darker (\checkmark) \\
& [tgt] The room is bright. Ali turns off the lights. The room is BLANK. & 0.900 & 35.50 & dark / bright (\checkmark) \\
\midrule
\multirow{5}{*}{\shortstack[l]{Low gemma-3-27b\\High human}}
& [ctx] Ali is standing on BLANK. Ali drops the glass. The glass breaks. & 1.000 & 18.00 & tile / carpet (\checkmark) \\
& [tgt] The wall was painted 5 minutes ago. Ali touches the wall. Ali's hand is BLANK. & 1.000 & 22.00 & painted / clean (--) \\
& [tgt] The cup is empty. Ali pours water in the cup. The cup is BLANK. & 1.000 & 28.25 & wet / dry (--) \\
& [ctx] The glass is on the BLANK. Ali bumps the table. The glass breaks. & 0.955 & 20.75 & table / floor (\checkmark) \\
& [ctx] The soup is on the BLANK. Ali turns on the stove. The soup gets warmer. & 1.000 & 29.75 & stove / table (\checkmark) \\
\midrule
\multirow{5}{*}{\shortstack[l]{Low human\\Low gemma-3-27b}}
& [tgt] Ali is standing on carpet. Ali drops the glass. The glass BLANK. & 0.818 & $-41.75$ & doesn't break / breaks ($\times$) \\
& [ctx] The icecube is in the freezer. Ali puts the icecube in the BLANK. The icecube melts. & 0.810 & 14.00 & frying pan / snow (--) \\
& [ctx] The wall was painted BLANK ago. Ali touches the wall. Ali's hand is painted. & 0.750 & 20.25 & 5 minutes / 1 month (\checkmark) \\
& [ctx] The window is BLANK Ali. Ali throws the baseball forward. The window breaks. & 0.857 & 15.75 & in front of / behind (\checkmark) \\
& [ctx] The room is bright. Ali BLANK the lights. The room is dark. & 0.867 & 0.00 & turns off / turns on (\checkmark) \\
\bottomrule
\end{tabular}
\end{table}

\begin{table}[H]
\centering
\small
\caption{ state\_nonfollows (n=36)}
\begin{tabular}{@{}p{0.18\textwidth} p{0.49\textwidth} p{0.06\textwidth} p{0.08\textwidth} p{0.15\textwidth}@{}}
\toprule
Quadrant & Prompt & Hum. & Gemma $\Delta$ & A / B (correct) \\
\midrule
\multirow{5}{*}{\shortstack[l]{High human\\High gemma-3-27b}}
& [tgt] The full glass is on the table. Ali knocks the glass over. The table is BLANK. & 0.950 & 33.00 & wet / dry (\checkmark) \\
& [ctx] The paper is in one piece. Ali BLANK the paper. The paper is in one piece. & 0.955 & 21.25 & folds / rips (\checkmark) \\
& [tgt] The cup is empty. Ali pours sand in the cup. The cup is BLANK. & 1.000 & 10.75 & dry / wet (--) \\
& [ctx] The BLANK glass is on the table. Ali knocks the glass over. The table is wet. & 0.933 & 21.00 & full / empty (\checkmark) \\
& [ctx] The stove is BLANK. Ali puts his hand on the stove. Ali's hand is not burnt. & 0.857 & 27.00 & off / on (\checkmark) \\
\midrule
\multirow{5}{*}{\shortstack[l]{High gemma-3-27b\\Low human}}
& [ctx] The glass is on the floor. Ali bumps the BLANK. The glass doesn't break. & 0.400 & 24.75 & table / glass (\checkmark) \\
& [ctx] The glass and the cup are on the table. Ali pushes the BLANK off the table. The glass doesn't break. & 0.524 & 26.00 & cup / glass (\checkmark) \\
& [ctx] The ice cube and the banana are in the fridge. Ali takes the BLANK out of the fridge. The ice cube doesn't melt. & 0.619 & 28.75 & banana / ice cube (\checkmark) \\
& [tgt] The glass and the cup are on the table. Ali pushes the cup off the table. The glass BLANK. & 0.450 & 6.00 & doesn't break / breaks (\checkmark) \\
& [tgt] The ice cube is on the floor. Ali moves the ice cube to the counter. The ice cube gets BLANK. & 0.810 & 29.00 & warmer / colder (\checkmark) \\
\midrule
\multirow{5}{*}{\shortstack[l]{Low gemma-3-27b\\High human}}
& [tgt] The wine glass is on the floor. Ali touches the glass. The glass BLANK. & 0.938 & $-34.25$ & doesn't break / breaks ($\times$) \\
& [ctx] The soup is on the BLANK. Ali turns on the stove. The soup gets colder. & 0.952 & $-15.75$ & table / stove ($\times$) \\
& [tgt] The icecube is in the freezer. Ali puts the icecube in the snow. The icecube BLANK. & 0.952 & $-14.00$ & doesn't melt / melts ($\times$) \\
& [tgt] The paper is in one piece. Ali folds the paper. The paper is in BLANK. & 0.900 & $-20.00$ & one piece / two pieces ($\times$) \\
& [tgt] Ali is holding the book. Ali drops the book. The book BLANK. & 0.952 & $-7.25$ & doesn't break / breaks (--) \\
\midrule
\multirow{5}{*}{\shortstack[l]{Low human\\Low gemma-3-27b}}
& [tgt] The soup is on the table. Ali turns on the stove. The soup gets BLANK. & 0.312 & $-30.00$ & colder / warmer ($\times$) \\
& [tgt] The glass is on the floor. Ali bumps the table. The glass BLANK. & 0.636 & $-35.75$ & doesn't break / breaks ($\times$) \\
& [tgt] The empty glass is on the table. Ali knocks the glass over. The table is BLANK. & 0.619 & $-32.00$ & dry / wet ($\times$) \\
& [tgt] The window is behind Ali. Ali throws the baseball forward. The window BLANK. & 0.455 & $-20.00$ & doesn't break / breaks ($\times$) \\
& [ctx] The cup is empty. Ali pours BLANK in the cup. The cup is dry. & 0.762 & $-27.00$ & sand / water ($\times$) \\
\bottomrule
\end{tabular}
\end{table}

\begin{table}[H]
\centering
\small
\caption{ two\_refs\_egocentric (n=42)}
\begin{tabular}{@{}p{0.18\textwidth} p{0.49\textwidth} p{0.06\textwidth} p{0.08\textwidth} p{0.15\textwidth}@{}}
\toprule
Quadrant & Prompt & Hum. & Gemma $\Delta$ & A / B (correct) \\
\midrule
\multirow{5}{*}{\shortstack[l]{High human\\High gemma-3-27b}}
& [tgt] Ali and Mark are facing away from eachother. Mark turns around. Mark is facing BLANK Ali. & 0.900 & 28.75 & towards / away from (\checkmark) \\
& [tgt] Ali is facing away from Mark. Ali turns around. Ali is facing BLANK Mark. & 0.850 & 34.75 & towards / away from (\checkmark) \\
& [tgt] Mark is behind Ali. Ali turns around. Ali is facing BLANK Mark. & 0.909 & 19.75 & towards / away from (\checkmark) \\
& [ctx] Mark is to the BLANK Ali. Mark looks to his left and Ali looks to his right. Ali is facing away from Mark. & 0.867 & 23.50 & left of / towards (\checkmark) \\
& [tgt] Mark is behind Ali. Mark walks forwards. Mark is getting BLANK Ali. & 0.750 & 32.75 & closer to / further from (\checkmark) \\
\midrule
\multirow{5}{*}{\shortstack[l]{High gemma-3-27b\\Low human}}
& [ctx] Mark is BLANK Ali. Mark walks backwards. Mark is getting further from Ali. & 0.476 & 25.50 & behind / in front of (\checkmark) \\
& [ctx] Mark and Ali are both facing BLANK. Ali turns to his left. Ali is facing away from Mark. & 0.429 & 2.00 & right / left (--) \\
& [tgt] Ali and Mark are facing eachother. Mark turns around. Ali is facing BLANK Mark. & 0.438 & 3.50 & towards / away from (\checkmark) \\
& [ctx] Mark is facing BLANK Ali. Ali turns around. Mark is facing away Ali. & 0.364 & $-8.00$ & away from / towards ($\times$) \\
& [ctx] Ali is to Mark's BLANK. Mark turns around. Ali is to Mark's left. & 0.688 & 18.00 & right / left (\checkmark) \\
\midrule
\multirow{5}{*}{\shortstack[l]{Low gemma-3-27b\\High human}}
& [ctx] Mark is BLANK Ali. Ali turns around. Ali is facing away from Mark. & 0.818 & $-17.75$ & in front of / behind ($\times$) \\
& [tgt] Ali is to the right of Mark. Mark looks to his left and Ali looks to his right. Ali is facing BLANK Mark. & 0.762 & $-19.00$ & away from / towards ($\times$) \\
& [tgt] Ali is to the right of Mark. Mark looks to his left. Mark BLANK Ali. & 0.750 & $-20.75$ & can't see / sees ($\times$) \\
& [ctx] Ali is to the BLANK Mark. Mark looks to his left. Mark can't see Ali. & 0.875 & $-8.50$ & right of / left of ($\times$) \\
& [ctx] Ali and Mark are facing BLANK eachother. Mark turns around. Mark is facing towards Ali. & 0.867 & $-9.50$ & away from / towards ($\times$) \\
\midrule
\multirow{5}{*}{\shortstack[l]{Low human\\Low gemma-3-27b}}
& [ctx] Ali is to Mark's BLANK. Ali turns around. Ali is to Mark's right. & 0.333 & $-17.50$ & right / left ($\times$) \\
& [ctx] Mark and Ali are facing BLANK eachother. Mark turns around. Ali is facing away from Mark. & 0.190 & $-16.00$ & away from / towards ($\times$) \\
& [tgt] Mark is facing away from Ali. Ali turns around. Mark is facing BLANK Ali. & 0.500 & $-24.25$ & away / towards ($\times$) \\
& [tgt] Mark and Ali are facing away from eachother. Mark turns around. Ali is facing BLANK Mark. & 0.550 & $-30.50$ & away from / towards ($\times$) \\
& [tgt] Mark is in front of Ali. Mark walks forwards. Mark is getting BLANK Ali. & 0.545 & $-27.50$ & further from / closer to ($\times$) \\
\bottomrule
\end{tabular}
\end{table}

\begin{table}[H]
\centering
\small
\caption{ two\_refs\_geocentric (n=30)}
\begin{tabular}{@{}p{0.18\textwidth} p{0.49\textwidth} p{0.06\textwidth} p{0.08\textwidth} p{0.15\textwidth}@{}}
\toprule
Quadrant & Prompt & Hum. & Gemma $\Delta$ & A / B (correct) \\
\midrule
\multirow{5}{*}{\shortstack[l]{High human\\High gemma-3-27b}}
& [ctx] Mark is East of Budapest and Ali is West of Budapest. Ali walks directly BLANK Mark. Ali is heading East. & 0.875 & 49.25 & towards / away from (\checkmark) \\
& [tgt] Mark is facing East and Ali is facing West. Mark turns around. Mark is facing BLANK. & 0.800 & 24.25 & West / East (\checkmark) \\
& [tgt] Mark is East of Budapest and Ali is West of Budapest. Mark walks directly towards Ali. Mark is heading BLANK. & 0.857 & 18.75 & West / East (\checkmark) \\
& [ctx] Mark and Ali are both directly West of Budapest. Mark walks BLANK. Ali is North of Mark. & 0.867 & 13.00 & South / North (\checkmark) \\
& [ctx] Mark is facing East and Ali is facing BLANK. Mark turns 180 degrees. Mark and Ali are facing the same direction. & 0.750 & 17.25 & West / East (\checkmark) \\
\midrule
\multirow{5}{*}{\shortstack[l]{High gemma-3-27b\\Low human}}
& [ctx] Mark is East of Budapest and Ali is West of Budapest. Mark walks directly BLANK Ali. Mark is heading West. & 0.571 & 40.75 & towards / away from (\checkmark) \\
& [ctx] Milwaukee is South of Mark. Mark goes BLANK. Milwaukee is Southeast of Mark. & 0.312 & 6.00 & West / East (--) \\
& [tgt] Mark is West of Ali facing East. Ali is East of Mark facing West. Mark is facing BLANK Ali. & 0.438 & 12.00 & towards / away from (\checkmark) \\
& [tgt] Mark and Ali are both directly East of Budapest. Mark walks North. Ali is BLANK of Mark. & 0.562 & 14.25 & South / North (\checkmark) \\
& [tgt] Mark is East of Budapest and Ali is West of Budapest. Ali walks directly towards Mark. Ali is heading BLANK. & 0.650 & 30.25 & East / West (\checkmark) \\
\midrule
\multirow{5}{*}{\shortstack[l]{Low gemma-3-27b\\High human}}
& [tgt] Milwaukee is North of Mark. Mark goes East. Milwaukee is BLANK of Mark. & 0.762 & $-13.00$ & Northwest / Southeast ($\times$) \\
& [ctx] Mark and Ali are both directly East of Budapest. Mark walks BLANK. Ali is South of Mark. & 0.909 & 1.25 & North / South (\checkmark) \\
& [tgt] Mark is East of Ali. Mark turns right. Mark is BLANK of Ali. & 0.909 & 2.00 & East / West (--) \\
& [tgt] Mark is East of Ali facing East. Ali is West of Mark facing West. Mark is facing BLANK Ali. & 0.750 & $-6.50$ & away from / towards ($\times$) \\
& [ctx] Mark is North of Budapest and Ali is South of Budapest. Ali walks directly BLANK Mark. Ali is walking South. & 0.636 & $-24.50$ & away from / towards ($\times$) \\
\midrule
\multirow{5}{*}{\shortstack[l]{Low human\\Low gemma-3-27b}}
& [ctx] Mark is BLANK of Ali. Mark turns right. Mark is East of Ali. & 0.400 & $-16.25$ & East / West ($\times$) \\
& [ctx] Milwaukee is South of Mark. Mark goes BLANK. Mark is Northeast of Milwaukee. & 0.273 & $-3.50$ & West / East (--) \\
& [tgt] Milwaukee is South of Mark. Mark goes West. Mark is BLANK of Milwaukee. & 0.312 & $-2.50$ & Northest / Southeast ($\times$) \\
& [ctx] Milwaukee is North of Mark. Mark goes BLANK. Milwaukee is Northwest of Mark. & 0.286 & $-1.50$ & East / West (--) \\
& [tgt] Mark is facing South and Ali is facing East. Mark turns 90 degrees to his right. Mark and Ali are facing BLANK. & 0.476 & $-9.50$ & the same direction / different directions ($\times$) \\
\bottomrule
\end{tabular}
\end{table}
\newpage
%Bibliography
\bibliographystyle{plainnat}
\bibliography{references}
\clearpage

\end{document}